\documentclass[10pt,twocolumn,letterpaper]{article}

\usepackage[pagenumbers]{cvpr} 

\usepackage[accsupp]{axessibility}  
\usepackage{graphicx}
\usepackage{amsmath}
\usepackage{amssymb}
\usepackage{booktabs}

\usepackage{pifont}
\newcommand{\cmark}{\ding{51}}
\newcommand{\xmark}{\ding{55}}
\usepackage{enumitem}
\usepackage{float}
\usepackage{adjustbox}

\usepackage[pagebackref,breaklinks,colorlinks]{hyperref}

\usepackage[capitalize]{cleveref}
\crefname{section}{Sec.}{Secs.}
\Crefname{section}{Section}{Sections}
\Crefname{table}{Table}{Tables}
\crefname{table}{Tab.}{Tabs.}

\def\cvprPaperID{7461} 
\def\confName{CVPR}
\def\confYear{2023}

\begin{document}

\title{Modernizing Old Photos Using Multiple References \\ via Photorealistic Style Transfer}

\author{Agus Gunawan$^{1}$ \quad Soo Ye Kim$^{1,2}$\footnotemark[1] \quad Hyeonjun Sim$^{1}\footnotemark[2]$ \quad Jae-Ho Lee$^{3}$ \quad Munchurl Kim$^{1}$\footnotemark[3]\\
[0.2em]
$^1$KAIST \quad
$^2$Adobe Research \quad
$^3$ETRI
\\
{\tt\small \{agusgun, flhy5836, mkimee\}@kaist.ac.kr \quad sooyek@adobe.com \quad jhlee3@etri.re.kr}
}

\twocolumn[{%
\renewcommand\twocolumn[1][]{#1}%
\maketitle

\begin{center}
\centering
\captionsetup{type=figure}
\vspace{-1.5em}
\includegraphics[width=0.98\textwidth]{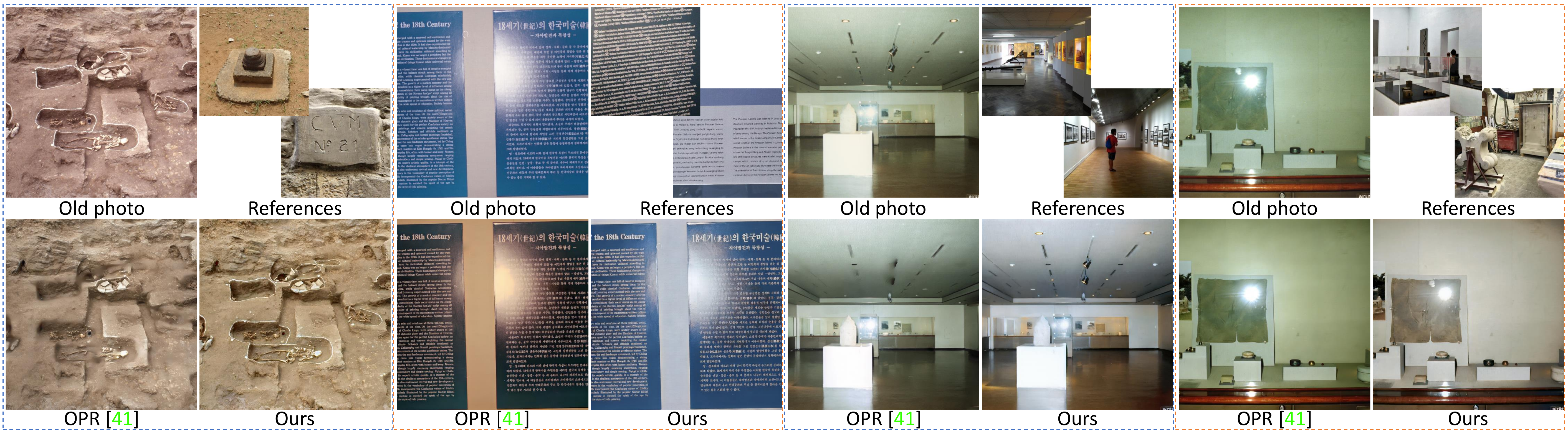}
\vspace{-0.8em}
\captionof{figure}{The results of old photo modernization produced by our method. Our method is able to generate more modern-looking images that resemble the style of input reference images \textbf{without the use of old photos during training}. Please visit our webpage at \href{https://kaist-viclab.github.io/old-photo-modernization}{https://kaist-viclab.github.io/old-photo-modernization} for dataset and code.}
\label{fig:results_first_page}
\vspace{0em}
\end{center}%
}]

{
    \renewcommand{\thefootnote}
        {\fnsymbol{footnote}}
    \footnotetext[1]{Soo Ye Kim is currently affiliated with Adobe Research.}
    \footnotetext[2]{Hyeonjun Sim is currently affiliated with Qualcomm.}
    \footnotetext[3]{Corresponding author.}
}

\begin{abstract}
   This paper firstly presents old photo modernization using multiple references by performing stylization and enhancement in a unified manner. 
   In order to modernize old photos, we propose a novel multi-reference-based old photo modernization (MROPM) framework consisting of a network MROPM-Net and a novel synthetic data generation scheme. 
   MROPM-Net stylizes old photos using multiple references via photorealistic style transfer (PST) and further enhances the results to produce modern-looking images.
   Meanwhile, the synthetic data generation scheme trains the network to effectively utilize multiple references to perform modernization.
   To evaluate the performance, we propose a new old photos benchmark dataset (CHD) consisting of diverse natural indoor and outdoor scenes. 
   Extensive experiments show that the proposed method outperforms other baselines in performing modernization on real old photos, even though no old photos were used during training. 
   Moreover, our method can appropriately select styles from multiple references for each semantic region in the old photo to further improve the modernization performance.
\end{abstract}
%

\section{Introduction}
\label{sec:intro}
Old photos taken a long time ago may contain important information that carry cultural and heritage values, e.g., photos of Queen Elizabeth II's coronation.
Such old images may contain multiple degradations, e.g., scratches, and old photo artifacts, e.g., color fading, often preventing people from understanding the scene.
To restore these images, a skilled expert needs to perform laborious manual processes such as degradation restoration and modernization, i.e., colorization or enhancement, to make them look modern \cite{whitt2014beginning}.
Consequently, early studies \cite{stanco2003towards, giakoumis2005digital} try to restore damaged old photos automatically by using traditional inpainting techniques.
However, solely re-synthesizing damaged regions in the image is inadequate to ensure old photos look modern, as the overall style remains similar.

Recent work \cite{luo2021time} formulates the task as time-travel rephotography which aims to translate old photos into a modern photos space.
The authors considered a multi-task problem consisting of two main tasks:
(i) restoration of old photos with both unstructured (noise, blur) and structured (scratch, crack) degradations;
(ii) modernization which aims to change old photos' characteristics to look like modern images, e.g., better color saturation and contrast by using colorization \cite{luo2021time, xu2022pik} or enhancement \cite{wan2020bringing}.
However, simply using an enhancement method \cite{wan2020bringing}  fails to modernize old photos, as shown in Fig. \ref{fig:results_first_page}, since the overall look still remains similar to old photos, e.g., with a sepia color.

In this paper, we propose to modernize old color photos of natural scenes by changing their styles and enhancing them to look modern.
For this, a novel unified framework is proposed which leverages multiple modern photo references in solving the modernization task of old photos by utilizing photorealistic style transfer (PST).
Although one prior work \cite{xu2022pik} is also reference-based, it only relies on a single reference to colorize greyscale portrait photos.
However, in natural scene cases, it is challenging to find a single modern photo as a reference that can well match the whole semantics of an old photo.
Moreover, changing only the color is not sufficient to alter the overall look of an image \cite{hu2020aesthetic}.
Thus, our framework uses multiple references to modernize old photos by changing the \textit{style} instead of only the color.
Since there is no public old photos benchmark dataset of natural scenes, we propose a new Cultural Heritage Dataset (CHD) with 644 indoor and outdoor old color photos collected from various national museums in Korea.

Our multiple-reference-based old photo modernization framework (MROPM) consists of two main parts: (i) \textit{MROPM-Net} and (ii) \textit{a novel training strategy} that enables the network to utilize multiple references.
The MROPM-Net consists of two different subnets: The first is a single stylization subnet that transfers both global and local styles without any semantic segmentation from a modern photo into an old photo;
Specifically, we propose an improved version of WCT2 \cite{yoo2019photorealistic}, inspired by its universal generalization, as the backbone of the single stylization subnet, and present a new architecture that can perform both global PST and local PST without requiring any semantic segmentation;
The second is a merging-refinement subnet that merges multiple stylization results from multiple references based on semantic similarities and further refines the merged result to produce a modernized version of the old photo.
To effectively train the MROPM-Net, we propose a synthetic data generation scheme that uses the style-variant (i.e., color jittering and unstructured degradation) and -invariant (i.e., rotation, flipping, and translation) transformations.
Our MROPM can modernize old photos better than the state-of-the-art (SOTA) old photo restoration method \cite{wan2020bringing}, even without using any old photos during training, thanks to the generalization of PST.
Our contributions are summarized as follows:
\begin{itemize}
 \itemsep 0em
 \item We propose the \textit{first} old photo modernization framework (MROPM) that allows the usage of \textit{multiple references} to guide the modernization process.
 \item Our \textit{photorealistic multi-stylization network} and training strategy enable the MROPM-Net to utilize multiple style references in modernizing old photos.
 \item Our training strategy based on synthetic data allows the MROPM-Net to modernize \textit{real} old photos even without using any old photos during training.
 \item We propose a new old photo dataset of natural scenes, called Cultural Heritage Dataset (CHD), with 644 outdoor and indoor cultural heritage images.
\end{itemize}

\section{Related Work}
\label{sec:related_work}

\noindent
\textbf{Reference-based color-related tasks.}\quad
One way to change the overall look of an image is by changing color, which is one of the style components \cite{hu2020aesthetic}.
To change the color of old photos, one can employ two methods: \textit{exemplar-based colorization} \cite{he2018deep, su2020instance, xu2020stylization, lu2020gray2colornet, wu2021towards, yin2021yes} and \textit{color transfer} or \textit{recolorization} \cite{liao2017visual, he2019progressive, afifi2021histogan, lee2020deep}.
However, exemplar-based colorization methods cannot utilize the color information in the input images for matching, although color is an important feature representing object semantics \cite{singh2020assessing}, limiting the methods for the modernization of old color photos.
Color transfer aims to transfer the reference image's color statistics into the input image.
Early deep learning works \cite{liao2017visual, he2019progressive} use deep feature matching from features extracted with pre-trained VGG19 \cite{simonyan2014very} to perform the color transfer, which can also be extended to multi-reference cases \cite{he2019progressive}.
Due to the long execution time of the optimization process, recent works develop end-to-end networks, where Lee \textit{et al}. \cite{lee2020deep} utilize color histogram analogy, and Afifi \textit{et al}. \cite{afifi2021histogan} utilize a color-controlled generative adversarial network (GAN).
However, recent works can only use a single reference, where finding a single reference image containing similar semantics as the input old photo can be challenging.
Thus, from the perspective of color transfer, our work is the first end-to-end network that can utilize multiple references to handle content mismatch without any slow optimization technique.

\noindent
\textbf{Photorealistic style transfer (PST).}\quad
The PST aims at achieving photorealistic rendering of an image with the style of another image.
Since the development of postprocessing and regularization techniques \cite{mechrez2017photorealistic, luan2017deep}, PST has gained much popularity.
Recent works can be categorized into architecture \cite{li2018closed, yoo2019photorealistic, an2020ultrafast, xia2020joint, qu2021non,chiu2022pca, chiu2022photowct2} and feature transformation \cite{li2017universal, li2019learning, huo2021manifold} improvements to effectively and efficiently produce photorealistic results.
Specifically, WCT2 \cite{yoo2019photorealistic} utilizes wavelet-based skip connection and progressive stylization to achieve better PST where the method can work universally without re-training to pre-defined styles.
Due to these benefits, we base our network architecture on WCT2.
However, WCT2 produces unnatural style transfer results when performing global and local stylization with unreliable semantic segmentation (shown in Supplementary Material), which hinders the application to old photos.
Thus, our MROPM-Net is designed to enable local stylization without any semantic segmentation, which in consequence, can perform multi-style PST in one unified framework without specifying any masks.
To the best of our knowledge, this is the first work in multi-style PST, although there is one work in multi-style artistic style transfer (AST) \cite{huang2019style}.
Note that the AST is different from the PST in that it utilizes learning-dependent feature transformation, which can cause severe visual artifacts in PST.

\noindent
\textbf{Old photo restoration.}\quad
Early works in old photo restoration focus on detecting and restoring structured degradation (scratch and crack) of images using traditional inpainting techniques \cite{stanco2003towards, giakoumis2005digital}.
Besides the structured degradation, \cite{wan2020bringing, liu2021cg, xu2022pik} incorporate additional spatially-uniform unstructured degradation, e.g., blur and noise, using synthetic degradation and formulate the problem as mixed degradation restoration.
However, restoring mixed degradation is not enough to ensure that old photos look modern.
Consequently, Luo \textit{et al}. \cite{luo2021time} formally introduce the time-travel rephotography problem, which aims to translate old photos to look like ones taken in the modern era.
This problem adds modernization, synthesizing the missing colors and enhancing the details, on top of degradation restoration.
To solve the modernization problem, Luo \textit{et al}. \cite{luo2021time} use a StyleGAN2-generated \cite{karras2020analyzing} sibling image to serve as a reference for old portrait photos.
However, generating complex natural scene photos via GAN to be used as references is challenging \cite{brock2018large}, making the method unable to be applied to natural scene old photos.
Another work \cite{xu2022pik} proposes to use a single reference image to colorize an old greyscale photo.
However, using a single reference is not enough to cover the whole semantics of old photos (shown in Fig. \ref{fig:results_first_page}).
Thus, different from previous methods, we propose to modernize old photos by stylizing and enhancing old photos in a unified manner using \textit{multiple references} to better cover the entire semantics of old photos.

\section{Proposed Cultural Heritage Dataset (CHD)}
\label{sec:dataset}
Although some public datasets such as Historical Wiki Face Dataset (HWFD) \cite{luo2021time} and RealOld \cite{xu2022pik} have been released recently, these datasets only contain portrait or face photos which are much simpler compared to natural scenes.
In addition, these datasets only contain greyscale photos and disregard color photos produced during the 20th century using reversal films \cite{machidon2018digital}, which have specific degradations such as color dye fading and have not been analyzed before.
Therefore, we propose a Cultural Heritage Dataset (CHD) consisting of 644 old color photos produced in the 20th century.
Specifically, we collect these old photos in the form of reversal films or papers from three national museums in Korea, which are then scanned in resolutions varying from 4K to 8K.
The photos have been well preserved and stored carefully due to their value, containing little structured degradation, e.g., scratches, but varying degrees of unstructured degradation, e.g., noises.
These photos contain indoor and outdoor scenes of cultural heritage, such as special exhibitions and excavation ruins.
After collection, all photos are divided into train and test sets by randomly splitting with a proportion of 8 (514 photos):2 (130 photos).
The train set is only used for other baselines that need to be trained using real old photos.
Since our task is reference-based old photo modernization, we further collect modern photos as references by crawling images with similar contexts from the internet.
In total, we obtain 130 old photos in the test set, each of which has one or two references selected manually.
Further details can be found in \textit{Supplementary Material}.

\begin{figure}[t]
\centering
\includegraphics[width=0.80\columnwidth]{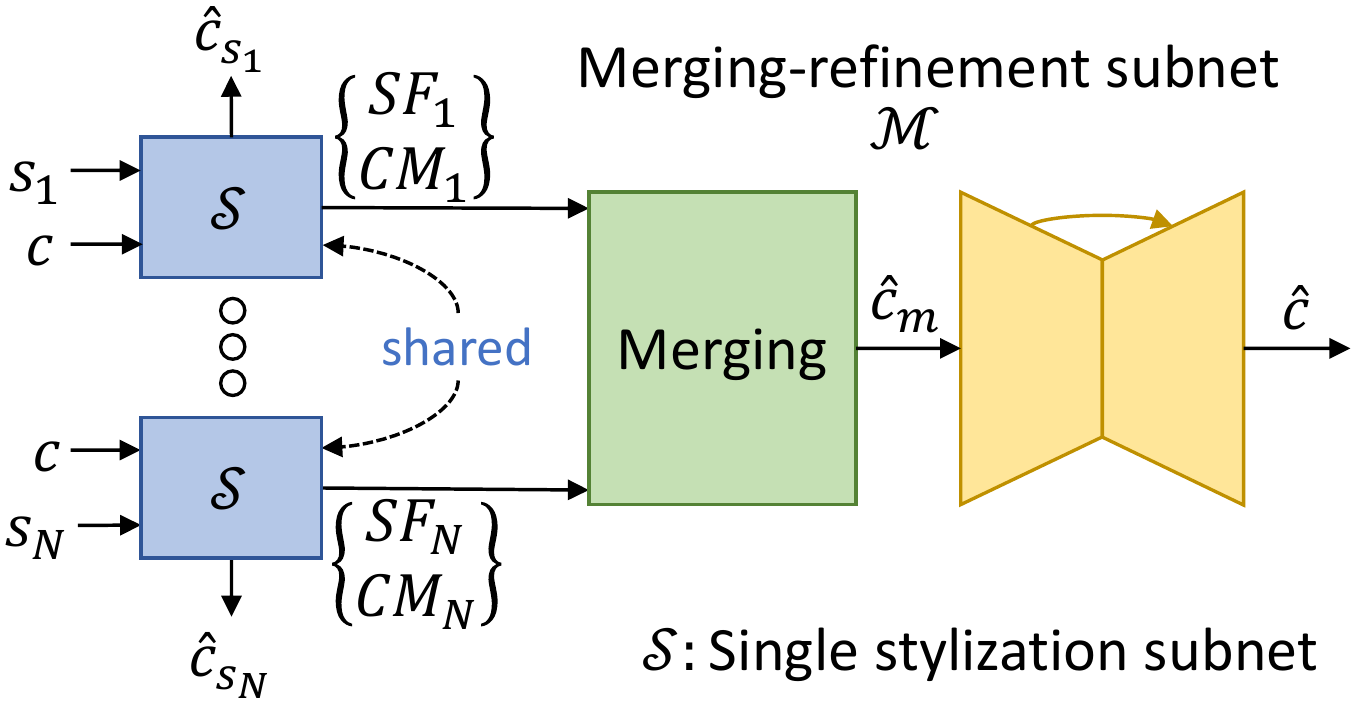}
\vspace{-0.8em}
\caption{The overall framework of our multiple-reference-based old photo modernization network (MROPM-Net).}
\label{fig:method_framework}
\vspace{-1.2em}
\end{figure}

\begin{figure*}[t]
\centering
\includegraphics[width=0.92\textwidth]{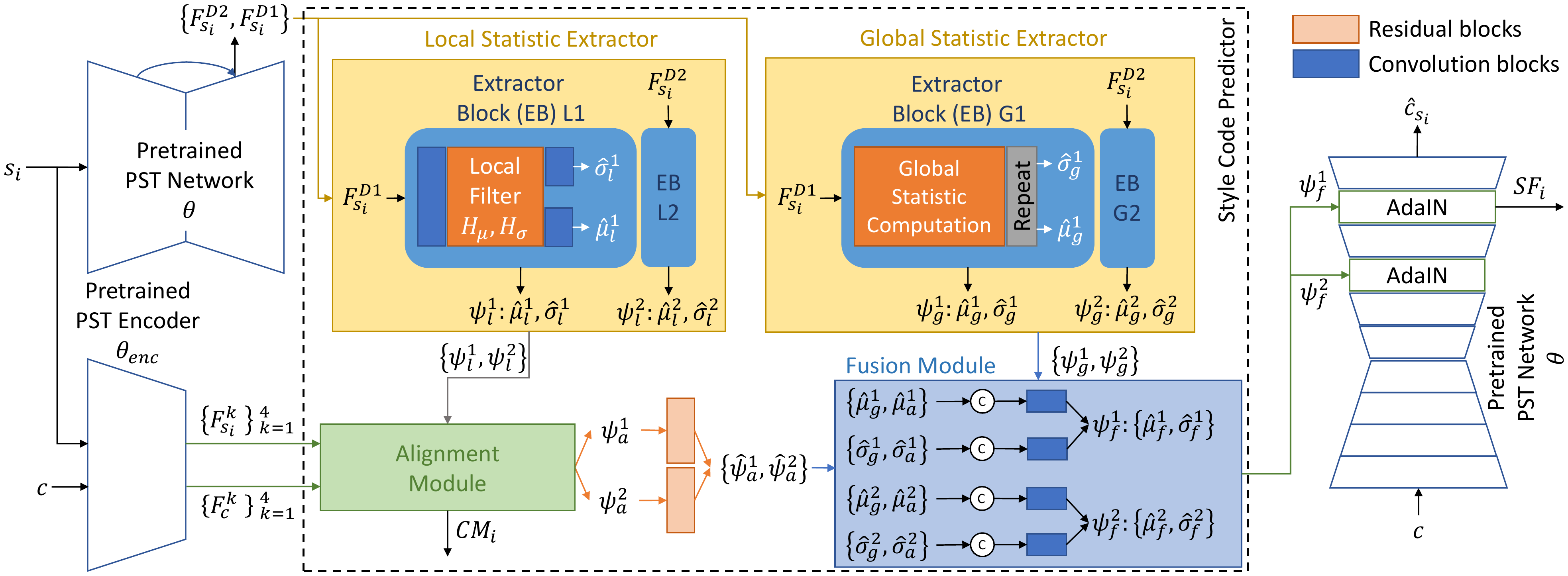}
\vspace{-0.8em}
\caption{The architecture of the single stylization subnet $\mathcal{S}$.}
\label{fig:method_single_stylization}
\vspace{-1em}
\end{figure*}

\section{Proposed Method}
\label{sec:method}

\subsection{Overall Framework}
Fig. \ref{fig:method_framework} shows our proposed multi-reference-based old photo modernization network (MROPM-Net) with a shared single stylization subnet $\mathcal{S}$ and a merging-refinement subnet $\mathcal{M}$.
We denote an old photo input as content $c \in \mathbb{R}^{H \times W \times 3}$ and $N$ number of modern photos as styles $\boldsymbol{s}=\{s_i\}_{i=1}^N \in \mathbb{R}^{N \times H \times W \times 3}$.
Our goal is to modernize $c$ using $\boldsymbol{s}$.
In the first step, we utilize $\mathcal{S}$, which is built based on a photorealistic style transfer (PST) backbone, to stylize $c$ using each $s_i$, yielding $N$ stylized features and correlation matrices $\{SF_i, CM_i\}_{i=1}^N$.
After having multiple stylization results, we merge the features $\{SF_i\}_{i=1}^N$ based on the semantic similarity $\{CM_i\}_{i=1}^N$ between $c$ and $\boldsymbol{s}$ and further refine the merging result via $\mathcal{M}$.
Specifically, $\mathcal{M}$ selects the appropriate styles for each semantic region based on multiple stylization results $\{SF_i\}_{i=1}^N$ to produce an intermediate merging image output $\hat{c}_m$,
e.g., selecting the most appropriate feature for a sky region from $SF_1$ that contains a sky style, not from $SF_{i \neq 1}$, which do not contain sky styles.
Then, $\hat{c}_{m}$ is further refined to get the final result $\hat{c}$.
Given relevant references, $\hat{c}$ becomes a modern version with a modern style and enhanced details for old photo input $c$.

\subsection{Network Architecture}

\noindent
\textbf{Single stylization subnet $\mathcal{S}$.}\quad
Fig. \ref{fig:method_single_stylization} shows a detailed structure of $\mathcal{S}$.
For given multiple references, our single stylization subnet is shared for all input pairs and takes a single pair of an old photo $c$ and a reference $s_i$ at a time.
Given a pair of $(c, s_i)$, $\mathcal{S}$ stylizes $c$ based on the style code of $s_i$ locally and globally, resulting in a stylized feature $SF_i$, a stylized old photo $\hat{c}_{s_i}$, and a correlation matrix $CM_i$.
This subnet $\mathcal{S}$ consists of two main parts: (i) \textit{an improved PST network} and (ii) \textit{a style code predictor}.

For the PST network, we improve some drawbacks of the concatenated version of WCT2 \cite{yoo2019photorealistic}.
We observed that the stylization only affects the last decoder block due to the design of its skip connection, where this issue is called a ``short circuit" in \cite{an2020ultrafast}.
Thus, instead of transferring three different high-frequency components as in the WCT2, we propose to simplify it by transferring a single high-frequency component in level-0 of the Laplacian pyramid representation \cite{burt1987laplacian}.
Second, we only apply feature transformation in the network's decoder part, especially the last two decoder blocks, which achieves the best trade-off between the stylization effect and the photorealism.
Third, we use the differentiable adaptive instance normalization (AdaIN) \cite{huang2017arbitrary} instead of the non-differentiable WCT \cite{li2017universal} to learn and predict the local style rather than compute it.

The second part of $\mathcal{S}$ is a style code predictor.
This part aims to predict style codes $\psi = \{\mu, \sigma\}$ consisting of mean and standard deviation (std), which are statistics used to perform stylization in AdaIN \cite{huang2017arbitrary}.
We propose to predict $\psi$ instead of computing it as in AdaIN to perform local style transfer without requiring any semantic segmentation.
The first step (yellow) of the style code predictor is to extract local style codes $\psi_l^j =\{\hat{\mu}_l^j,\hat{\sigma}_l^j\}$ and global style codes $\psi_g^j=\{\hat{\mu}_g^j, \hat{\sigma}_g^j\}$ from the \textit{j}-th level feature \{$F_{s_i}^{Dj}$\} extracted by the last two decoder blocks ($j = 1, 2$) of the pre-trained PST network as shown in Fig. \ref{fig:method_single_stylization}.
In this regard, $\psi_l^j$ is extracted using a local statistic extractor which consists of a local mean filter $H_{\mu}$ and local std filter $H_{\sigma}$ with a kernel size of 3 and convolution blocks to refine both filtered outputs.
Meanwhile, the global statistic extractor extracts $\psi_g^j$ by computing channel-wise mean and std values, which are then spatially repeated to the same spatial size of $\psi_l^j$.
After style code extraction, the second step (green) is to align $\psi_l^j$ to $c$ by using non-local attention \cite{wang2018non}.
Specifically, we extract multi-level feature maps $\{F_{c}^{k}\}_{k=1}^4$ and $\{F_{s_i}^{k}\}_{k=1}^4$ for both $c$ and $s_i$, respectively, map them into the same feature space using shared convolution blocks, and perform matrix multiplication between mapped features to obtain correlation matrix $CM_i$.
Then, we align $\psi^j_l$ to $c$ by using $CM_i$ via matrix multiplication.
The aligned style code $\psi_a^j$ is further refined to prevent interpolation artifacts by using residual blocks \cite{he2016deep}, resulting in a refined version $\hat{\psi}_{a}^j=\{\hat{\mu}_{a}^j, \hat{\sigma}_{a}^j\}$.
After obtaining $\hat{\psi}^j_{a}$, we fuse it with $\psi_g^j$ via the fusion module to obtain a fused style code $\psi_f^j=\{\hat{\mu}_f^j, \hat{\sigma}_f^j\}$.
The fusion module performs channel-wise concatenation for $\hat{\psi}^j_{a}$ and $\psi_g^j$, which is then fed into the following convolution blocks as shown in the blue part of Fig. \ref{fig:method_single_stylization}.

\begin{figure}[t]
\centering
\includegraphics[width=0.9\columnwidth]{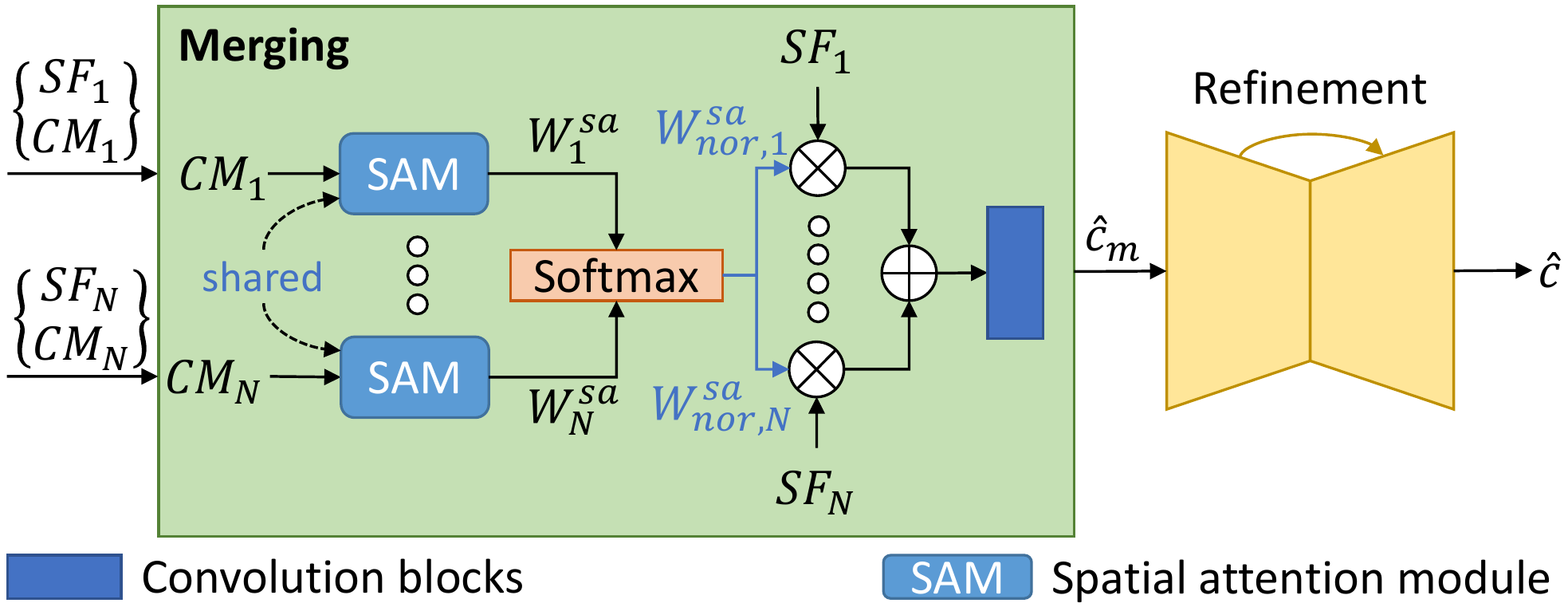}
\vspace{-0.7em}
\caption{The architecture of the merging-refinement subnet $\mathcal{M}$.}
\label{fig:merging_subnet}
\vspace{-1.3em}
\end{figure}

Finally, after performing all the operations from the local and global statistic extractors to the fusion module, we obtain $\psi_f^1$ and $\psi_f^2$.
These fused style codes are then used for stylizing $c$.
We use our PST network with AdaIN to perform the stylization as shown in the right part of Fig. \ref{fig:method_single_stylization}.

\vspace{0.1em}
\noindent
\textbf{Merging-refinement subnet $\mathcal{M}$.}\quad
After stylizing an old photo $c$ with $N$ different modern photos $\boldsymbol{s}=\{s_i\}_{i=1}^N$ using $\mathcal{S}$, we obtain multiple stylized features and correlation matrices $\{SF_i, CM_i\}_{i=1}^N$.
The next step is to select the most appropriate styles from $\{SF_i\}_{i=1}^N$ for each semantic region via the merging part of $\mathcal{M}$, as shown in Fig. \ref{fig:merging_subnet}.
For this, a spatial attention module (SAM) \cite{woo2018cbam} is employed, which strengthens and dampens semantically related and unrelated spatial features, respectively, in the merging process of the stylized features.
The SAM computes spatial attention weights $W_i^{sa}$ by using $CM_i$ for the corresponding $SF_i$. 
Then, we normalize all the spatial attention weights by using Softmax, thus having ${\boldsymbol{W}^{sa}_{nor}} = \{W_{nor,i}^{sa}\}_{i=1}^N$.
All normalized attention weights $W_{nor,i}^{sa}$ are multiplied with their corresponding $SF_i$, whose results are summed and then fed into the final convolution blocks to obtain a merging result $\hat{c}_m$ as an intermediate multi-style PST image.
We further refine $\hat{c}_m$ via the U-Net \cite{ronneberger2015u} based refinement subnet to produce a final modern version $\hat{c}$ for old photo input $c$.

\subsection{Training Strategy}

\begin{figure}[t]
\centering
\includegraphics[width=\columnwidth]{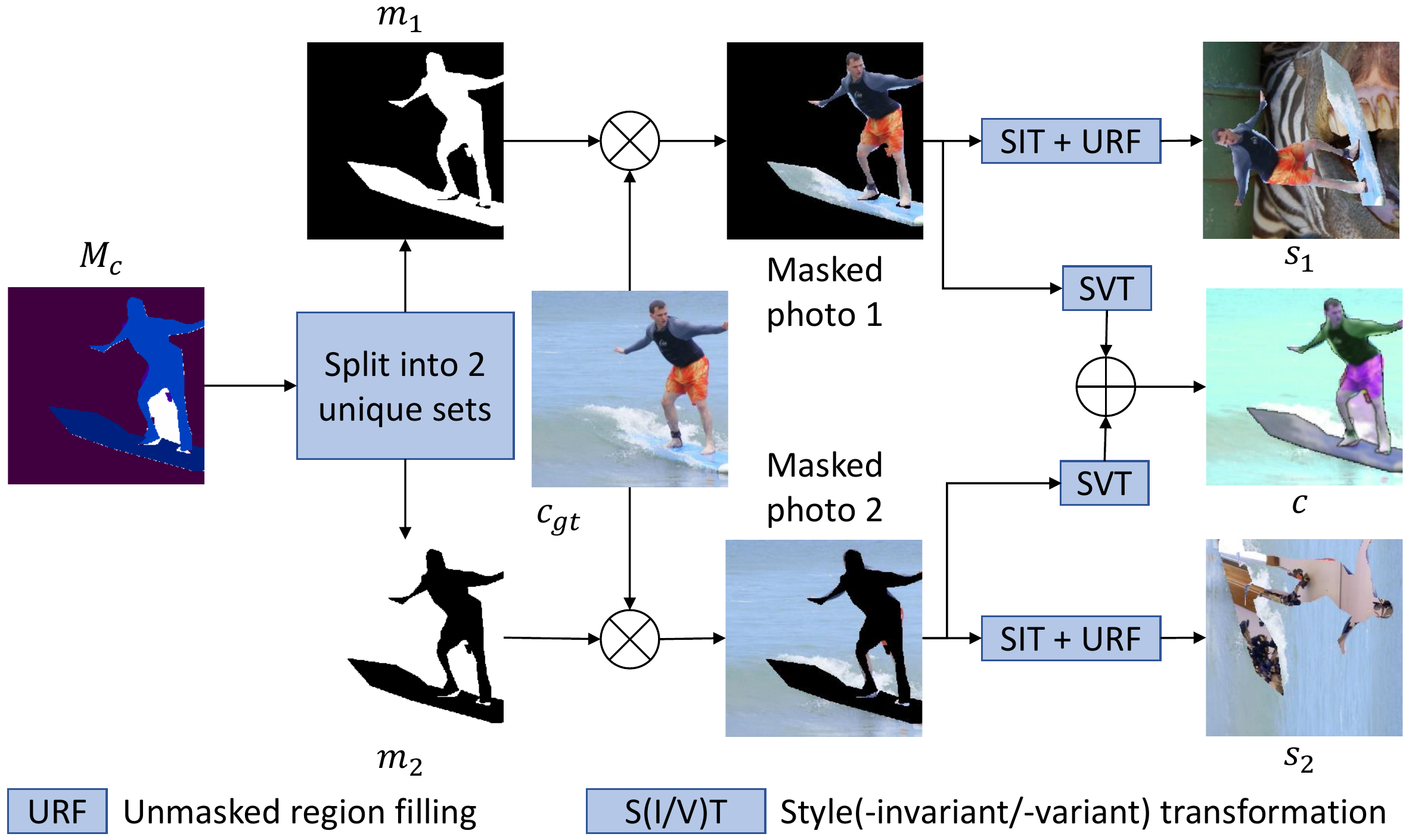}
\vspace{-1.8em}
\caption{Our synthetic data generation pipeline.}
\label{fig:synthetic_data_creation}
\vspace{-1.2em}
\end{figure}

\noindent
\textbf{Synthetic data generation.}\quad
Since there is no ground truth for multiple-reference-based old photo modernization tasks, we generate synthetic data for training the network in a self-supervised manner.
For this, the COCO-stuff dataset \cite{caesar2018coco} is utilized, which has a semantic segmentation mask for each image.
Fig. \ref{fig:synthetic_data_creation} shows the pipeline of generating the synthetic data where each sample consists of a \textit{synthetic} old photo $c$, its corresponding $N$ different references $\boldsymbol{s}=\{s_i\}_{i=1}^N$, and its ground truth $c_{gt}$.
We use two style references with $N=2$ for each sample throughout our experiments.
First, we randomly select a photo from the COCO-stuff dataset as ground truth $c_{gt}$ and its corresponding semantic mask $M_c$ at each iteration during the training.
Then, all the semantic regions in $M_c$ are randomly separated into $N$ non-overlapping parts $\{m_i\}_{i=1}^N$. 
For the example shown in Fig. \ref{fig:synthetic_data_creation}, $M_c$ consists of two semantic regions, \textit{surfer} and \textit{sea}, and thus separated to: one mask $m_1$ with the surfer, and the other mask $m_2$ with the sea.
The next step is to generate $N$ different masked photos using $\{m_i\}_{i=1}^{N}$ by element-wise multiplication between $m_i$ and $c_{gt}$.
Then, we use these masked photos to generate multiple references $\{s_i\}_{i=1}^N$ and $c$ via style-invariant (SIT) and -variant transformations (SVT) respectively.

The properties of style-variant and -invariant transformations are determined by whether the transformations alter the mean and std of any semantic region.
Hence, we use random translation (only for the regions that can be translated), rotation, and flipping as our SIT.
Meanwhile, random color jittering and unstructured degradation, i.e. blur, noise, resizing, and compression artifacts, are used for SVT.
Other types of degradation, e.g., scratches can be included in the SVT to make the method able to generalize to these types of degradation.
To generate $c$, we apply different SVTs for each masked photo and sum up the results. 
Meanwhile, to generate $\{s_i\}_{i=1}^N$, randomly selected SITs are applied for each masked photo, and then the unmasked region is filled (URF) with another photo randomly selected from the same COCO-Stuff dataset.
Our MROPM-Net can work reasonably well for real old photos after training with this synthetic data.
This is because the synthetic data make our MROPM-Net able to (i) robustly find local semantic correspondences between degraded synthetic old photo $c$ and semantically confusing synthetic modern photo $s_i$, (ii) accurately transfer the styles of each $s_i$ to $c$ locally, and (iii) merge and refine multiple stylization results from multiple styles $\{s_i\}_{i=1}^N$ to produce an output similar to $c_{gt}$. 
Thus, our synthetic data creation pipeline can be effectively used for multi-reference-based old photo modernization.

Our MROPM-Net is trained in multiple stages: 
(i) \textbf{Stage 1}: Our PST network is trained using a similar training strategy to \cite{yoo2019photorealistic};
(ii) \textbf{Stage 2}: Our single stylization subnet $\mathcal{S}$ is trained, while the pre-trained PST network is freezed;
(iii) \textbf{Stage 3}: We train our merging-refinement subnet $\mathcal{M}$, with both the pre-trained PST network and $\mathcal{S}$ freezed.

\noindent
\textbf{Loss function.}\quad
In Stage 2 of training, our goal is to obtain a faithful stylization result from each of the style reference images $\{s_i\}_{i=1}^N$.
Specifically, we use a weighted sum of the following different losses:
\begingroup
\setlength{\belowdisplayskip}{4pt} \setlength{\belowdisplayshortskip}{4pt}
\setlength{\abovedisplayskip}{4pt} \setlength{\abovedisplayshortskip}{4pt}
\begin{equation}
    \begin{aligned}
    \mathcal{L}_{s_i}^{Stage2} &= \lambda_{ML} \cdot \mathcal{L}_{ML}(\hat{c}_{s_i},c_{gt}) + \lambda_{p} \cdot \mathcal{L}_{p}(\hat{c}_{s_i},c_{gt})\\
    &+ \lambda_{CX} \cdot \mathcal{L}_{CX}(\hat{c}_{s_i}, c_{gt})
    \label{eq:loss_s1}
    \end{aligned}
\end{equation}
\endgroup
where $\mathcal{L}_{ML}$, $\mathcal{L}_{p}$ and $\mathcal{L}_{CX}$ represent masked reconstruction, perceptual \cite{johnson2016perceptual} and contextual \cite{mechrez2018contextual} losses respectively,
and the $\lambda$'s control relative weights for their respective losses.
We use the features extracted from VGG-19 \cite{simonyan2014very} at layer $relu4\_1$ for $\mathcal{L}_{p}$, and $relu3\_1$ and $relu4\_1$ for $\mathcal{L}_{CX}$.
Different from \cite{mechrez2018contextual}, GT image $c_{gt}$ is used as the reference instead of style $s_i$ for $\mathcal{L}_{CX}$ to compare with our output $\hat{c}_{s_i}$ because using $s_i$ can cause severe structure distortion.
$\mathcal{L}_{ML}$ in Eq. \ref{eq:loss_s1} can be expressed as:
\begingroup
\setlength{\belowdisplayskip}{4pt} \setlength{\belowdisplayshortskip}{4pt}
\setlength{\abovedisplayskip}{4pt} \setlength{\abovedisplayshortskip}{4pt}
\begin{align}
    \mathcal{L}_{ML}(\hat{c}_{s_i}, c_{gt}) &= \lVert (\hat{c}_{s_i} - c_{gt})  \odot m_i \rVert_1
    \label{eq:masked_l1}
\end{align}
\endgroup
where $m_i$ is a mask used to generate $s_i$ in our data generation scheme as shown in Fig. \ref{fig:synthetic_data_creation}.
Correspondingly, these three losses are used to encourage $\mathcal{S}$ (i) to faithfully stylize $c$ at the pixel level for semantic regions that also appear in $s_i$ and disregard other unrelated semantic regions, (ii) to faithfully stylize $c$ at the semantic level, and (iii) to perform better semantic style transfer.
In Stage 2 of training, we only use a single style reference for each $c$ to reduce the computation complexity and stabilize the training of the single stylization subnet $\mathcal{S}$.

In Stage 3, we train our merging-refinement subnet $\mathcal{M}$ by using weighted sum of four different losses:
\begingroup
\setlength{\belowdisplayskip}{4pt} \setlength{\belowdisplayshortskip}{4pt}
\setlength{\abovedisplayskip}{4pt} \setlength{\abovedisplayshortskip}{4pt}
\begin{equation}
    \begin{aligned}
        \mathcal{L}^{Stage3} &= \lambda_{L1} \cdot \mathcal{L}_{L1}(\hat{c}, c_{gt}) + \lambda_{p} \cdot \mathcal{L}_{p}(\hat{c}, c_{gt})\\
        &+ \lambda_{sm} \cdot \mathcal{L}_{sm}(\hat{c}) + \lambda_{adv} \cdot \mathcal{L}_{adv}(\hat{c}, c_{gt})
    \label{eq:loss_s2}
    \end{aligned}
\end{equation}
\endgroup
where $\mathcal{L}_{L1}$, $\mathcal{L}_{p}$, $\mathcal{L}_{sm}$ and $\mathcal{L}_{adv}$ are reconstruction, perceptual \cite{johnson2016perceptual}, local smoothness \cite{zhang2019deep} and least square adversarial \cite{mao2017least} losses respectively, 
and $\lambda$'s control relative weights for corresponding losses.
$\mathcal{L}_{p}$ in Eq. \ref{eq:loss_s2} and Eq. \ref{eq:loss_s1} refer to the same loss function.
We use these four losses accordingly to encourage the merging-refinement subnet $\mathcal{M}$ to produce: (i) accurate merging and better refinement, (ii) perceptually plausible output, (iii) spatially smooth output, and (iv) realistic output.

\begin{table}[t]
    \centering
    \small
    \begin{tabular}{l|ccc}
        \noalign{\hrule height 0.3mm}
        Method & PSNR${\uparrow}$ & SSIM${\uparrow}$ & LPIPS${\downarrow}$\\
        \hline
        ExColTran \cite{yin2021yes} + OPR-R & 19.5796 & 0.7885 & 0.2563\\
        ReHistoGAN \cite{afifi2021histogan} + OPR-R & \underline{20.0458} & \textbf{0.7987} & \underline{0.2109}\\
       MAST \cite{huo2021manifold} + OPR-R & 19.0148 & 0.7853 & 0.2270\\
       PCAPST \cite{chiu2022pca} + OPR-R & 19.1731 & 0.7908 & 0.2197\\
       \hline
        Ours & \textbf{21.2212} & \underline{0.7919} & \textbf{0.2027}\\
        \noalign{\hrule height 0.3mm}
    \end{tabular}
    \vspace{-0.7em}
    \caption{Quantitative results of modernization on synthetic dataset.}
    \label{tab:quantitative_synthetic_data}
    \vspace{-1em}
\end{table}

\begin{table}[t]
    \centering
    \small
    \begin{tabular}{l|cc}
        \noalign{\hrule height 0.3mm}
        Method & NIQE${\downarrow}$ & BRISQUE${\downarrow}$\\
        \hline
        OPR \cite{wan2020bringing} & 4.8705 & 21.4588\\
        ExColTran \cite{yin2021yes} + OPR & 4.9415 & 18.8971\\
        ReHistoGAN \cite{afifi2021histogan} + OPR & 4.8051 & 26.2557\\
        MAST \cite{huo2021manifold} + OPR & 4.8111 & 18.9555\\
        PCAPST \cite{chiu2022pca} + OPR & 4.7094 & 18.9860 \\
        Ours - Single & \underline{3.4737} & \underline{15.5152}\\
        \hline
        Ours - Multiple & \textbf{3.4487} & \textbf{15.4180}\\
        \noalign{\hrule height 0.3mm}
    \end{tabular}
    \vspace{-0.7em}
    \caption{Quantitative results of modernization on real old photos.}
    \label{tab:quantitative_real_old_photos}
    \vspace{-1.1em}
\end{table}

\section{Experiments}

\subsection{Experimental Settings}
\label{subsec:experiment_settings}

\noindent
\textbf{Training details.}\quad
We use our proposed synthetic data generation scheme with the aforementioned multi-stage training strategy to train the network: (i) We train our PST network for five epochs; (ii)
Then, we train our single stylization subnet $\mathcal{S}$ based on $\mathcal{L}^{Stage2}$ in Eq. \ref{eq:loss_s1} for two epochs, not to be overfitted for synthetic data, while freezing our PST network, where we set $\lambda_{ML}=1$, $\lambda_{p}=1$, and $\lambda_{CX}=1$; (iii)
Finally, we train our merging-refinement subnet $\mathcal{M}$ for three epochs which is sufficient while freezing both $\mathcal{S}$ and our PST network.
The loss function to train $\mathcal{M}$ is $\mathcal{L}^{Stage3}$ in Eq. \ref{eq:loss_s2}, where we set $\lambda_{L1}=2$, $\lambda_{p}=1$, $\lambda_{sm}=3$, and $\lambda_{adv}=0.2$.
For all of the training, we use an ADAM optimizer \cite{kingma2014adam} with a learning rate of $1e-4$ and batch size of 1 to optimize our network and discriminator (PatchGAN discriminator \cite{isola2017image}).
In addition, we apply a linear learning decay in the last epoch of the $\mathcal{M}$ training.

\noindent
\textbf{Baselines.}\quad 
Our work can be seen as handling a joint task of stylization and enhancement by using multiple references for old photo modernization. 
Since there are no baselines in reference-based old photo modernization, we compare to sequential models consisting of stylization then enhancement, which can perform the same task.
Reversing the order (enhancement and then stylization), results in worse outcomes.
For stylization, we employ four state-of-the-art (SOTA) methods as baselines: (i) exemplar-based colorization: transformer-based method (ExColTran \cite{yin2021yes}); (ii) recolorization: recolorization using color-controlled GAN (ReHistoGAN \cite{afifi2021histogan}); photorealistic style transfer (PST): semantic PST (MAST \cite{huo2021manifold}) and PCA-based knowledge distillation PST (PCAPST \cite{chiu2022pca}).
Meanwhile, for enhancement, we employ SOTA no-reference-based old photo restoration method (OPR \cite{wan2020bringing}), which can perform similar enhancement as ours.
For the stylization baselines, we use their pre-trained models in the evaluation since they cannot be retrained with our synthetic data due to their different training strategies.
However, note that these pre-trained models have already been trained to achieve the same goal of changing the overall look of input images based on the given reference image.
Meanwhile, for enhancement, we use the pre-trained OPR model for real old photo evaluation, denoted as OPR, since it achieves better performance on real old photos, and retrain the OPR using our synthetic data and CHD training set for synthetic data evaluation, denoted as OPR-R.
Since the four baselines can only utilize a single reference, we average the results of using different sets of references in quantitative evaluation, and randomly select one of two references for each input to the baseline networks in qualitative evaluation and user study.

\begin{figure*}[ht]
\centering
\includegraphics[width=\textwidth]{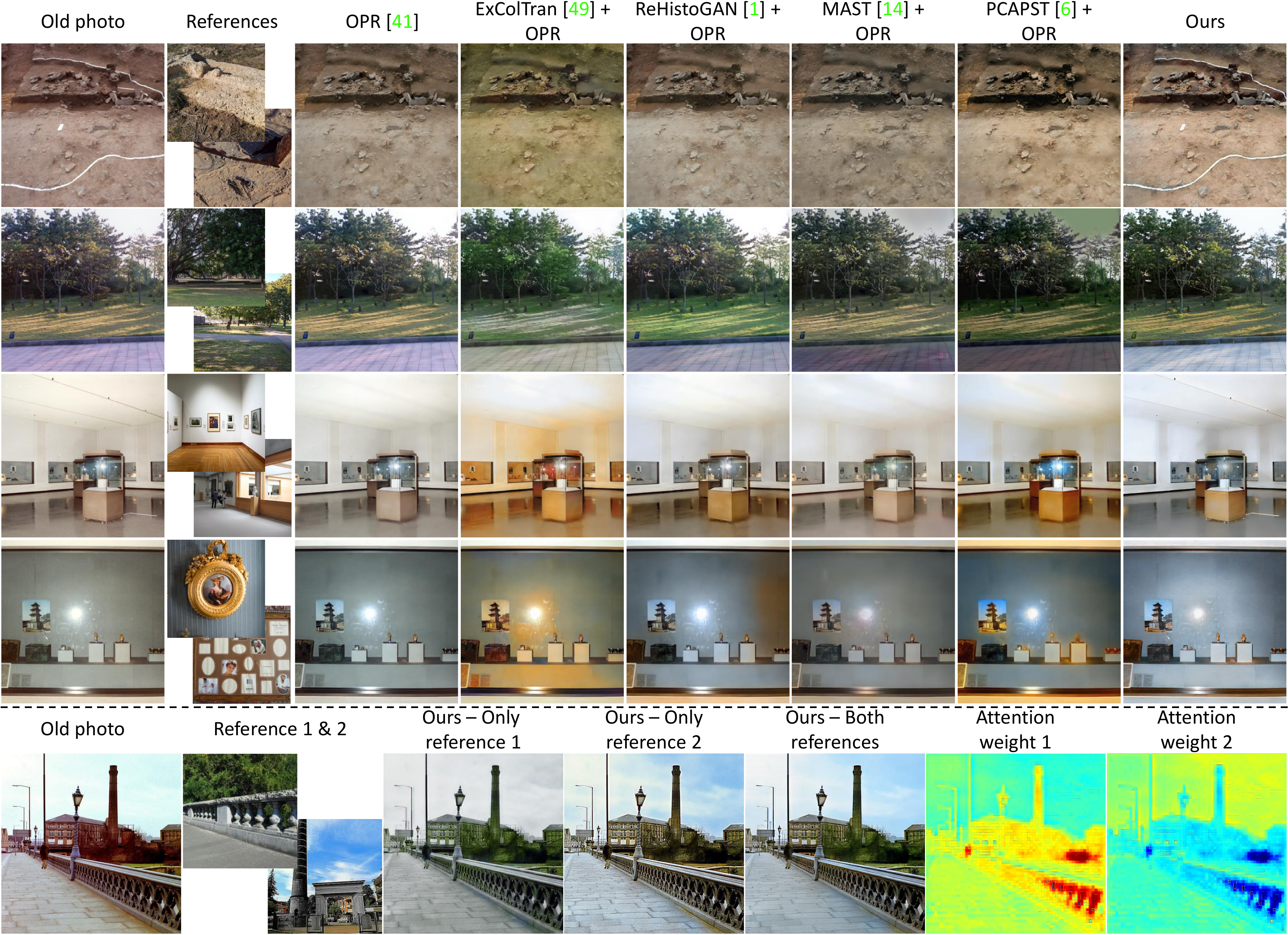}
\vspace{-2em}
\caption{\textbf{Top}: qualitative results of modernization on real outdoor and indoor old photos. The baselines use top-left reference as their reference. \textbf{Bottom}: Attention weight visualization, blue (lowest)-red (highest) color coded. Our method can select appropriate styles from multiple references depending on the availability of similar objects to achieve better modernization. (Zoom in for a better view)}
\label{fig:qualitative_result}
\vspace{-1em}
\end{figure*}

\begin{table}[t]
    \centering
    \small
    \begin{adjustbox}{max width=\linewidth}
        \begin{tabular}{l|ccccc}
            \noalign{\hrule height 0.3mm}
            Method & Top 1 & Top 2 & Top 3 & Top 4 & Top 5\\
            \hline
            OPR \cite{wan2020bringing} & \underline{17.44} & \underline{39.83} & 57.05 & 70.90 & 87.22\\
            ExColTran \cite{yin2021yes} + OPR & 1.62 & 5.13 & 10.77 & 24.87 & 47.27\\
            ReHistoGAN \cite{afifi2021histogan} + OPR & 7.91 & 32.27 & \underline{61.84} & \underline{83.80} & \underline{96.92}\\
            MAST \cite{huo2021manifold} + OPR & 5.68 & 21.62 & 41.92 & 66.33 & 86.20\\
            PCAPST \cite{chiu2022pca} + OPR & 10.98 & 28.50 & 44.87 & 61.97 & 84.66\\
            \hline
            Ours & \textbf{56.37} & \textbf{72.69} & \textbf{83.55} & \textbf{92.14} & \textbf{97.74}\\
            \noalign{\hrule height 0.3mm}
        \end{tabular}
    \end{adjustbox}
    \vspace{-0.7em}
    \caption{User study results. The percentage of user selection is shown.}
    \label{tab:user_study}
    \vspace{-1.1em}
\end{table}

\noindent
\textbf{Evaluation metrics.}\quad
We evaluate all the methods on synthetically degraded and real old photos (CHD testing set).
In synthetic degraded photos evaluation, we employ: 1) peak signal-to-noise ratio (PSNR) and structural similarity index (SSIM) to measure the pixel-level discrepancy between output and ground truth, 2) learned perceptual image patch similarity (LPIPS) \cite{zhang2018unreasonable} to measure the perceptual quality of the output.
For evaluation on real old photos, we employ no-reference image quality assessment metrics such as NIQE \cite{mittal2012making} and BRISQUE \cite{mittal2012no}, similar to \cite{wan2020bringing, wan2022bringing, luo2021time}, since the modernization ground truth photos do not exist.

\subsection{Experimental Results}

\noindent
\textbf{Quantitative comparison.}\quad
We evaluate our method and baselines on a synthetic dataset and real-world old photos.
In synthetic dataset evaluation, we evaluate all methods, including ours, in a single-reference-based scenario since the baselines cannot utilize more than one reference by using ADE20K validation set\cite{zhou2017scene} that includes semantic segmentation masks.
Specifically, we generate 1,000 evaluation pairs, each consisting of a synthetic degraded photo and a reference image, by randomly degrading half set of the semantic regions using our synthetic data generation scheme, e.g., only the surfer in Fig. \ref{fig:synthetic_data_creation}.
Table. \ref{tab:quantitative_synthetic_data} shows our method achieves the best PSNR and LPIPS score, which means our method can effectively utilize the references to jointly stylize and enhance the synthetically degraded images, thus generating an output similar to the ground truth both in the pixel and semantic levels.
In terms of SSIM, we achieve the second-best compared to recolorization (ReHistoGAN \cite{afifi2021histogan} + OPR-R) since our method can change the other aspects besides color, such as texture and luminance, which may result in a lower SSIM score.
Interestingly, compared to other PST baselines, especially MAST \cite{huo2021manifold}, which is designed to perform semantic style transfer, our method achieves the most accurate stylization (PSNR and LPIPS) while still preserving the structure (SSIM), which are two important aspects in PST.
A similar observation can be seen for real old photos evaluation shown in Table. \ref{tab:quantitative_real_old_photos}, where our method outperforms other baselines significantly using a single reference and further improves the performance by using multiple references.

\noindent
\textbf{Qualitative comparison.}\quad 
As shown in Fig. \ref{fig:qualitative_result}, no-reference OPR \cite{wan2020bringing} can restore both structured (SD) and unstructured degradations (UD).
However, SD restoration cannot generalize well to real old photos since it significantly degrades the important regions of the original photos.
In addition, it fails to modernize some old photos because the overall styles still remain similar to the original old photos.
ExColTran \cite{yin2021yes}, which can only use luminance information for semantic matching, fails to locally change the color of old photos, thus producing unnatural results.
Meanwhile, ReHistoGAN \cite{afifi2021histogan} can better recolorize old photos, producing more modern-looking images than only OPR.
Compared to other PST methods combined with OPR, and other baselines, our method achieves better local and global PST and yields enhancement, consequently achieving better modernization.
Moreover, our method can utilize multiple references better in all examples, e.g., the second row of Fig. \ref{fig:qualitative_result}, where styles of tree leaves and road come from the first and second references, respectively.
Meanwhile, the fourth row shows the generalization of our method, which can handle unrelated references well.
The bottom part under the dotted line in Fig. \ref{fig:qualitative_result} shows the visualization of spatial attention weights, where our method can select appropriate styles for each semantic object in an old photo from multiple references to achieve better modernization, e.g., the bridge and sky style in the first and second reference respectively.
All in all, our method can modernize old photos better than the baselines by leveraging multiple modern photo references, even though it has not been trained with any old photos.
These results also show that restoring the degradation of old photos cannot guarantee the outputs to look modernized, but changing their styles can contribute to the \textit{modern} look more than restoring the degradation.

\noindent
\textbf{User study.}\quad 
We conduct a user study to compare the modernization results of our method with those of the baselines. 
Specifically, we select 130 photos from the CHD testing set and ask 18 users to rank the modernization results.
As shown in Table. \ref{tab:user_study}, our method outperforms other baselines with 56.37\% chance selected as the best method.

\subsection{Analysis}
\begin{figure}[t]
\centering
\includegraphics[width=\columnwidth]{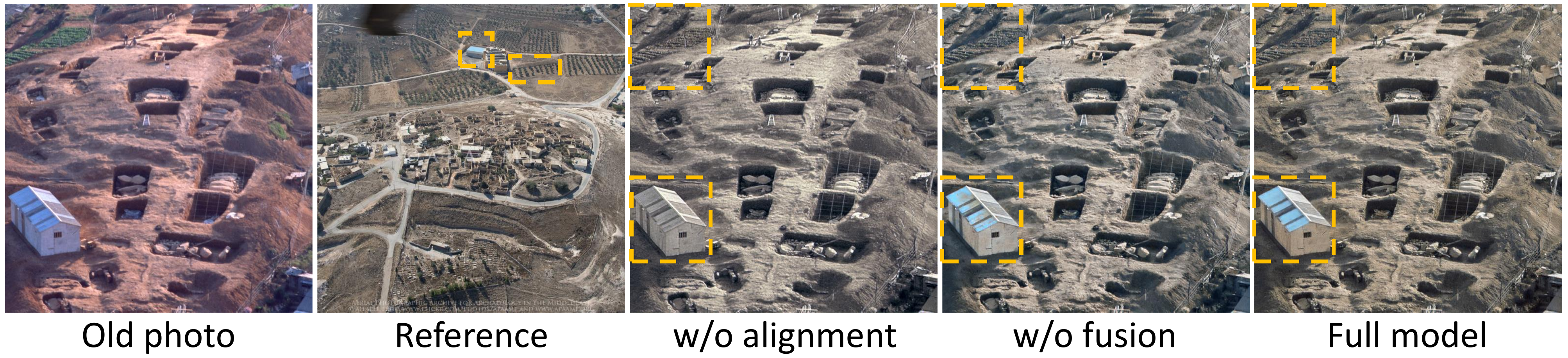}
\vspace{-1.9em}
\caption{Ablation study on the single stylization subnet.}
\label{fig:results_ablat_s}
\vspace{-1em}
\end{figure}

\begin{figure}[t]
\centering
\includegraphics[width=\columnwidth]{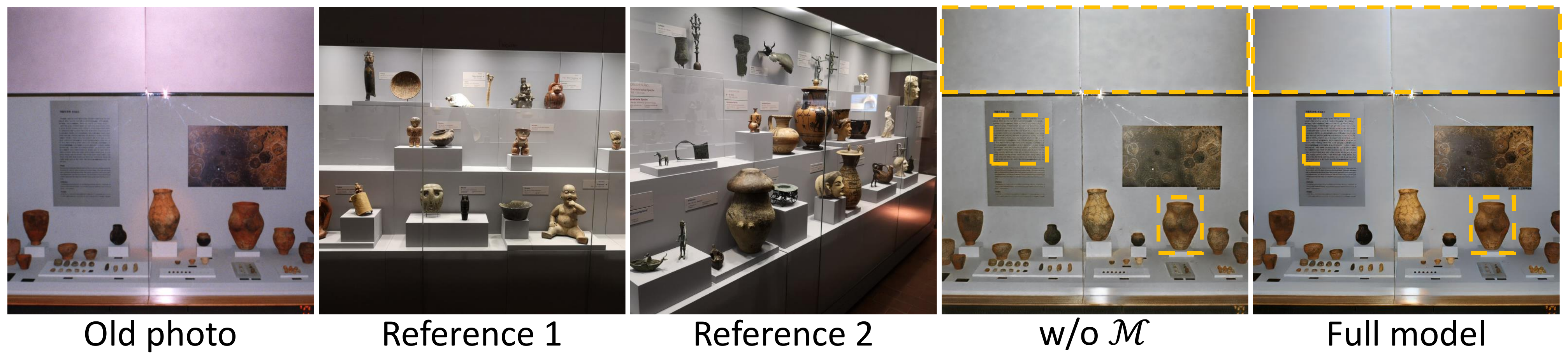}
\vspace{-1.9em}
\caption{Ablation study on the merging-refinement subnet.}
\label{fig:results_ablat_mr}
\vspace{-1.1em}
\end{figure}

\noindent
\textbf{Ablation study on the single stylization subnet.} 
We analyze the contribution of each module in the single stylization subnet $\mathcal{S}$.
As shown in Fig. \ref{fig:results_ablat_s}, the subnet fails to accurately transfer the local styles of objects, e.g., the styles of the blue building and grass, when the alignment module is removed. 
Even though the subnet can perform better local style transfer of building and grass regions with the alignment module, the stylization results are not smooth, which may produce unnatural results. 
Thus, adding a fusion module that merges global and local styles can produce smoother stylization locally and globally. 

\noindent
\textbf{Ablation study on the merging-refinement subnet.} 
To evaluate the contribution of the merging-refinement subnet $\mathcal{M}$, we change this subnet to a simple concatenation between multiple stylization features and feed the concatenated features into several convolution blocks (denoted as w/o $\mathcal{M}$).
As shown in Fig. \ref{fig:results_ablat_mr}, without $\mathcal{M}$, the network cannot select appropriate styles from different references and fail to enhance the results.
In addition, retraining the network is required to use a different number of references between inference and training.
With our $\mathcal{M}$, we can adaptively choose the number of references without retraining. 
Results of other ablation studies and modernization using more than two references are in \textit{Supplementary Material}.

\begin{figure}[t]
\centering
\includegraphics[width=\columnwidth]{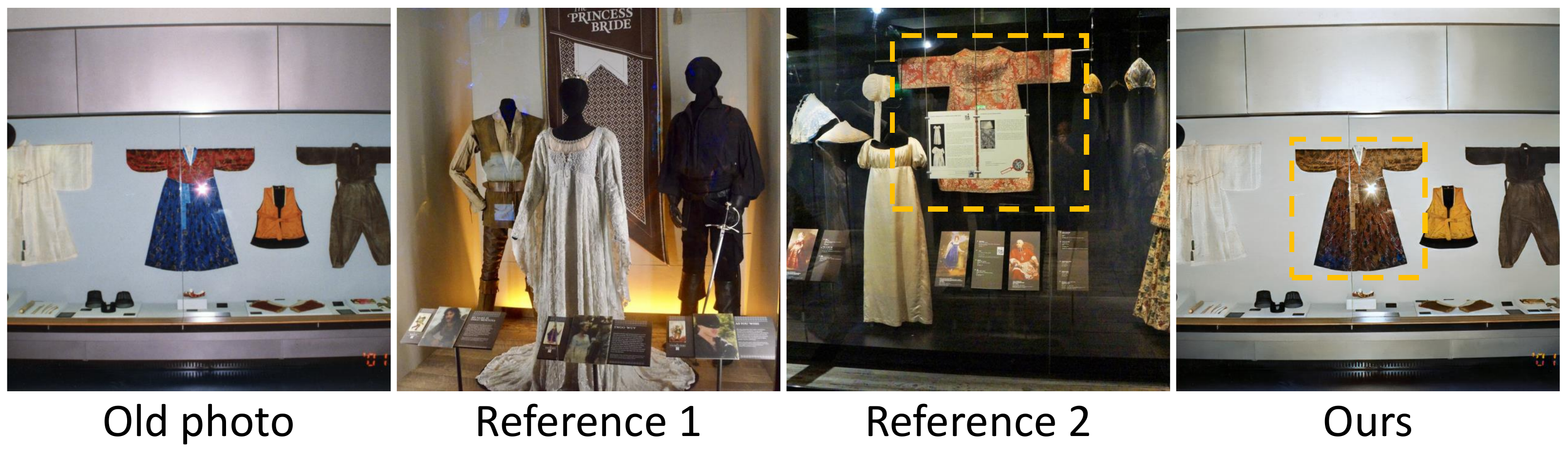}
\vspace{-1.8em}
\caption{Limitation of our method.}
\label{fig:limitation}
\vspace{-1.1em}
\end{figure}

\noindent
\textbf{Limitation.}
The limitation of our work mainly comes from the selection of references.
As shown in Fig. \ref{fig:limitation}, our method may produce unsatisfying modernization when a related object in the references has a style that does not enhance the old photo.
However, finding references that contain a similar local object with a modern style in an automated way is highly challenging using existing image retrieval methods.
Moreover, using VGG feature space matching similar to \cite{he2018deep} fails to produce semantically similar references due to the domain gap between old and modern photos.

\section{Conclusion}
\label{sec:conclusion}
In this paper, we first proposed old photo modernization by using multiple references.
In order to perform modernization, we proposed MROPM, which performs old photo stylization using multiple references via photorealistic style transfer and enhancement in one unified framework.
Thanks to the generalization of PST and our synthetic data generation scheme, our work outperforms baselines for real-world old photos, even without using any old photos during the training.
Furthermore, we analyze that our method can select appropriate styles from multiple references, further improving the modernization performance.
Also, we propose an old color photos dataset CHD consisting of natural indoor and outdoor scenes to spur future research in the domain.

\small{
\smallskip\noindent
\textbf{Acknowledgment.}\quad
This research was supported by Culture, Sports and Tourism R\&D Program through the Korea Creative Content Agency grant funded by the Ministry of Culture, Sports and Tourism in 2020 (Project Name: CHIC, Project Number: R2020040045, Contribution Rate: 100\%). We would like to thank Gimhae, Jeju, and National Museum of Korea for the old photos.
}

{\small
\bibliographystyle{ieee_fullname}
\bibliography{main}

\begin{thebibliography}{10}\itemsep=-1pt

\bibitem{afifi2021histogan}
Mahmoud Afifi, Marcus~A Brubaker, and Michael~S Brown.
\newblock Histogan: Controlling colors of gan-generated and real images via
  color histograms.
\newblock In {\em Proceedings of the IEEE/CVF conference on computer vision and
  pattern recognition}, pages 7941--7950, 2021.

\bibitem{an2020ultrafast}
Jie An, Haoyi Xiong, Jun Huan, and Jiebo Luo.
\newblock Ultrafast photorealistic style transfer via neural architecture
  search.
\newblock In {\em Proceedings of the AAAI Conference on Artificial
  Intelligence}, volume~34, pages 10443--10450, 2020.

\bibitem{brock2018large}
Andrew Brock, Jeff Donahue, and Karen Simonyan.
\newblock Large scale gan training for high fidelity natural image synthesis.
\newblock In {\em International Conference on Learning Representations}, 2018.

\bibitem{burt1987laplacian}
Peter~J Burt and Edward~H Adelson.
\newblock The laplacian pyramid as a compact image code.
\newblock In {\em Readings in computer vision}, pages 671--679. Elsevier, 1987.

\bibitem{caesar2018coco}
Holger Caesar, Jasper Uijlings, and Vittorio Ferrari.
\newblock Coco-stuff: Thing and stuff classes in context.
\newblock In {\em Proceedings of the IEEE conference on computer vision and
  pattern recognition}, pages 1209--1218, 2018.

\bibitem{chen2022vision}
Zhe Chen, Yuchen Duan, Wenhai Wang, Junjun He, Tong Lu, Jifeng Dai, and Yu
  Qiao.
\newblock Vision transformer adapter for dense predictions.
\newblock {\em arXiv preprint arXiv:2205.08534}, 2022.

\bibitem{chiu2022pca}
Tai-Yin Chiu and Danna Gurari.
\newblock Pca-based knowledge distillation towards lightweight and
  content-style balanced photorealistic style transfer models.
\newblock In {\em Proceedings of the IEEE/CVF Conference on Computer Vision and
  Pattern Recognition}, pages 7844--7853, 2022.

\bibitem{chiu2022photowct2}
Tai-Yin Chiu and Danna Gurari.
\newblock Photowct2: Compact autoencoder for photorealistic style transfer
  resulting from blockwise training and skip connections of high-frequency
  residuals.
\newblock In {\em Proceedings of the IEEE/CVF Winter Conference on Applications
  of Computer Vision}, pages 2868--2877, 2022.

\bibitem{giakoumis2005digital}
Ioannis Giakoumis, Nikos Nikolaidis, and Ioannis Pitas.
\newblock Digital image processing techniques for the detection and removal of
  cracks in digitized paintings.
\newblock {\em TIP}, 15(1):178--188, 2005.

\bibitem{he2016deep}
Kaiming He, Xiangyu Zhang, Shaoqing Ren, and Jian Sun.
\newblock Deep residual learning for image recognition.
\newblock In {\em Proceedings of the IEEE conference on computer vision and
  pattern recognition}, pages 770--778, 2016.

\bibitem{he2018deep}
Mingming He, Dongdong Chen, Jing Liao, Pedro~V Sander, and Lu Yuan.
\newblock Deep exemplar-based colorization.
\newblock {\em ACM Transactions on Graphics (TOG)}, 37(4):1--16, 2018.

\bibitem{he2019progressive}
Mingming He, Jing Liao, Dongdong Chen, Lu Yuan, and Pedro~V Sander.
\newblock Progressive color transfer with dense semantic correspondences.
\newblock {\em ACM Transactions on Graphics (TOG)}, 38(2):1--18, 2019.

\bibitem{hu2020aesthetic}
Zhiyuan Hu, Jia Jia, Bei Liu, Yaohua Bu, and Jianlong Fu.
\newblock Aesthetic-aware image style transfer.
\newblock In {\em Proceedings of the 28th ACM International Conference on
  Multimedia}, pages 3320--3329, 2020.

\bibitem{huang2017arbitrary}
Xun Huang and Serge Belongie.
\newblock Arbitrary style transfer in real-time with adaptive instance
  normalization.
\newblock In {\em Proceedings of the IEEE international conference on computer
  vision}, pages 1501--1510, 2017.

\bibitem{huang2019style}
Zixuan Huang, Jinghuai Zhang, and Jing Liao.
\newblock Style mixer: Semantic-aware multi-style transfer network.
\newblock In {\em Computer Graphics Forum}, volume~38, pages 469--480. Wiley
  Online Library, 2019.

\bibitem{huo2021manifold}
Jing Huo, Shiyin Jin, Wenbin Li, Jing Wu, Yu-Kun Lai, Yinghuan Shi, and Yang
  Gao.
\newblock Manifold alignment for semantically aligned style transfer.
\newblock In {\em Proceedings of the IEEE/CVF International Conference on
  Computer Vision}, pages 14861--14869, 2021.

\bibitem{isola2017image}
Phillip Isola, Jun-Yan Zhu, Tinghui Zhou, and Alexei~A Efros.
\newblock Image-to-image translation with conditional adversarial networks.
\newblock In {\em Proceedings of the IEEE conference on computer vision and
  pattern recognition}, pages 1125--1134, 2017.

\bibitem{johnson2016perceptual}
Justin Johnson, Alexandre Alahi, and Li Fei-Fei.
\newblock Perceptual losses for real-time style transfer and super-resolution.
\newblock In {\em European conference on computer vision}, pages 694--711.
  Springer, 2016.

\bibitem{karras2020analyzing}
Tero Karras, Samuli Laine, Miika Aittala, Janne Hellsten, Jaakko Lehtinen, and
  Timo Aila.
\newblock Analyzing and improving the image quality of stylegan.
\newblock In {\em Proceedings of the IEEE/CVF conference on computer vision and
  pattern recognition}, pages 8110--8119, 2020.

\bibitem{kingma2014adam}
Diederik~P Kingma and Jimmy Ba.
\newblock Adam: A method for stochastic optimization.
\newblock {\em arXiv preprint arXiv:1412.6980}, 2014.

\bibitem{lee2020deep}
Junyong Lee, Hyeongseok Son, Gunhee Lee, Jonghyeop Lee, Sunghyun Cho, and
  Seungyong Lee.
\newblock Deep color transfer using histogram analogy.
\newblock {\em The Visual Computer}, 36(10):2129--2143, 2020.

\bibitem{li2019learning}
Xueting Li, Sifei Liu, Jan Kautz, and Ming-Hsuan Yang.
\newblock Learning linear transformations for fast image and video style
  transfer.
\newblock In {\em Proceedings of the IEEE/CVF Conference on Computer Vision and
  Pattern Recognition}, pages 3809--3817, 2019.

\bibitem{li2017universal}
Yijun Li, Chen Fang, Jimei Yang, Zhaowen Wang, Xin Lu, and Ming-Hsuan Yang.
\newblock Universal style transfer via feature transforms.
\newblock {\em Advances in neural information processing systems}, 30, 2017.

\bibitem{li2018closed}
Yijun Li, Ming-Yu Liu, Xueting Li, Ming-Hsuan Yang, and Jan Kautz.
\newblock A closed-form solution to photorealistic image stylization.
\newblock In {\em Proceedings of the European Conference on Computer Vision
  (ECCV)}, pages 453--468, 2018.

\bibitem{liao2017visual}
Jing Liao, Yuan Yao, Lu Yuan, Gang Hua, and Sing~Bing Kang.
\newblock Visual attribute transfer through deep image analogy.
\newblock {\em ACM Transactions on Graphics (TOG)}, 36(4):1--15, 2017.

\bibitem{liu2021cg}
Jixin Liu, Rui Chen, Shipeng An, and Heng Zhang.
\newblock Cg-gan: Class-attribute guided generative adversarial network for old
  photo restoration.
\newblock In {\em Proceedings of the 29th ACM International Conference on
  Multimedia}, pages 5391--5399, 2021.

\bibitem{lu2020gray2colornet}
Peng Lu, Jinbei Yu, Xujun Peng, Zhaoran Zhao, and Xiaojie Wang.
\newblock Gray2colornet: Transfer more colors from reference image.
\newblock In {\em Proceedings of the 28th ACM International Conference on
  Multimedia}, pages 3210--3218, 2020.

\bibitem{luan2017deep}
Fujun Luan, Sylvain Paris, Eli Shechtman, and Kavita Bala.
\newblock Deep photo style transfer.
\newblock In {\em Proceedings of the IEEE conference on computer vision and
  pattern recognition}, pages 4990--4998, 2017.

\bibitem{luo2021time}
Xuan Luo, Xuaner Zhang, Paul Yoo, Ricardo Martin-Brualla, Jason Lawrence, and
  Steven~M Seitz.
\newblock Time-travel rephotography.
\newblock {\em ACM Transactions on Graphics (TOG)}, 40(6):1--12, 2021.

\bibitem{machidon2018digital}
Octavian-Mihai Machidon and Mihai Ivanovici.
\newblock Digital color restoration for the preservation of reversal film
  heritage.
\newblock {\em Journal of Cultural Heritage}, 33:181--190, 2018.

\bibitem{mao2017least}
Xudong Mao, Qing Li, Haoran Xie, Raymond~YK Lau, Zhen Wang, and Stephen
  Paul~Smolley.
\newblock Least squares generative adversarial networks.
\newblock In {\em Proceedings of the IEEE international conference on computer
  vision}, pages 2794--2802, 2017.

\bibitem{mechrez2017photorealistic}
Roey Mechrez, Eli Shechtman, and Lihi Zelnik-Manor.
\newblock Photorealistic style transfer with screened poisson equation.
\newblock {\em arXiv preprint arXiv:1709.09828}, 2017.

\bibitem{mechrez2018contextual}
Roey Mechrez, Itamar Talmi, and Lihi Zelnik-Manor.
\newblock The contextual loss for image transformation with non-aligned data.
\newblock In {\em Proceedings of the European conference on computer vision
  (ECCV)}, pages 768--783, 2018.

\bibitem{mittal2012no}
Anish Mittal, Anush~Krishna Moorthy, and Alan~Conrad Bovik.
\newblock No-reference image quality assessment in the spatial domain.
\newblock {\em IEEE Transactions on image processing}, 21(12):4695--4708, 2012.

\bibitem{mittal2012making}
Anish Mittal, Rajiv Soundararajan, and Alan~C Bovik.
\newblock Making a “completely blind” image quality analyzer.
\newblock {\em IEEE Signal processing letters}, 20(3):209--212, 2012.

\bibitem{qu2021non}
Ying Qu, Zhenzhou Shao, and Hairong Qi.
\newblock Non-local representation based mutual affine-transfer network for
  photorealistic stylization.
\newblock {\em IEEE Transactions on Pattern Analysis and Machine Intelligence},
  44(10):7046--7061, 2021.

\bibitem{ronneberger2015u}
Olaf Ronneberger, Philipp Fischer, and Thomas Brox.
\newblock U-net: Convolutional networks for biomedical image segmentation.
\newblock In {\em International Conference on Medical image computing and
  computer-assisted intervention}, pages 234--241. Springer, 2015.

\bibitem{simonyan2014very}
Karen Simonyan and Andrew Zisserman.
\newblock Very deep convolutional networks for large-scale image recognition.
\newblock {\em arXiv preprint arXiv:1409.1556}, 2014.

\bibitem{singh2020assessing}
Aditya Singh, Alessandro Bay, and Andrea Mirabile.
\newblock Assessing the importance of colours for cnns in object recognition.
\newblock In {\em NeurIPS 2020 Workshop SVRHM}, 2020.

\bibitem{stanco2003towards}
F Stanco, Giovanni Ramponi, and A De~Polo.
\newblock Towards the automated restoration of old photographic prints: a
  survey.
\newblock In {\em The IEEE Region 8 EUROCON 2003. Computer as a Tool.},
  volume~2, pages 370--374. IEEE, 2003.

\bibitem{su2020instance}
Jheng-Wei Su, Hung-Kuo Chu, and Jia-Bin Huang.
\newblock Instance-aware image colorization.
\newblock In {\em Proceedings of the IEEE/CVF Conference on Computer Vision and
  Pattern Recognition}, pages 7968--7977, 2020.

\bibitem{ulyanov2016instance}
Dmitry Ulyanov, Andrea Vedaldi, and Victor Lempitsky.
\newblock Instance normalization: The missing ingredient for fast stylization.
\newblock {\em arXiv preprint arXiv:1607.08022}, 2016.

\bibitem{wan2022bringing}
Ziyu Wan, Bo Zhang, Dongdong Chen, and Jing Liao.
\newblock Bringing old films back to life.
\newblock In {\em Proceedings of the IEEE/CVF Conference on Computer Vision and
  Pattern Recognition}, pages 17694--17703, 2022.

\bibitem{wan2020bringing}
Ziyu Wan, Bo Zhang, Dongdong Chen, Pan Zhang, Dong Chen, Jing Liao, and Fang
  Wen.
\newblock Bringing old photos back to life.
\newblock In {\em proceedings of the IEEE/CVF conference on computer vision and
  pattern recognition}, pages 2747--2757, 2020.

\bibitem{wang2018non}
Xiaolong Wang, Ross Girshick, Abhinav Gupta, and Kaiming He.
\newblock Non-local neural networks.
\newblock In {\em Proceedings of the IEEE conference on computer vision and
  pattern recognition}, pages 7794--7803, 2018.

\bibitem{whitt2014beginning}
Phillip Whitt.
\newblock {\em Beginning photo retouching and restoration using GIMP}.
\newblock Springer, 2014.

\bibitem{woo2018cbam}
Sanghyun Woo, Jongchan Park, Joon-Young Lee, and In~So Kweon.
\newblock Cbam: Convolutional block attention module.
\newblock In {\em Proceedings of the European conference on computer vision
  (ECCV)}, pages 3--19, 2018.

\bibitem{wu2021towards}
Yanze Wu, Xintao Wang, Yu Li, Honglun Zhang, Xun Zhao, and Ying Shan.
\newblock Towards vivid and diverse image colorization with generative color
  prior.
\newblock In {\em Proceedings of the IEEE/CVF International Conference on
  Computer Vision}, pages 14377--14386, 2021.

\bibitem{xia2020joint}
Xide Xia, Meng Zhang, Tianfan Xue, Zheng Sun, Hui Fang, Brian Kulis, and Jiawen
  Chen.
\newblock Joint bilateral learning for real-time universal photorealistic style
  transfer.
\newblock In {\em European Conference on Computer Vision}, pages 327--342.
  Springer, 2020.

\bibitem{xu2022pik}
Runsheng Xu, Zhengzhong Tu, Yuanqi Du, Xiaoyu Dong, Jinlong Li, Zibo Meng,
  Jiaqi Ma, Alan Bovik, and Hongkai Yu.
\newblock Pik-fix: Restoring and colorizing old photo.
\newblock {\em arXiv preprint arXiv:2205.01902}, 2022.

\bibitem{xu2020stylization}
Zhongyou Xu, Tingting Wang, Faming Fang, Yun Sheng, and Guixu Zhang.
\newblock Stylization-based architecture for fast deep exemplar colorization.
\newblock In {\em Proceedings of the IEEE/CVF Conference on Computer Vision and
  Pattern Recognition}, pages 9363--9372, 2020.

\bibitem{yin2021yes}
Wang Yin, Peng Lu, Zhaoran Zhao, and Xujun Peng.
\newblock Yes," attention is all you need", for exemplar based colorization.
\newblock In {\em Proceedings of the 29th ACM International Conference on
  Multimedia}, pages 2243--2251, 2021.

\bibitem{yoo2019photorealistic}
Jaejun Yoo, Youngjung Uh, Sanghyuk Chun, Byeongkyu Kang, and Jung-Woo Ha.
\newblock Photorealistic style transfer via wavelet transforms.
\newblock In {\em Proceedings of the IEEE/CVF International Conference on
  Computer Vision}, pages 9036--9045, 2019.

\bibitem{zhang2019deep}
Bo Zhang, Mingming He, Jing Liao, Pedro~V Sander, Lu Yuan, Amine Bermak, and
  Dong Chen.
\newblock Deep exemplar-based video colorization.
\newblock In {\em Proceedings of the IEEE/CVF Conference on Computer Vision and
  Pattern Recognition}, pages 8052--8061, 2019.

\bibitem{zhang2018unreasonable}
Richard Zhang, Phillip Isola, Alexei~A Efros, Eli Shechtman, and Oliver Wang.
\newblock The unreasonable effectiveness of deep features as a perceptual
  metric.
\newblock In {\em Proceedings of the IEEE conference on computer vision and
  pattern recognition}, pages 586--595, 2018.

\bibitem{zhou2017scene}
Bolei Zhou, Hang Zhao, Xavier Puig, Sanja Fidler, Adela Barriuso, and Antonio
  Torralba.
\newblock Scene parsing through ade20k dataset.
\newblock In {\em Proceedings of the IEEE conference on computer vision and
  pattern recognition}, pages 633--641, 2017.

\end{thebibliography}
}

\clearpage
\twocolumn[{%
\renewcommand\twocolumn[1][]{#1}%
\begin{center}
\centering
\textbf{\large Supplementary Material}
\end{center}%
\begin{center}
\centering
\captionsetup{type=figure}
\includegraphics[width=\textwidth]{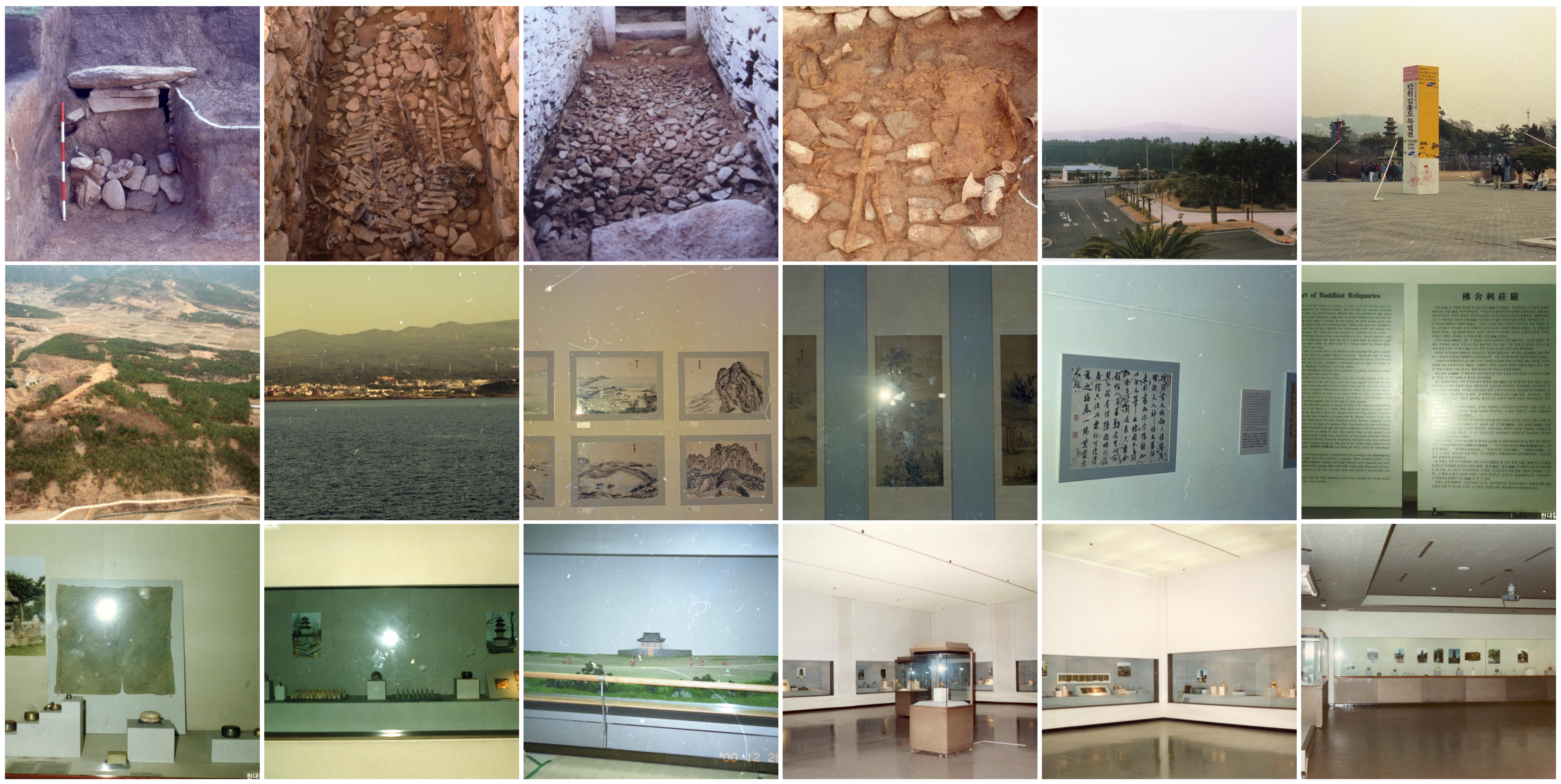}
\vspace{-1.8em}
\captionof{figure}{Diverse examples of outdoor and indoor scenes from our dataset. The images contain various kinds of degradations, especially color fading.}
\label{fig:1_1_dataset_examples}
\vspace{0.4em}
\end{center}%
}]

\section{Details of Proposed CHD Dataset}

\begin{figure*}[ht]
\centering
\includegraphics[width=0.8\textwidth]{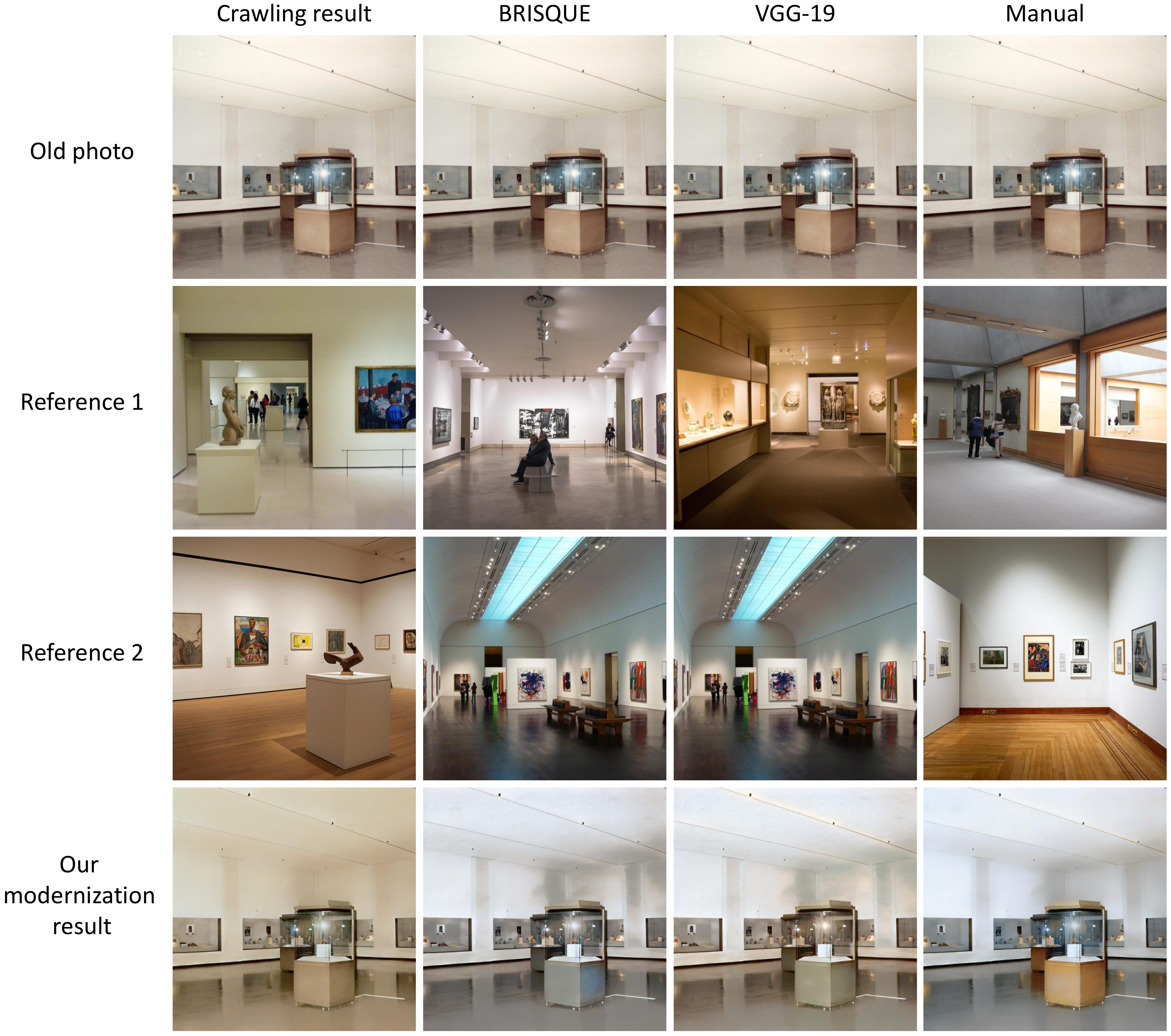}
\vspace{-0.5em}
\caption{Automatic references selection trial. We try to automate reference selection by using BRISQUE\cite{mittal2012no} and VGG-19\cite{simonyan2014very} feature similarity.}
\label{fig:1_3_automatic_selection}
\vspace{-1em}
\end{figure*}

\subsection{Details of Our CHD Dataset}

In order to curate our Cultural Heritage Dataset (CHD), we collect old photos produced in the 20th century.
Specifically, we collect these old photos in the form of reversal films or papers from three national museums in Korea, i.e., the National Museum of Korea, Gimhae National Museum, and Jeju National Museum.
After collecting old photos, we scan the photos in resolution varying from 4K to 8K.
The content of the photos is indoor and outdoor scenes of cultural heritage, such as special exhibitions and excavation ruins.
For the degradation, the photos contain a little scratch and crack degradations since they have been well preserved and stored carefully due to their important values.
However, they contain various degrees of unstructured degradations and color fading.
Fig. \ref{fig:1_1_dataset_examples} shows the diversity of indoor and outdoor scenes in our dataset with varying degrees of degradations such as blur, noise, scratch and crack, and color fading.
In addition to these degradations, our dataset also contains various real-world artifacts that may provide benefits for the community such as reflection, flash, etc.

After the collection, we filter out several old photos that contain sensitive information, e.g., front-facing faces, distinguished faces, and license plates.
In total, 644 old color photos are obtained through filtering, where 383, 147, and 114 photos are from the National Museum of Korea, Gimhae National Museum, and Jeju National Museum respectively.
Then, we randomly divide these 644 old photos into train and test sets with a proportion of 8:2.
Note that we also preserve the same ratio of images in the train and test set for each museum name.
In total, we obtain 514 photos for the train set and 130 photos for the test set.
The train set is used to train the old photo restoration baseline that needs to be trained using real old photos since it works by reducing the domain gap between real and synthetic old photos.
Meanwhile, our method does not use any old photos during the training since our method utilizes photorealistic style transfer that can work on any photo, including old photos.
Since the scanned photos have a resolution of 4K to 8K, we further preprocess the photos.
Specifically, we resize these photos to make the short side (width or height) have a resolution of 1024, then we center-crop the images, resulting in a resolution of $1024\times1024$.

Since our task is reference-based old photo modernization, we further collect photos as references by automatically crawling CC-Licensed images with similar contexts from an internet search using the crawling tool\footnote{\url{https://github.com/hardikvasa/google-images-download}} for the test set, where each old photo serves as the query of the search.
Approximately 100 reference photos for each old photo in the test set are obtained.
Then, we select one to two modern photos manually as the references for each old photo, which are then resized and center-cropped, similar to the resize and crop operation applied to the old photos resulting in a resolution of $1024\times1024$.
We tried to perform the selection automatically by selecting reference photos that have the largest cosine similarity in the VGG-19 \cite{simonyan2014very} feature space, using a similar idea to \cite{he2018deep, xu2022pik}, and the best BRISQUE \cite{mittal2012no} score.
However, Fig. \ref{fig:1_3_automatic_selection} shows that the automatic selection sometimes fails to obtain modern photo references with a modern style.
The references from the crawling result have a similar context to the old photo. 
However, the references have the characteristic of old photos, i.e., hazy and unsaturated colors.
This observation is similar to the automatic selection using the VGG-19 feature space, where the selection algorithm tends to select photos with similar old photos characteristics, e.g., sepia color.
Meanwhile, selecting references with the best BRISQUE score cannot obtain references with similar contexts, e.g., no similar objects for the showcase. 
Note that since we do not own any of the reference photos, we will only release the link for the reference images and the attribution in the dataset.

\begin{figure*}[ht]
\centering
\includegraphics[width=\textwidth]{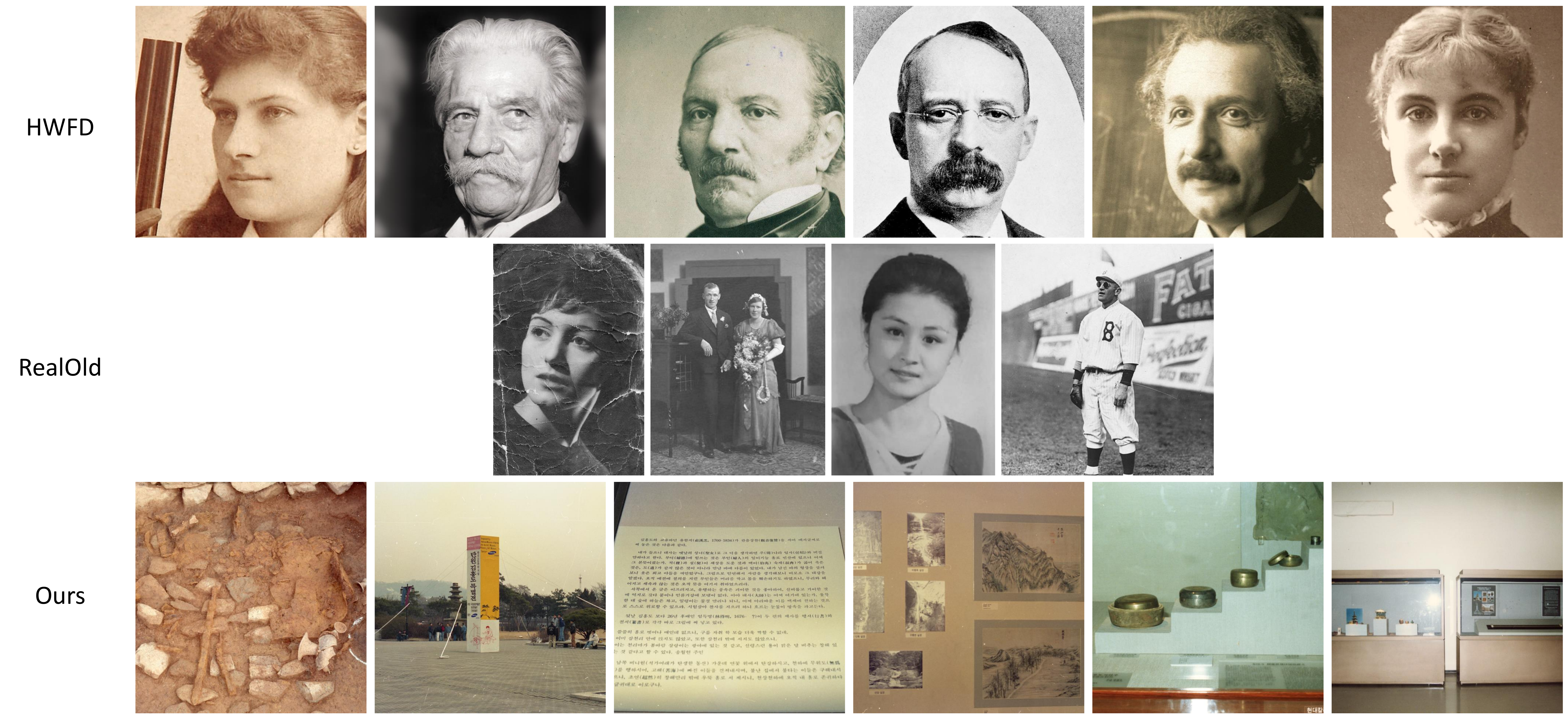}
\vspace{-2em}
\caption{Comparison between our CHD dataset and other datasets (HWFD \cite{luo2021time} and RealOld \cite{xu2022pik}). Our CHD dataset has the most complex and diverse scenes compared to other datasets. In addition, our dataset also contains unique color fading artifacts.}
\label{fig:1_2_dataset_comparison}
\vspace{-1em}
\end{figure*}

\begin{table}[t]
    \centering
    \small
    \begin{adjustbox}{max width=\linewidth}
        \begin{tabular}{l|ccc}
            \noalign{\hrule height 0.3mm}
             & HWFD \cite{luo2021time} & RealOld \cite{xu2022pik} & Ours \\
            \hline
            \begin{tabular}{@{}l@{}}Number \\ of images\end{tabular} & 224 & 200 & 644 \\
            \hline
            Era & \begin{tabular}{@{}c@{}}19-20th \\ century\end{tabular} & \begin{tabular}{@{}c@{}} - \end{tabular} & \begin{tabular}{@{}c@{}}20th \\ century\end{tabular}\\
            \hline
            Content type & \begin{tabular}{@{}c@{}} Face \end{tabular} & \begin{tabular}{@{}c@{}} Portrait \end{tabular} & \begin{tabular}{@{}c@{}}Indoor \& outdoor \\ natural scenes\end{tabular} \\
            \hline
            Color space & Greyscale & Greyscale & Color\\
            \hline
            Resolution & \begin{tabular}{@{}c@{}} 133 $\times$ 133 until \\ 1024 $\times$ 1024\end{tabular} & - & 1024 $\times$ 1024\\
            \hline
            \begin{tabular}{@{}l@{}}Expert \\ ground-truth\end{tabular} & \xmark & \cmark & \xmark\\
            \noalign{\hrule height 0.3mm}
        \end{tabular}
    \end{adjustbox}
    \vspace{-0.7em}
    \caption{Comparison between our dataset and other public old photos datasets in several factors.}
    \label{tab:dataset_comparison}
    \vspace{-1em}
\end{table}

\subsection{Comparison of CHD and Other Old Photos Dataset}

There are two other public old photo datasets, such as the Historical Wiki Face Dataset (HWFD) \cite{luo2021time} and RealOld \cite{xu2022pik}.
However, when this paper was submitted, RealOld \cite{xu2022pik} had not been published yet.
Table. \ref{tab:dataset_comparison} shows the comparison between our and other datasets.
Our dataset is mainly focused on old color photos produced during the 20th century using reversal film \cite{machidon2018digital}, which have specific degradations such as color fading (shown in Fig. \ref{fig:1_1_dataset_examples}) and have not yet been analyzed before.
Meanwhile, other datasets contain greyscale photos.
Regarding the diversity of content, our dataset contains complex and diverse scenes of indoor and outdoor natural scenes, as shown in Fig. \ref{fig:1_1_dataset_examples}. 
In contrast, other datasets only contain portrait and face photos which are much simpler than natural scene photos.
In addition to the complexity of the scenes, our dataset also has a larger number of images compared to the two other datasets.
Fig. \ref{fig:1_2_dataset_comparison} shows some visual examples of our and other datasets.

\section{Results on Real Old Photos in The Wild}

\begin{figure*}[ht]
\centering
\includegraphics[width=0.80\textwidth]{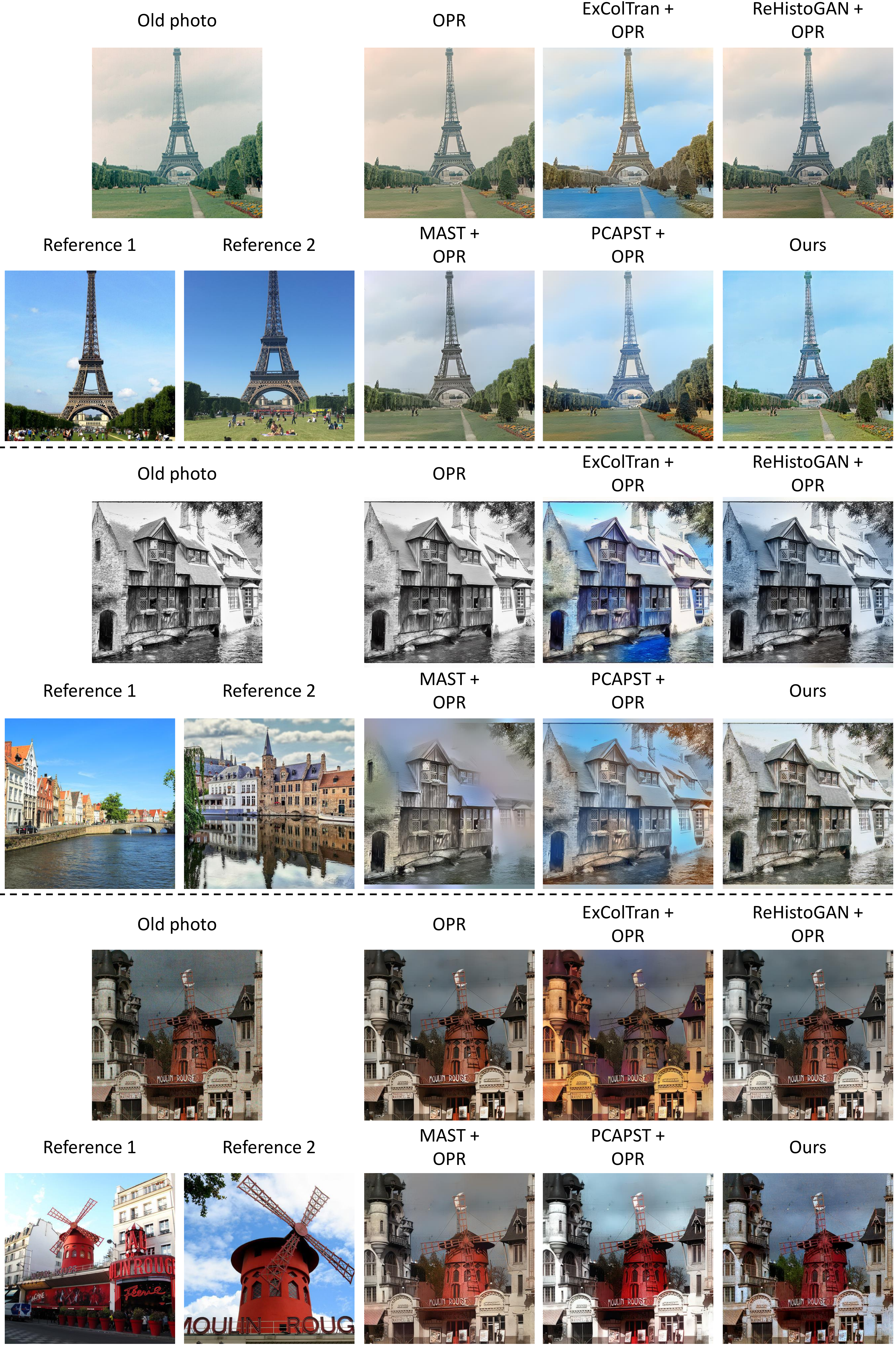}
\vspace{-1em}
\caption{Comparison of modernization on old photos in the wild between our method and other baselines. In most cases, our method outperforms other methods (OPR \cite{wan2020bringing}, ExColTran\cite{yin2021yes} + OPR, ReHistoGAN\cite{afifi2021histogan} + OPR, MAST\cite{huo2021manifold} + OPR, and PCAPST\cite{chiu2022pca} + OPR) in modernizing old photos in the wild showing the robustness of our method. Other reference-based baselines use reference 1 as their reference.}
\label{fig:7_2_real_old_part_1}
\vspace{-1em}
\end{figure*}

\begin{figure*}[ht]
\centering
\includegraphics[width=0.8\textwidth]{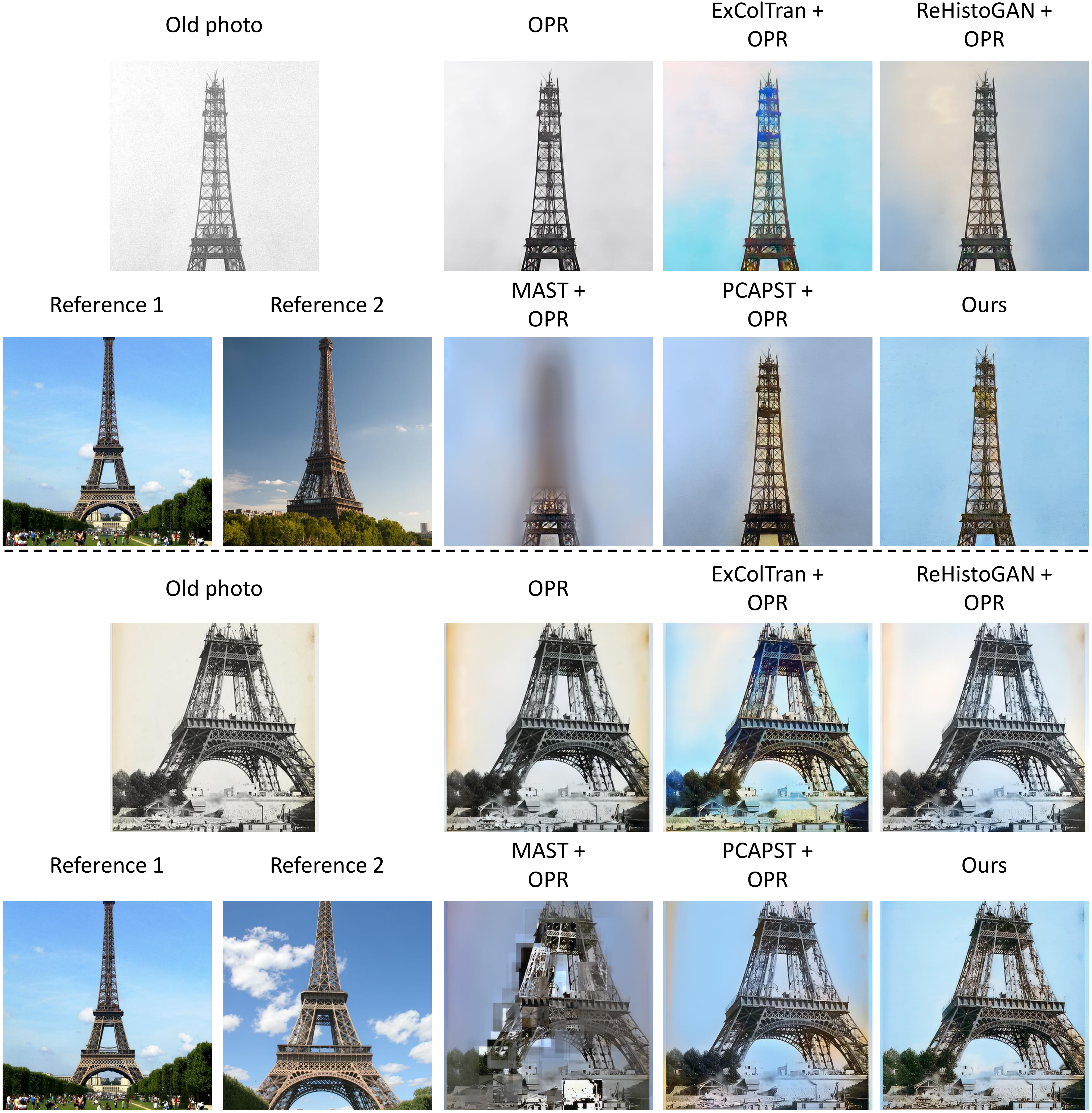}
\vspace{-0.5em}
\caption{Comparison of modernization on greyscale old photos in the wild between our method and other baselines. Our method outperforms other methods (OPR \cite{wan2020bringing}, ExColTran\cite{yin2021yes} + OPR, ReHistoGAN\cite{afifi2021histogan} + OPR, MAST\cite{huo2021manifold} + OPR, and PCAPST\cite{chiu2022pca} + OPR) in modernizing greyscale old photos in the wild showing the generalization of our method. Other reference-based baselines use reference 1 as their reference. In these examples, our method can better match the corresponding semantic regions between the old photo and multiple references even though the \textbf{viewpoint and scale are significantly different} (e.g., the viewpoint and scale of the tower).}
\label{fig:7_3_real_old_part_2}
\vspace{-0.5em}
\end{figure*}

Fig. \ref{fig:7_2_real_old_part_1} and Fig. \ref{fig:7_3_real_old_part_2} show the generalization and robustness of our method when applied to real old photos in the wild.
The first and second examples of Fig. \ref{fig:7_2_real_old_part_1} show that our method outperforms other baselines in modernizing old color photos and even can work for greyscale photos.
Interestingly, our method can achieve natural modernization results on greyscale old photos (second example) even when compared to the colorization baseline (ExColTran\cite{yin2021yes} + OPR).
For the third example in Fig. \ref{fig:7_2_real_old_part_1}, our method achieves the second-best performance compared to `PCAPST\cite{chiu2022pca} + OPR', where our method can better stylize the trees but fail to stylize the big castle, caused by our alignment module that may think that castle and building are different.

Fig. \ref{fig:7_3_real_old_part_2} shows additional examples of modernization on greyscale old photos in the wild.
The same observation can be seen where our method outperforms other baselines even when compared to the colorization baseline (ExColTran\cite{yin2021yes} + OPR).
Interestingly, we can handle paper blotches in the second example of Fig. \ref{fig:7_3_real_old_part_2} even though our method is not trained with this kind of artifact.
Meanwhile, the baseline OPR trained with this kind of artifact further highlights the paper blotches artifact instead of removing them.
In addition, compared to other reference-based methods, our method can better match similar semantic regions between the old photo and references even though the viewpoint and scale are significantly different.
For example, the Eiffel tower in the old photo of the first and second examples of Fig. \ref{fig:7_3_real_old_part_2} have different viewpoints and scales compared to the references.
However, our method can faithfully match the Eiffel tower in the old photo and references, thus resulting in better stylization and modernization results.
Note that our network can achieve all these results without using old photos during training.

\section{Details \& Analyses of Our Photorealistic Style Transfer (PST) Network}

\subsection{Comparison Between Our Photorealistic Style Transfer (PST) Network and WCT2 \cite{yoo2019photorealistic}}
\begin{figure*}[ht]
\centering
\includegraphics[width=0.6\textwidth]{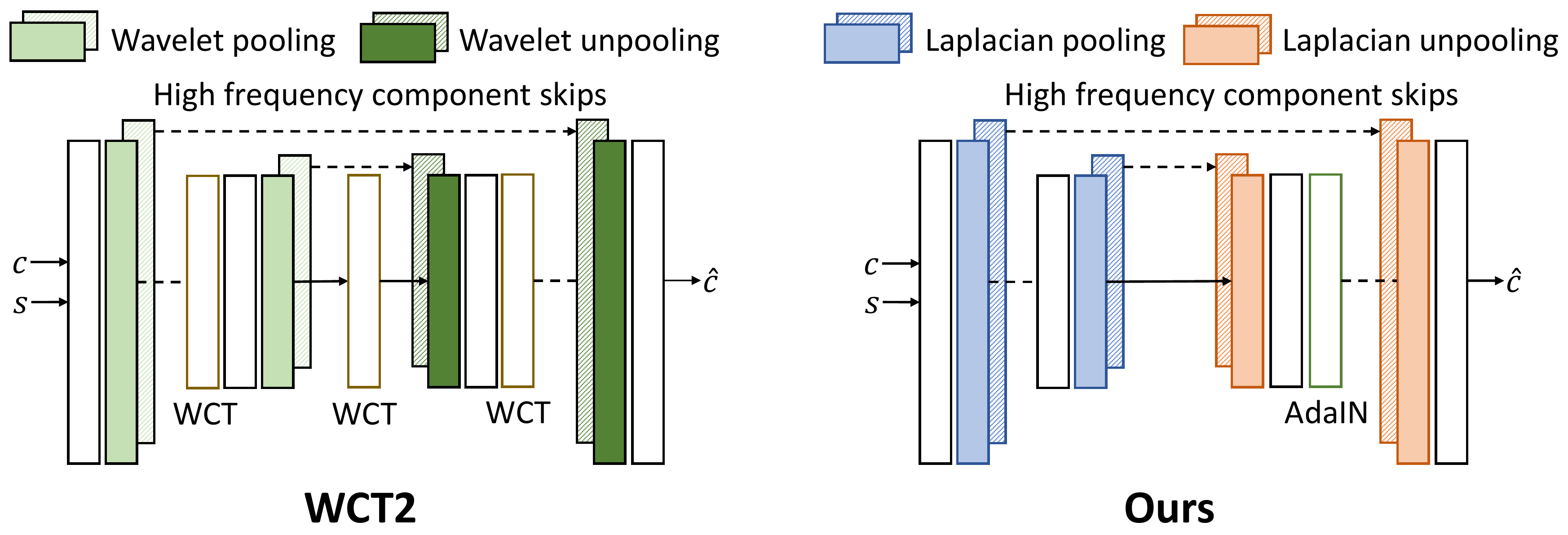}
\vspace{-0.8em}
\caption{Comparison between our PST network and WCT2 \cite{yoo2019photorealistic}.}
\label{fig:2_1_photost_architecture_comparison}
\vspace{-1em}
\end{figure*}

Fig. \ref{fig:2_1_photost_architecture_comparison} shows the comparison between our PST network and WCT2 \cite{yoo2019photorealistic} architecture.
We propose our photorealistic style transfer (PST) network to address some drawbacks of the concatenated version of the WCT2 network with progressive stylization (style transfer). 
Specifically, our PST network only transfers a single high-frequency component in level-0 of the Laplacian pyramid representation \cite{burt1987laplacian}.
Meanwhile, WCT2 \cite{yoo2019photorealistic} transfers three different high-frequency components of wavelet-based skip connection.
This modification addresses the "short circuit" issue explored in \cite{an2020ultrafast}, which makes the stylization of WCT2's decoder part only has an effect when applied in the last decoder block (shown in Fig. \ref{fig:2_4_stylization_decoder}).
Then, we only apply progressive stylization in the decoder part, especially the last two decoder blocks, which achieves the best trade-off between the stylization effect and the photorealism.
The last improvement is we use differentiable adaptive instance normalization (AdaIN) \cite{huang2017arbitrary} instead of the non-differentiable whitening-and-coloring transformation (WCT) \cite{li2017universal} to enable the learning and prediction of local style, which can enable our single stylization subnet to perform local style transfer without any semantic segmentation mask.
Further analyses of the drawbacks of WCT2\cite{yoo2019photorealistic} are explained in the following subsection.

\subsection{Additional Analyses}

\begin{figure}[ht]
\centering
\includegraphics[width=\columnwidth]{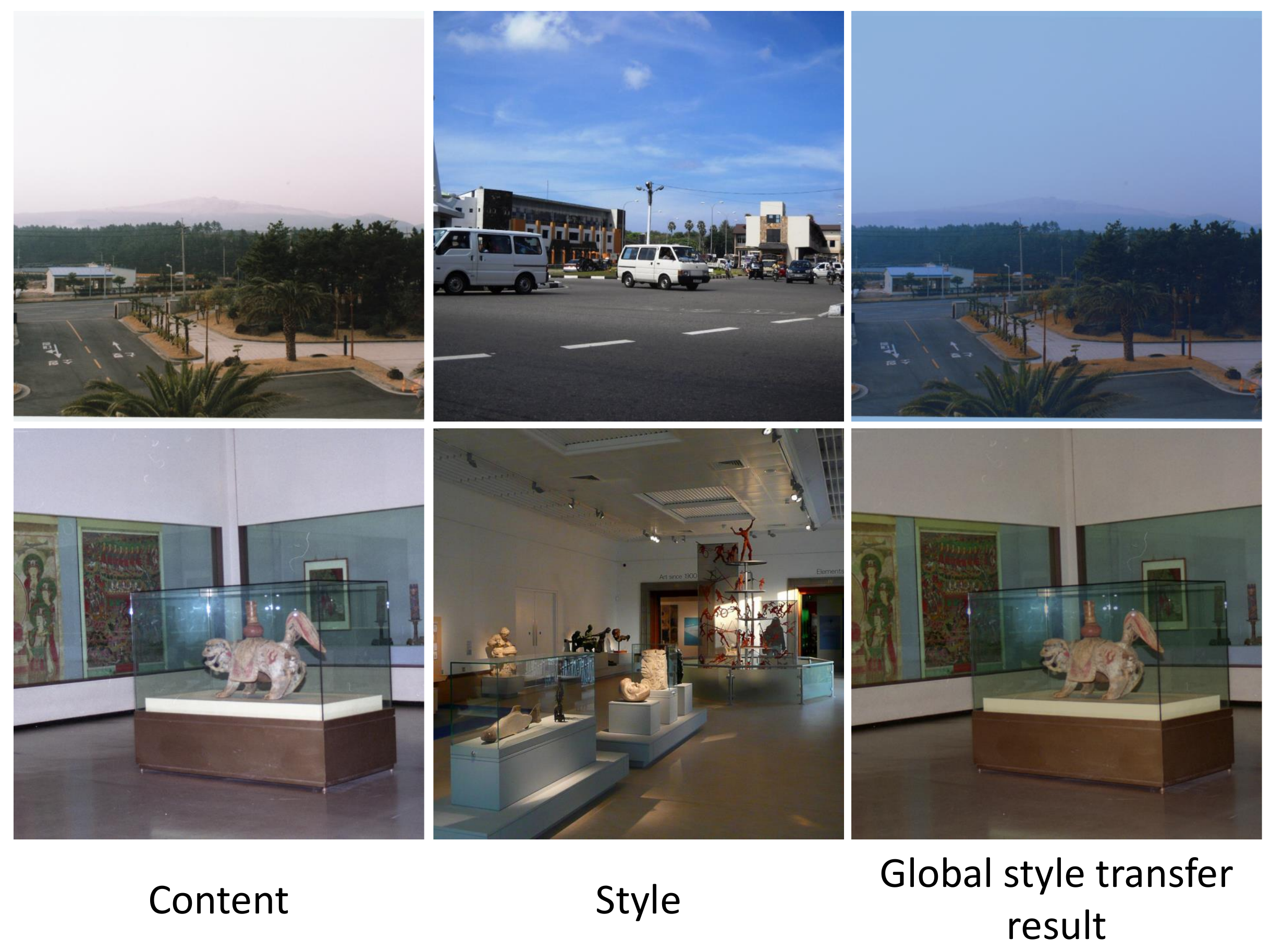}
\vspace{-1.5em}
\caption{Unnatural global style transfer result of WCT2\cite{yoo2019photorealistic}.}
\label{fig:2_2_unnatural_global_st}
\vspace{-0.5em}
\end{figure}

\begin{figure}[ht]
\centering
\includegraphics[width=\columnwidth]{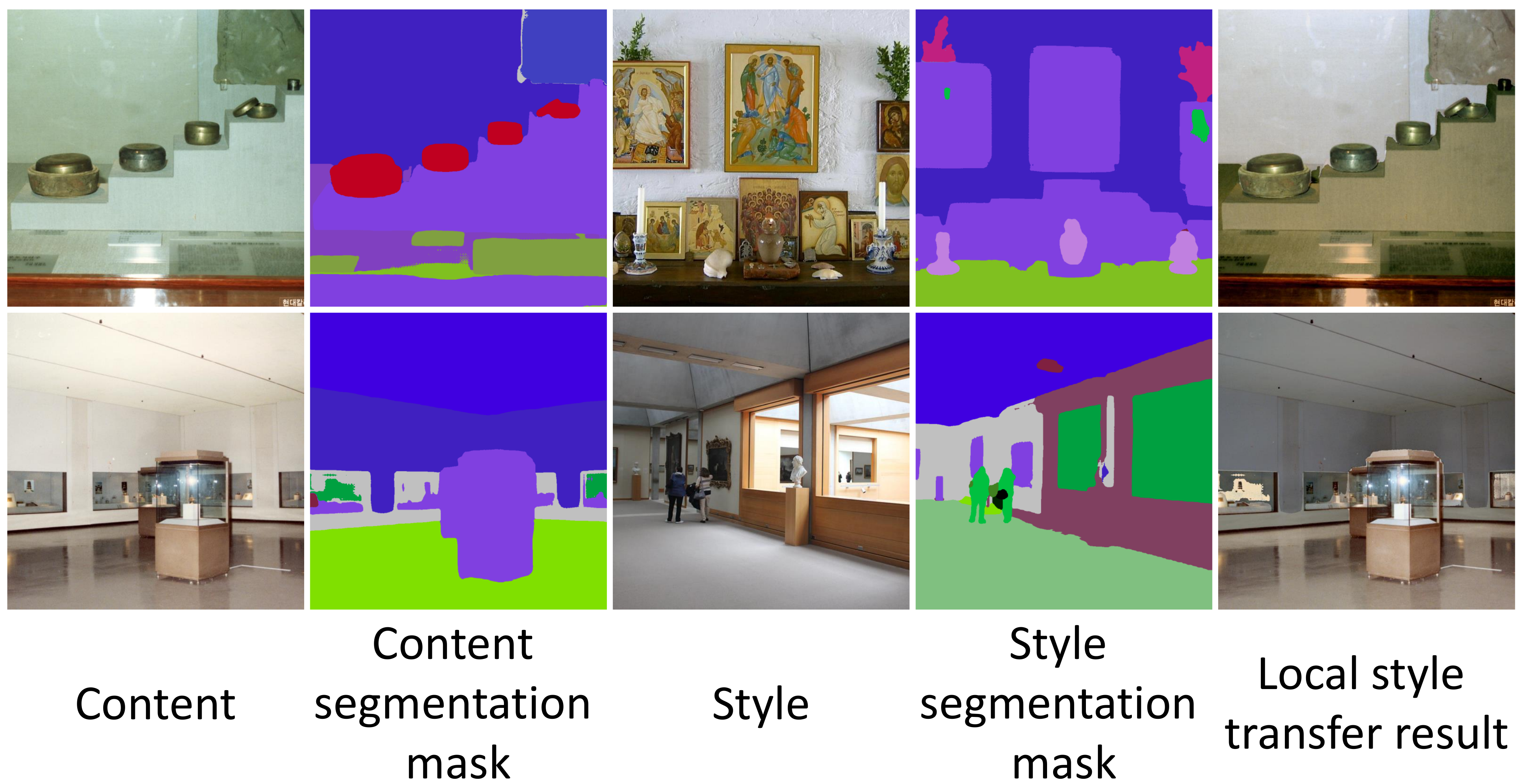}
\vspace{-1.5em}
\caption{Unnatural local style transfer result of WCT2\cite{yoo2019photorealistic}.}
\label{fig:2_3_unnatural_local_st}
\vspace{-0.5em}
\end{figure}

\noindent
\textbf{Limitations of WCT2 \cite{yoo2019photorealistic} for old photo modernization.}
There are two main limitations of WCT2 \cite{yoo2019photorealistic} that can prevent its application on real-world old photos.
The first limitation is the unnatural global style transfer results shown in Fig. \ref{fig:2_2_unnatural_global_st}, where this unnatural result may make the photo look like an old photo instead of modernizing them.
Meanwhile, Fig. \ref{fig:2_3_unnatural_local_st} shows the second limitation of WCT2.
We generate the semantic segmentation masks using VIT-Adapter\cite{chen2022vision}, which is one of the state-of-the-art models in the semantic segmentation task for the examples in Fig. \ref{fig:2_3_unnatural_local_st}.
As can be seen, WCT2 needs a near-perfect semantic segmentation mask to produce satisfactory results of local style transfer.
However, generating a near-perfect segmentation mask moreover for old photos is highly challenging even with one of the SOTA networks.
Therefore, we propose our single stylization subnet to overcome these limitations, especially to perform local style transfer without any semantic segmentation mask and produce natural global and local style transfer results.

\begin{figure}[ht]
\centering
\includegraphics[width=\columnwidth]{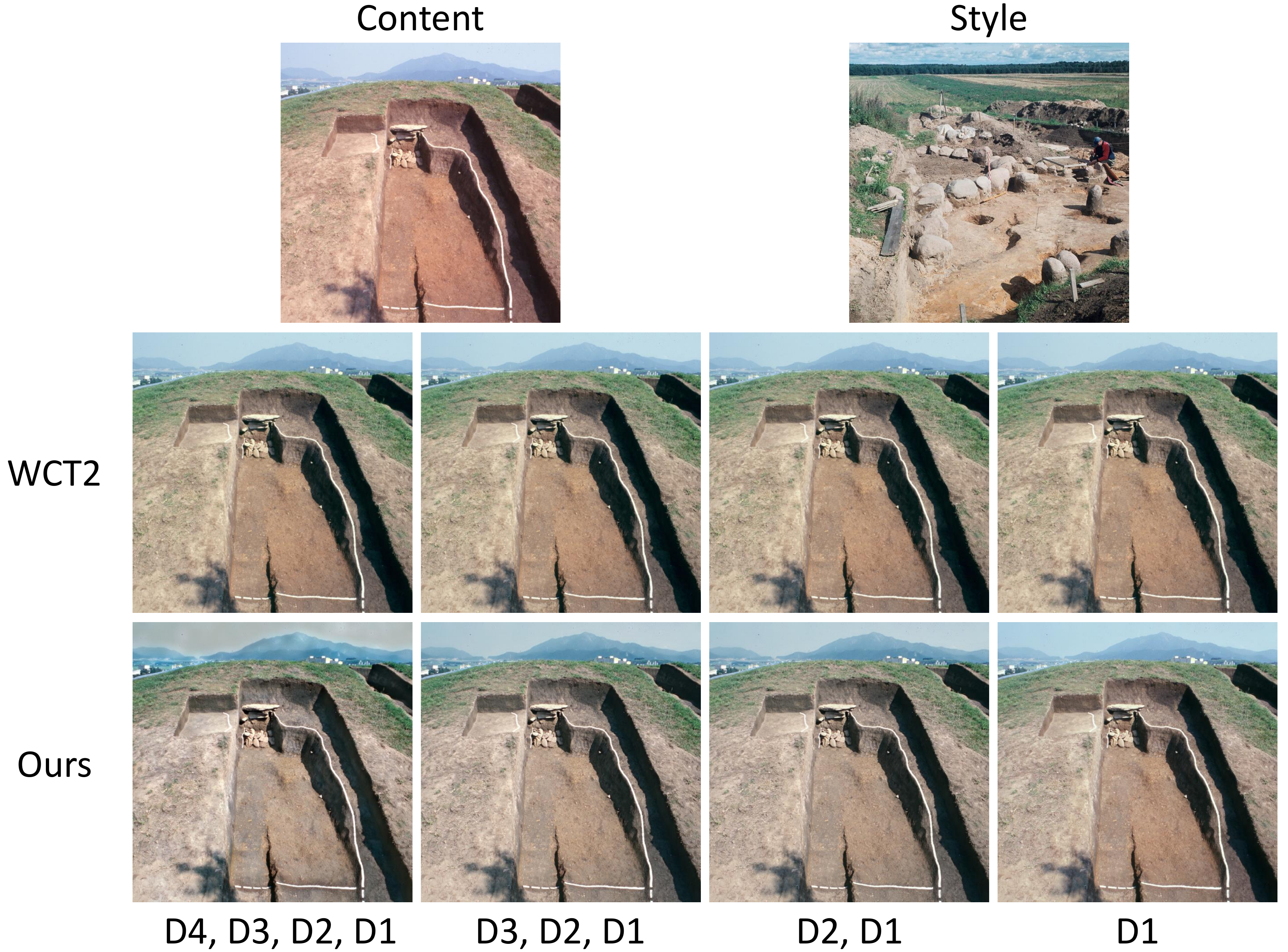}
\vspace{-1.5em}
\caption{Applying feature transformation in different decoder blocks of WCT2 \cite{yoo2019photorealistic} and our PST network. D denotes the decoder block, while the number denotes the decoder block number (a higher number denotes the decoder block in deeper feature space).}
\label{fig:2_4_stylization_decoder}
\vspace{-1em}
\end{figure}

\noindent
\textbf{The trade-off between stylization and photorealism.}
Fig. \ref{fig:2_4_stylization_decoder} shows that applying progressive stylization in different decoder blocks does not affect the original concatenated version of the WCT2 \cite{yoo2019photorealistic} network.
Thus, we modify the skip connection using the aforementioned laplacian-based skip connection to overcome this limitation.
In terms of progressive stylization, we only apply feature transformation on the decoder part using AdaIN to perform style transfer, especially the last two decoder blocks, to achieve the best trade-off between stylization and photorealism.
As shown in Fig. \ref{fig:2_4_stylization_decoder}, applying feature transformation in the deep feature space (D4 or D3) produces stylization artifacts in the output, making them look non-photorealistic.
Thus, applying feature transformation in the shallow feature space (D1 and D2) achieves the best stylization and photorealistic results.

\begin{figure}[ht]
\centering
\includegraphics[width=\columnwidth]{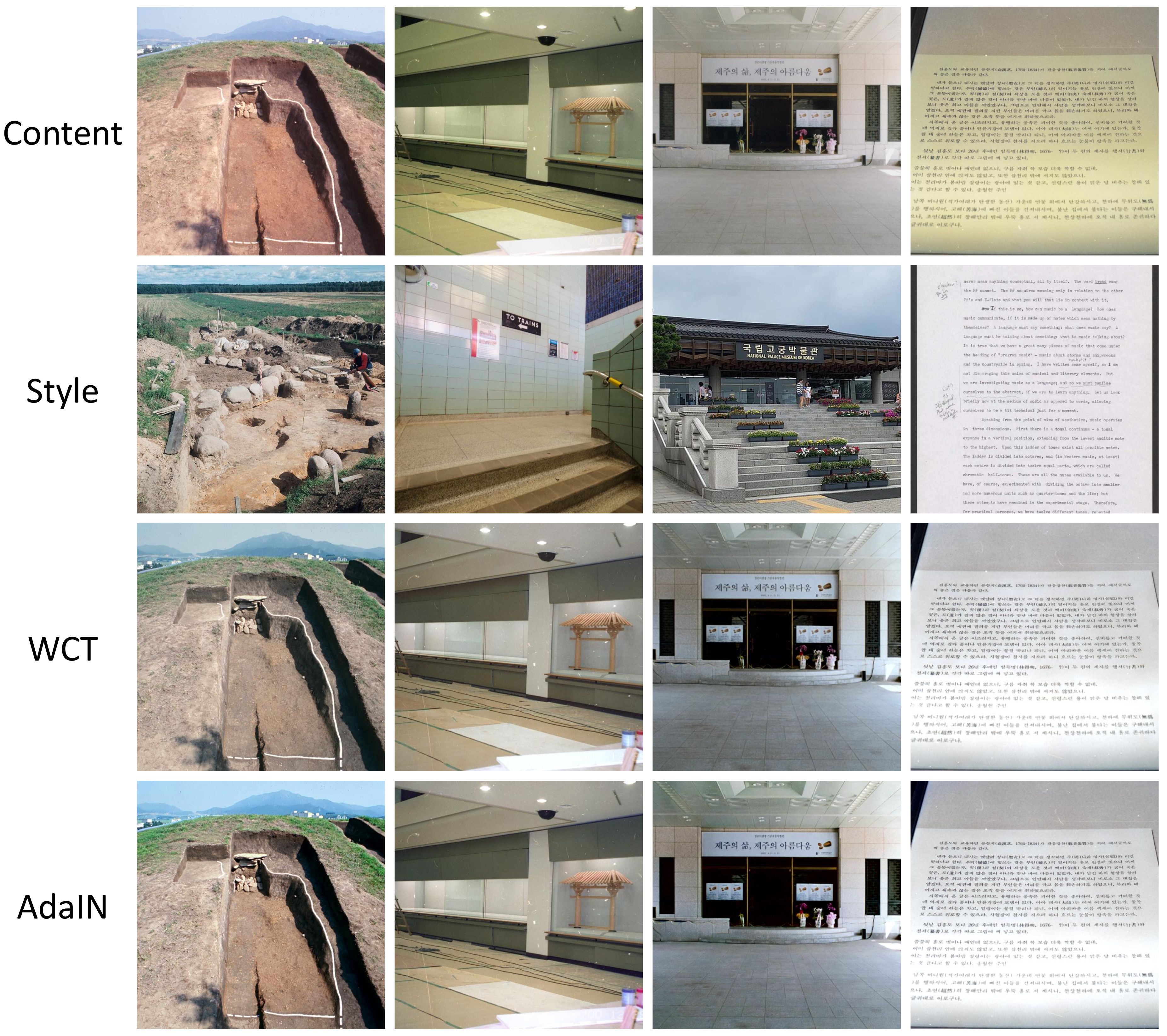}
\vspace{-1.5em}
\caption{Visual results comparison between AdaIN \cite{huang2017arbitrary} and WCT \cite{li2017universal} feature transformations.}
\label{fig:2_5_stylization_transformation}
\vspace{-1em}
\end{figure}

\noindent
\textbf{Comparison between different feature transformations.}
In our PST network, we use AdaIN instead of WCT, which is commonly used as the feature transformation to perform photorealistic style transfer.
We observe that using AdaIN achieves more improved stylization with better color saturation which can help us to achieve superior modernization as shown in Fig. \ref{fig:2_5_stylization_transformation}.
In addition, AdaIN feature transformation is also differentiable, which can help us achieve local style transfer without any semantic segmentation mask since we want to learn and predict the local styles instead of computing them.

\section{Details of MROPM-Net Architecture}

\subsection{Single Stylization Subnet}

\begin{figure*}[ht]
\centering
\includegraphics[width=\textwidth]{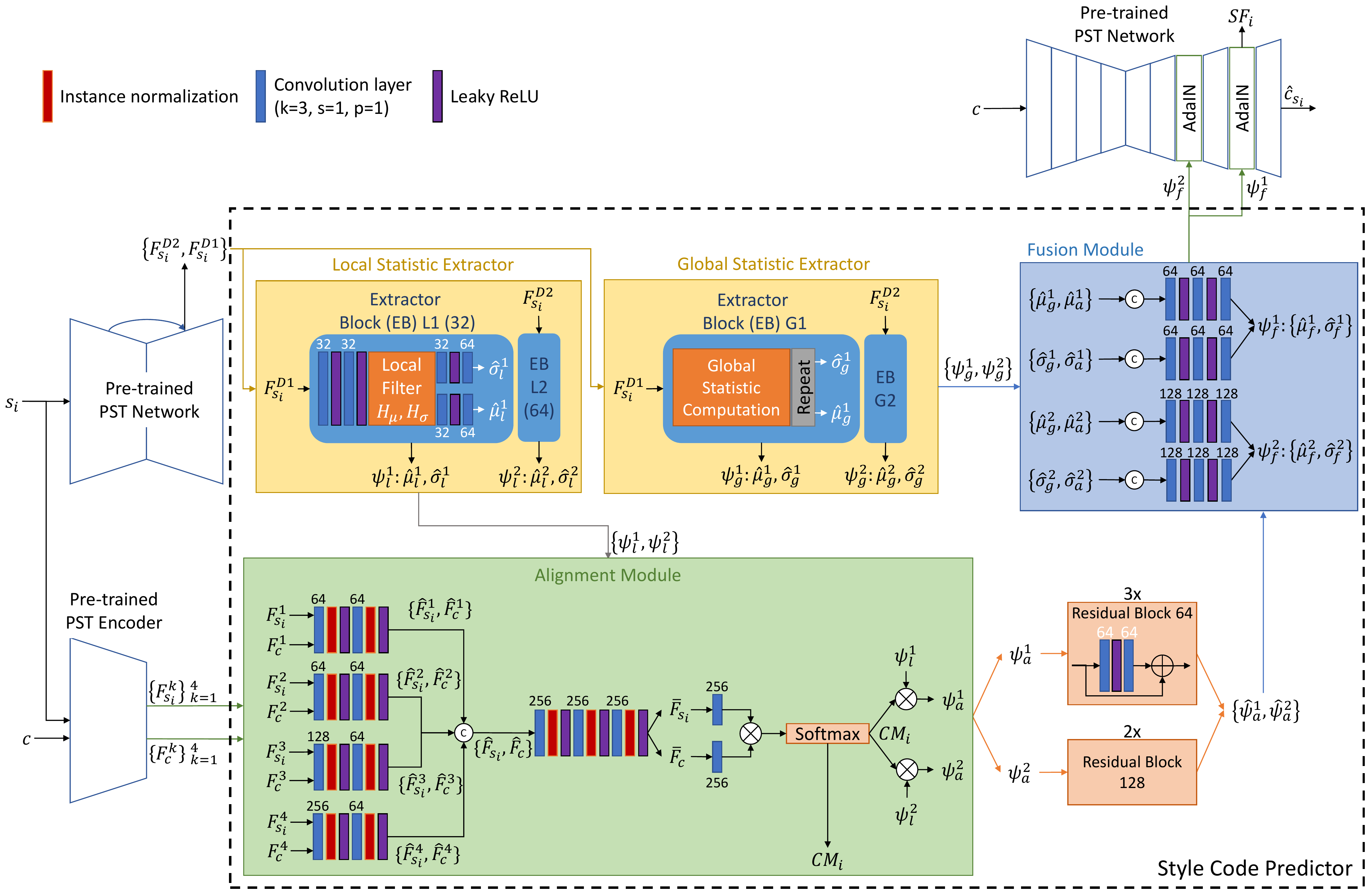}
\vspace{-2em}
\caption{Detailed architecture of our single stylization subnet $\mathcal{S}$.}
\label{fig:3_1_single_stylization_subnet}
\vspace{-0.5em}
\end{figure*}

The detailed architecture of our single stylization subnet $\mathcal{S}$ can be seen in Fig. \ref{fig:3_1_single_stylization_subnet}.
We only describe additional parts that have not been described in the main paper, which is the alignment module (green part).
Given an old photo $c$, modern style reference $s_i$, and extracted local style code $(\psi_l^1, \psi_l^2)$, the alignment module aligns the local style code of $s_i$ to $c$.
In the alignment module, we map the extracted multi-level feature maps $\{F_{c}^{k}\}_{k=1}^4$ and $\{F_{s_i}^{k}\}_{k=1}^4$ for both $c$ and $s_i$, respectively using shared convolution blocks, and perform matrix multiplication between mapped features to obtain correlation matrix $CM_i$ similar to non-local attention \cite{wang2018non}.
Since different feature maps have different spatial resolutions, we map them into the same spatial resolution, which is the spatial resolution of the deepest features, i.e., the spatial resolution of $F_{c}^4$, using nearest neighbor interpolation.
The next step is to align the local style code $(\psi_l^1, \psi_l^2)$ using correlation matrix $CM_i$ via matrix multiplication, thus resulting in aligned style codes $(\psi_{a}^1, \psi_{a}^2)$. 
Since different $\{\psi_l^k\}_{k=1}^2$ have a different spatial resolution than $CM_i$, we use nearest neighbor interpolation to map $\{\psi_l^k\}_{k=1}^2$ to the same spatial resolution of $CM_i$ and then map it back to the original spatial resolution after multiplication with $CM_i$.
Then, we use three residual blocks to refine $\psi_{a}^1$ and two residual blocks to refine $\psi_{a}^2$.

\begin{figure*}[ht]
\centering
\includegraphics[width=0.6\textwidth]{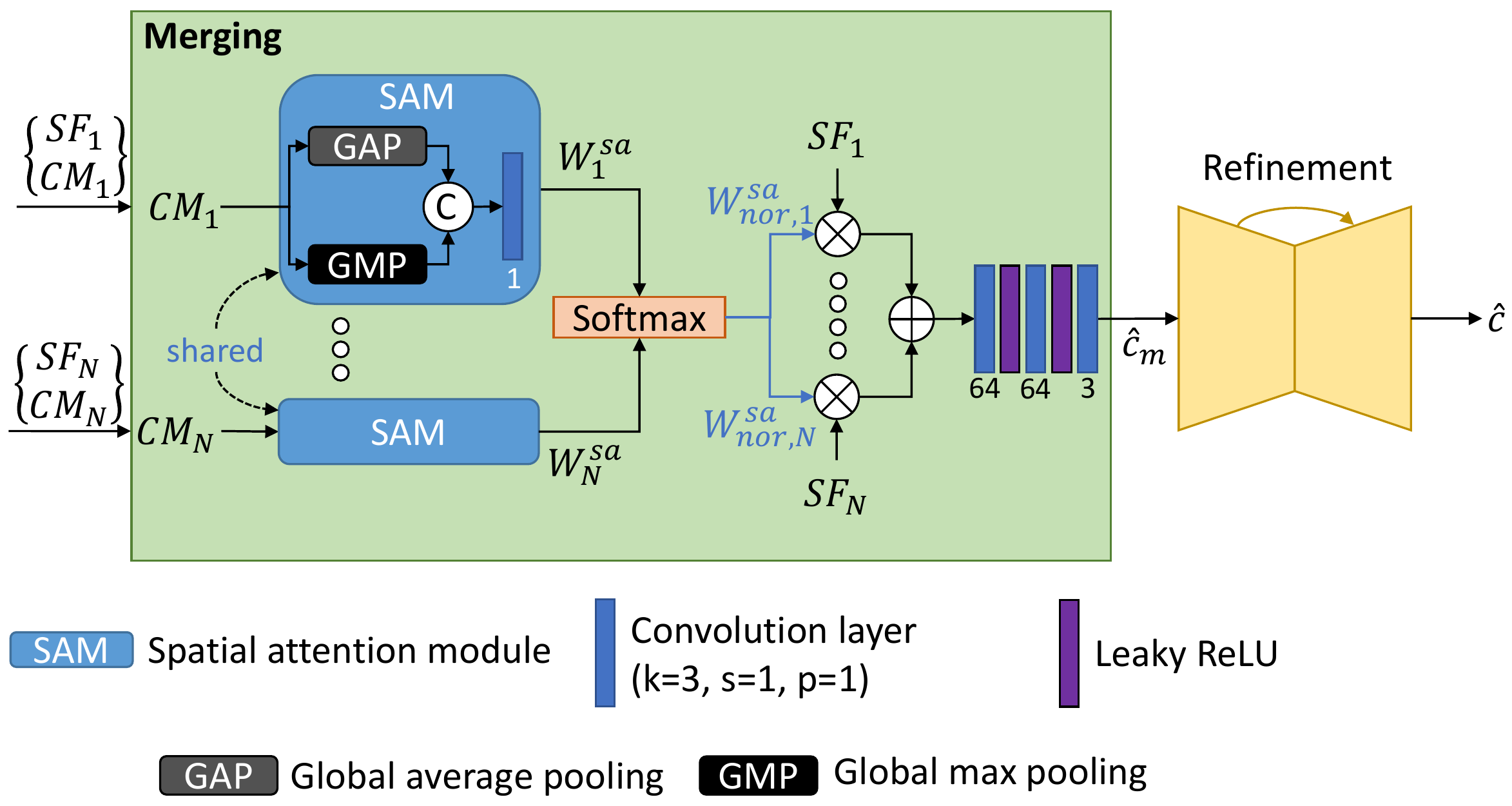}
\vspace{-0.5em}
\caption{Detailed architecture of our merging-refinement subnet $\mathcal{M}$.}
\label{fig:3_2_merging_refinement_subnet}
\vspace{-1em}
\end{figure*}

\subsection{Merging-Refinement Subnet}

Fig. \ref{fig:3_2_merging_refinement_subnet} shows the detailed architecture of our merging-refinement subnet $\mathcal{M}$.
We show the details of the spatial attention module \cite{woo2018cbam}.
Additionally, we show the details of convolution blocks that consist of several convolution layers and leaky ReLU activation, in order to get the intermediate merging output $\hat{c}_m$.
For the details of the refinement subnet, we follow the notation of U-Net \cite{ronneberger2015u} architecture in \cite{isola2017image}.
Specifically, the encoder-decoder architecture is based on the following:

\noindent \textbf{encoder}:

\noindent $C64-C128-C256-C512-C512-C512-C512$

\noindent \textbf{decoder}:

\noindent $CD512-CD512-CD512-CD256-CD128-CD64$

The activation functions in the encoder are leaky ReLUs with a slope of 0.2, while the activation functions in the decoder are ReLUs.
Then, we use a single convolution layer, followed by a single Tanh function, to map the features of the last layer decoder to the RGB channels representing modernized images.
We use instance normalization layers \cite{ulyanov2016instance} in the U-Net architecture.

\section{Additional Details of Synthetic Data Generation Scheme}
In this section, we describe additional details of the synthetic data generation scheme, such as the style variant transformation which includes the color jittering and unstructured degradation, and the details of the style invariant transformations.
To get the degraded images via the style variant transformation, we first perform color jittering on the images by randomly changing the brightness, contrast, saturation, and hue with the magnitude of 0.2, 0.2, 0.4, and 0.4, respectively.
In addition, we also apply a random sequence of mixed unstructured degradation after color jittering. 
Specifically, we choose a random sequence of the following degradations:
\begin{itemize}
\item Gaussian blur with a probability of 50\%, where the kernel size is chosen randomly between 3, 5, and 7 and the standard deviation $\sigma = 0.004-0.02$.
\item Random noise with a probability of 50\%, where we choose randomly between gaussian noise with ($\mu=0, \sigma=0.02-0.04$), and speckle noise with ($\mu=0, \sigma=0.02-0.08$).
\item Random resizing artifacts with a probability of 50\%. The resizing artifact is generated by downsampling the spatial resolution of the image to the half size using bicubic downsampling and then upsampling the downsampled image back to the original spatial resolution by using nearest or bilinear interpolation, which is chosen randomly.
\item Random JPEG artifact with a probability of 50\% where the compressed quality is a random number between 40\% to 100\% (no artifact).
\end{itemize}
\noindent
\textbf{How to adapt to new degradation.} 
In this work, we focus more on unstructured degradation since at the time this work was published, no public scratches data were available.
However, one can easily add new degradation or special artifacts into our style variant transformation.
By doing so, the network can adapt and handle new degradation or artifacts.

For the style invariant transformations, we apply a sequence of the following operators:
\begin{enumerate}
\item Random $k \times 90^{\circ}$ rotations chosen randomly between $90^{\circ}$, $180^{\circ}$, and $270^{\circ}$.
\item Random translation for regions that can be translated (the translated regions still remain inside the boundary of the image after the translation). 
\item Random left-to-right flipping.
\item Random up-to-down flipping.
\end{enumerate}

\section{Additional Experiments on Baselines}

In the main paper, we propose to use a sequence of stylization and enhancement as the baselines to compare with our method since our method can perform both stylization and enhancement jointly.
In this section, we first show the results of retraining the baseline  OPR\cite{wan2020bringing} using our synthetic data and our CHD training set since we use the original pretrained baseline model in the main paper.
Then, we show the results of using only stylization to show that an enhancement method is required to further improve the results.
Furthermore, we show the results of using a sequence of enhancement and stylization (reversed order) as the baselines compared to a sequence of stylization and enhancement.

\noindent
\textbf{Baselines.}
We choose four different state-of-the-art (SOTA) stylization methods, from exemplar-based colorization\cite{yin2021yes}, recolorization\cite{afifi2021histogan}, and photorealistic style transfer\cite{huo2021manifold,chiu2022pca}.
Even though exemplar-based colorization and recolorization can only change the color, we still use them as the baseline since changing the color can also affect the look of an image.
Our user study also shows that the recolorization baseline achieves better results than other baselines.
Specifically, we choose the following baselines that act as the stylization:
\begin{itemize}
\item exemplar-based colorization: transformer-based method (ExColTran \cite{yin2021yes})
\item recolorization: color-controlled GAN method (ReHistoGAN \cite{afifi2021histogan})
\item photorealistic style transfer (PST): semantic PST (MAST \cite{huo2021manifold}) and PCA-based knowledge distillation PST (PCAPST \cite{chiu2022pca})
\end{itemize}
For the enhancement, we use the SOTA of old photo restoration (OPR \cite{wan2020bringing}) as the baseline.
Note that, OPR is used for enhancement since OPR can handle both unstructured degradation and structured degradation.
Thus, it is used as an enhancement method in conjunction with stylization baselines for fair comparison since our method can perform both stylization and enhancement.

\begin{figure*}[ht]
\centering
\includegraphics[width=0.85\textwidth]{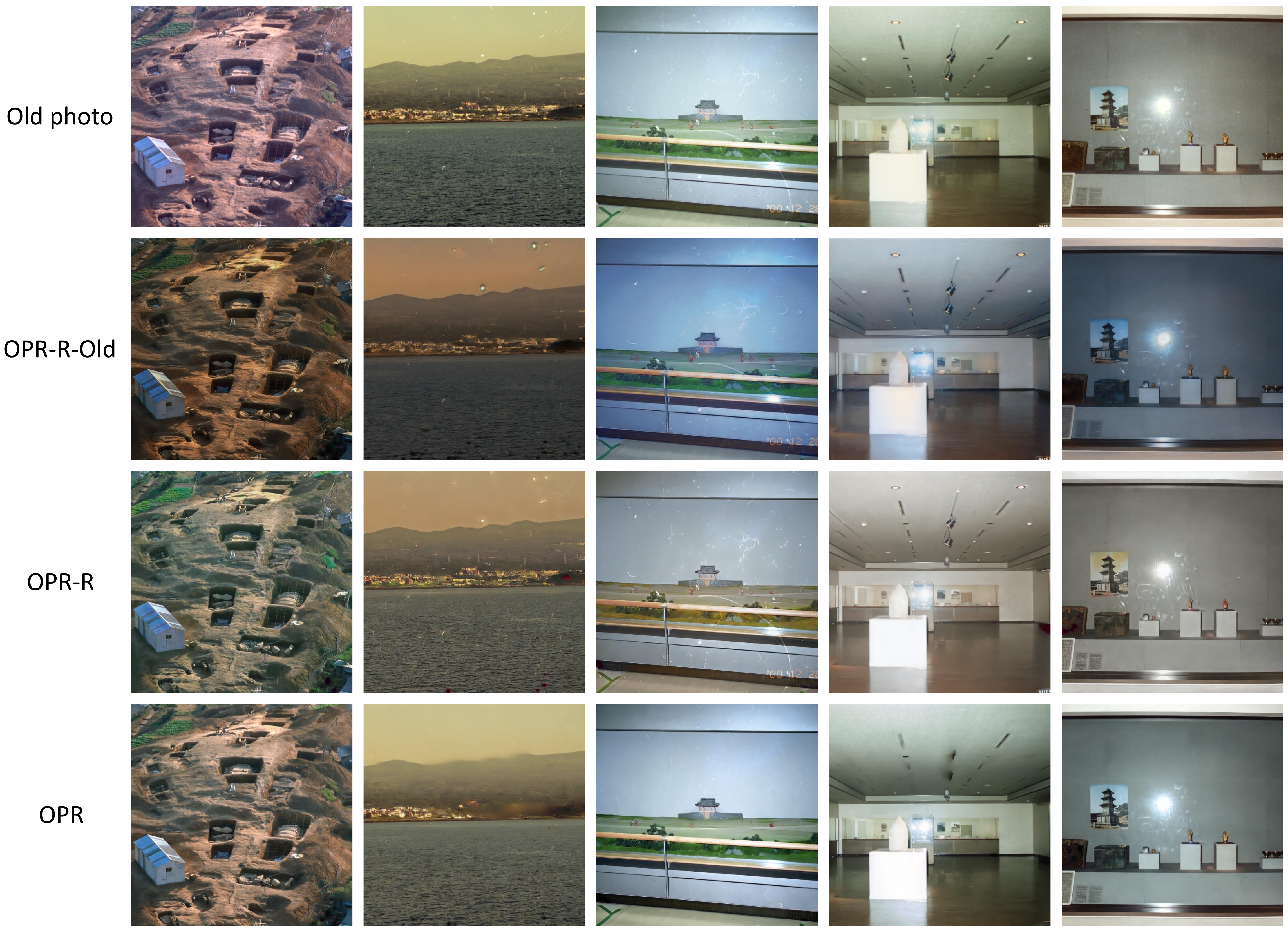}
\vspace{-1em}
\caption{The results of retraining OPR. OPR \cite{wan2020bringing} denotes the original pretrained model. OPR-R-Old denotes the model retrained using the original synthetic training data and our CHD training set, while OPR-R denotes the model retrained using our proposed synthetic data and our CHD training set.}
\label{fig:5_1_opr_training}
\vspace{-1em}
\end{figure*}

\vspace{0.3em}
\noindent
\textbf{Qualitative results of retraining the baseline OPR \cite{wan2020bringing}.}
As mentioned in the main paper, we use the pretrained model of baseline OPR \cite{wan2020bringing} rather than retraining it for real old photo evaluation. 
The results with the retrained baseline OPR \cite{wan2020bringing} are shown in Fig. \ref{fig:5_1_opr_training} both when using their synthetic data and our CHD training set (denoted as OPR-R-Old) and when using our synthetic data and our CHD training set (denoted as OPR-R).
Interestingly, training using their synthetic data and our CHD training set (OPR-R-Old) results in worse performance in real old photos.
This may suggest that OPR \cite{wan2020bringing} requires a large number of old photos since the authors trained the OPR network with 5,718 private real old photos.
Meanwhile, our CHD training set only contains 514 old photos. 
Another hypothesis of the training failure is that our collection of old photos has a larger diversity compared to the portrait photos used to train the original OPR network\cite{wan2020bringing}. 
In addition, since the baseline OPR is not capable of handling diverse color jittering degradation, the results of the retrained baseline OPR using our synthetic training data and CHD training set (OPR-R) are inferior to those of the pretrained baseline OPR model\cite{wan2020bringing} in real old photos evaluation.

\begin{figure*}[ht]
\centering
\includegraphics[width=0.9\textwidth]{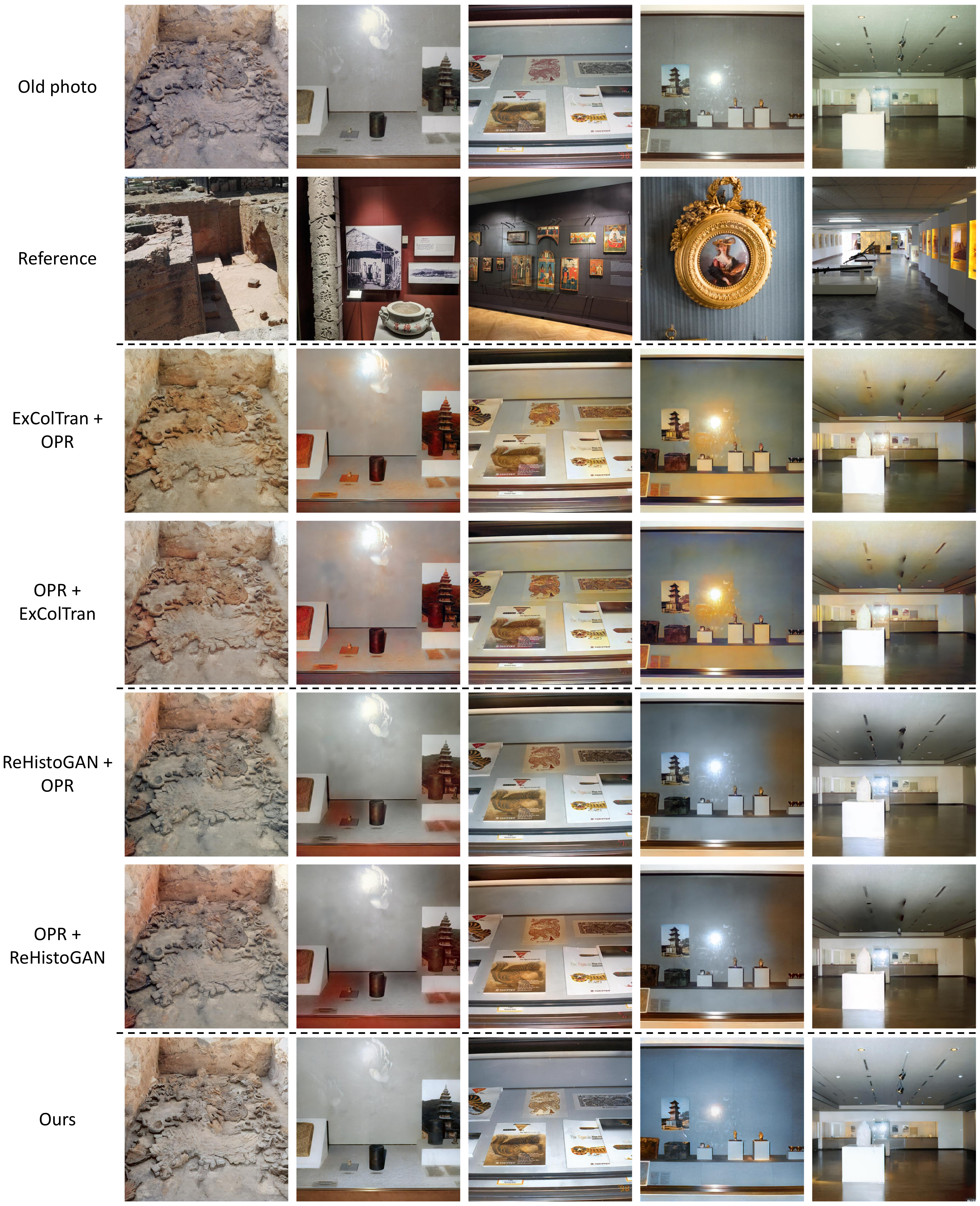}
\vspace{-1em}
\caption{Comparison between `stylization + enhancement' and `enhancement + stylization' for ExColTran \cite{yin2021yes} and ReHistoGAN \cite{afifi2021histogan}.}
\label{fig:5_2_stylization_enhnancement_part_1}
\vspace{-1em}
\end{figure*}

\begin{figure*}[ht]
\centering
\includegraphics[width=0.9\textwidth]{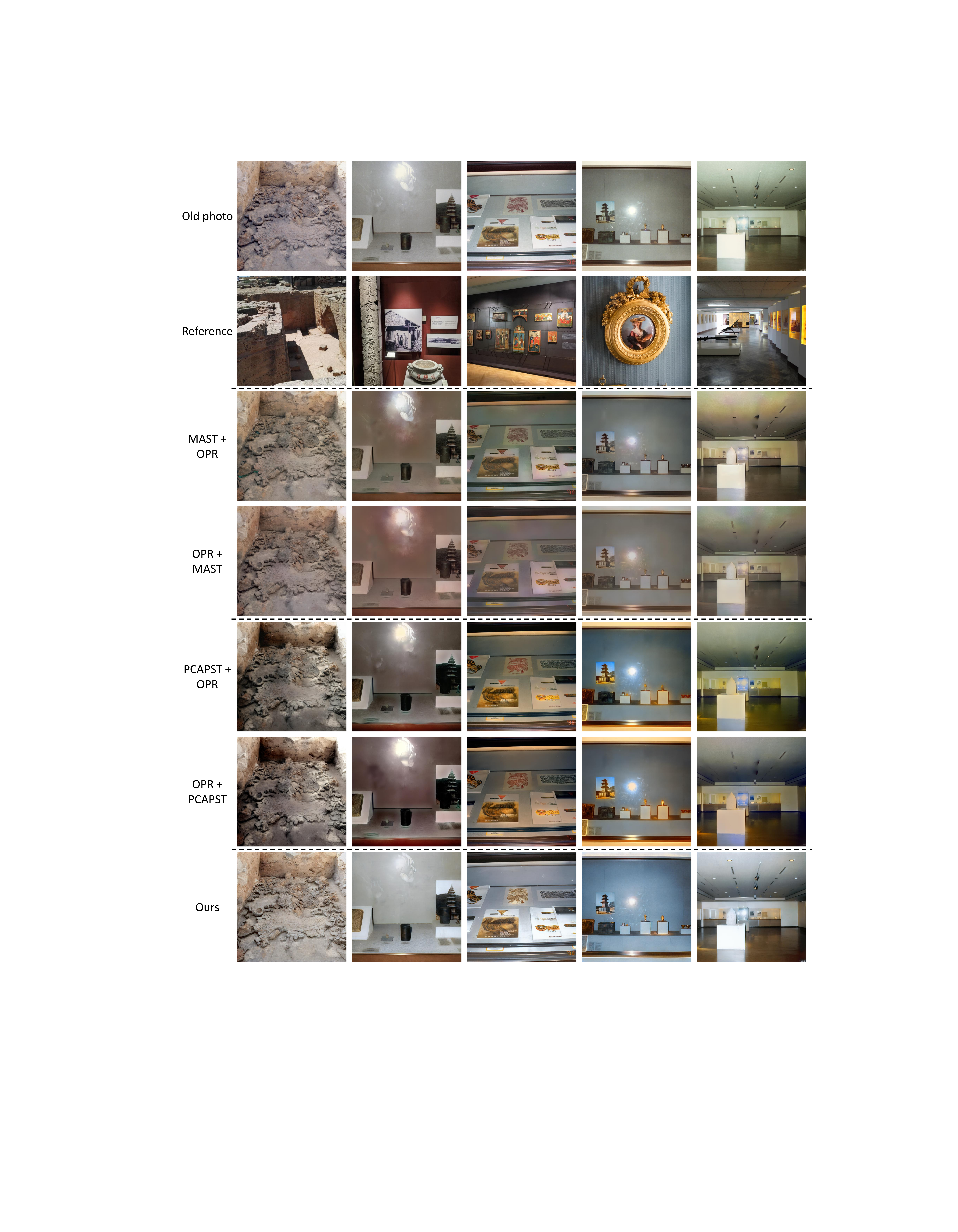}
\vspace{-1em}
\caption{Comparison between `stylization + enhancement' and `enhancement + stylization' for MAST \cite{huo2021manifold} and PCAPST \cite{chiu2022pca}.}
\label{fig:5_3_stylization_enhnancement_part_2}
\vspace{-1em}
\end{figure*}

\begin{figure*}[ht]
\centering
\includegraphics[width=0.75\textwidth]{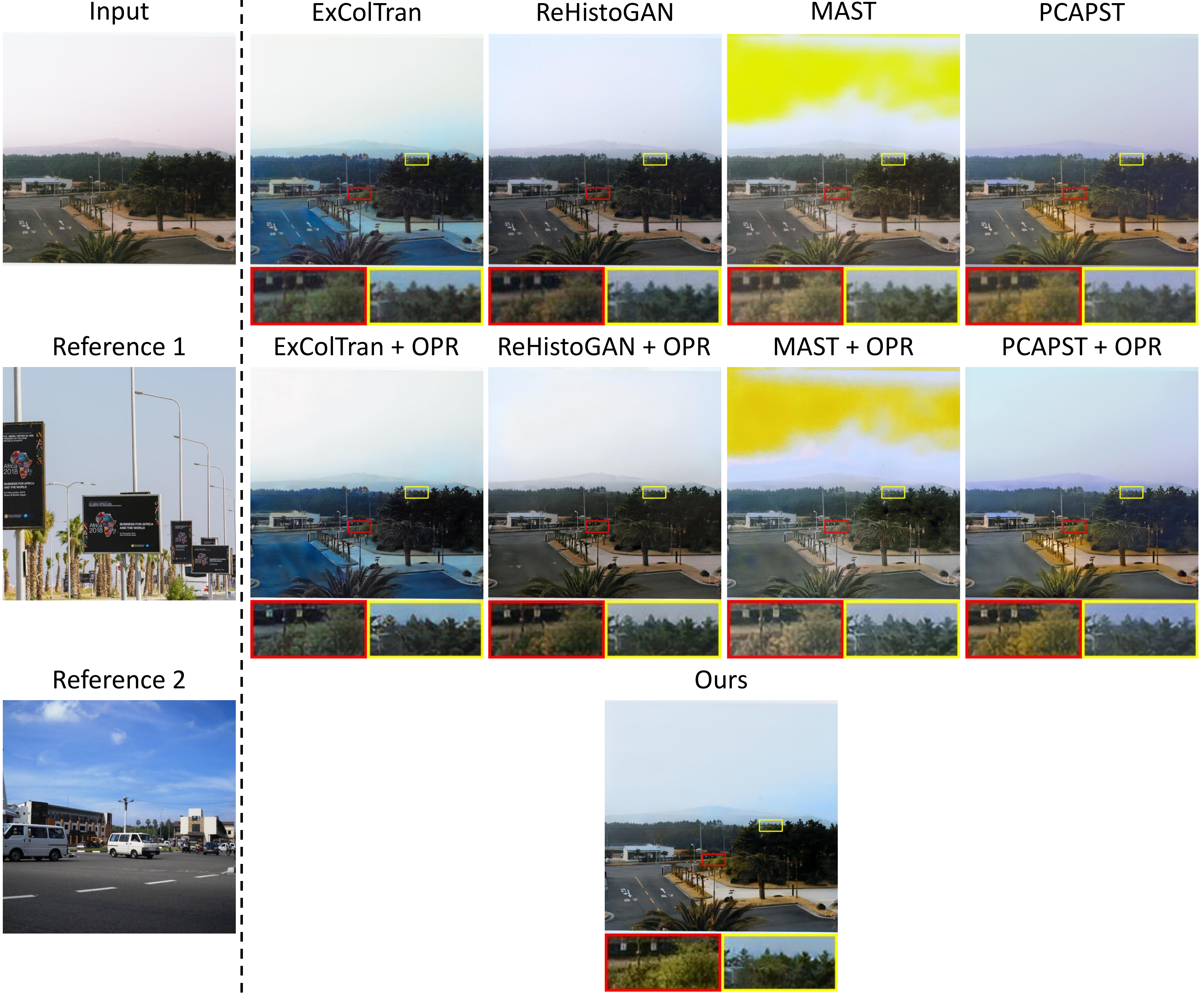}
\vspace{-1em}
\caption{Comparison between only `stylization' and `stylization + enhancement' for all of the stylization baselines: ExColTran\cite{yin2021yes}, ReHistoGAN\cite{afifi2021histogan}, MAST\cite{huo2021manifold}, and PCAPST\cite{chiu2022pca}. OPR\cite{wan2020bringing} is used for the enhancement method.}
\label{fig:5_4_stylization_enhancement_only_stylization}
\vspace{-1em}
\end{figure*}

\begin{table}[t]
    \centering
    \small
    \begin{tabular}{l|ccc}
        \noalign{\hrule height 0.3mm}
        Method & PSNR${\uparrow}$ & SSIM${\uparrow}$ & LPIPS${\downarrow}$\\
        \hline
        ExColTran \cite{yin2021yes} & \underline{20.1637} & \textbf{0.8123} & 0.2735\\
        ReHistoGAN \cite{afifi2021histogan} & 19.8240 & \underline{0.8044} & 0.2467\\
        MAST \cite{huo2021manifold} & 17.5653 & 0.7685 & 0.2726\\
        PCAPST \cite{chiu2022pca} & 17.3873 & 0.7834 & 0.2671\\
        Average & 18.7351 & 0.7922 & 0.2650\\
        \hline
        ExColTran \cite{yin2021yes} + OPR & 18.9152 & 0.7144 & 0.3044\\
        ReHistoGAN \cite{afifi2021histogan} + OPR & 18.9767 & 0.7220 & 0.2748\\
        MAST \cite{huo2021manifold} + OPR & 18.1063 & 0.7042 & 0.2855\\
        PCAPST \cite{chiu2022pca} + OPR & 17.8949 & 0.7061 & 0.2874\\
        Average & 18.4733 & 0.7117 & 0.2880\\
        \hline
        ExColTran \cite{yin2021yes} + OPR-R & 19.5796 & 0.7885 & 0.2563\\
        ReHistoGAN \cite{afifi2021histogan} + OPR-R & 20.0458 & 0.7987 & \underline{0.2109}\\
        MAST \cite{huo2021manifold} + OPR-R & 19.0148 & 0.7853 & 0.2270\\
        PCAPST \cite{chiu2022pca} + OPR-R & 19.1731 & 0.7908 & 0.2197\\
        Average & 19.4533 & 0.7908 & 0.2285\\
        \hline
        OPR-R + ExColTran \cite{yin2021yes} & 20.1565 & 0.7989 & 0.2400\\
        OPR-R + ReHistoGAN \cite{afifi2021histogan} & 19.8990 & 0.7932 & 0.2115\\
        OPR-R + MAST \cite{huo2021manifold} & 17.6050 & 0.7591 & 0.2374\\
        OPR-R + PCAPST \cite{chiu2022pca} & 17.7387 & 0.7702 & 0.2317\\ 
        Average & 18.8498 & 0.7803 & 0.2302\\
       \hline
        Ours & \textbf{21.2212} & 0.7919 & \textbf{0.2027}\\
        \noalign{\hrule height 0.3mm}
    \end{tabular}
    \vspace{-0.7em}
    \caption{Quantitative results of modernization on synthetic dataset.}
    \label{tab:quantitative_synthetic_data}
    \vspace{-1em}
\end{table}

\begin{table}[t]
    \centering
    \small
    \begin{tabular}{l|cc}
        \noalign{\hrule height 0.3mm}
        Method & NIQE${\downarrow}$ & BRISQUE${\downarrow}$\\
        \hline
        OPR \cite{wan2020bringing} & 4.8705 & 21.4588\\
        OPR-R & 3.8616 & 25.2025\\
        \hline
        ExColTran \cite{yin2021yes} & 3.3852 & 28.5359\\
        ReHistoGAN \cite{afifi2021histogan} & \textbf{3.2115} & 32.4907\\
        MAST \cite{huo2021manifold} & 3.4060 & 26.6633\\
        PCAPST \cite{chiu2022pca} & \underline{3.2264} & 24.8812\\
        Average & 3.3073 & 28.1428\\
        \hline
        ExColTran \cite{yin2021yes} + OPR & 4.9415 & 18.8971\\
        ReHistoGAN \cite{afifi2021histogan} + OPR & 4.8051 & 26.2557\\
        MAST \cite{huo2021manifold} + OPR & 4.8111 & 18.9555\\
        PCAPST \cite{chiu2022pca} + OPR & 4.7094 & 18.9860 \\
        Average & 4.8168 & 20.7736\\
        \hline
        OPR + ExColTran \cite{yin2021yes} & 5.1461 & 22.7619\\
        OPR + ReHistoGAN \cite{afifi2021histogan} & 3.3192 & 33.7882\\
        OPR + MAST \cite{huo2021manifold} & 4.7573 & 22.5228\\
        OPR + PCAPST \cite{chiu2022pca} & 4.7087 & 22.7718\\
        Average & 4.4829 & 25.4612\\
        \hline
        Ours - Single & 3.4737 & \underline{15.5152}\\
        \hline
        Ours - Multiple & 3.4487 & \textbf{15.4180}\\
        \noalign{\hrule height 0.3mm}
    \end{tabular}
    \vspace{-0.7em}
    \caption{Quantitative results of modernization on real old photos.}
    \label{tab:quantitative_real_old_photos}
    \vspace{-1.1em}
\end{table}

\vspace{0.3em}
\noindent
\textbf{Comparison between a sequence of `stylization + enhancement', `enhancement + stylization', and only `stylization'.}
We show additional results of performing old photo modernization using three different variations: 1) `stylization + enhancement', 2) `enhancement + stylization', and 3) `stylization'.
Specifically, we provide additional quantitative results on a synthetic dataset and real old photos, and qualitative results on real old photos to show that `stylization + enhancement' is the best baseline over other variations. 
Table \ref{tab:quantitative_synthetic_data} shows the quantitative results of old photo modernization on the synthetic dataset, where on average, the `stylization + enhancement' baselines achieve better results than other baselines' variations.
Even though `ExColTran' \cite{yin2021yes} achieves higher PSNR and SSIM than other baselines, we still choose `stylization + enhancement' as the main baselines since this sequence provides the most stable results for all of the baselines.
In addition, Table. \ref{tab:quantitative_synthetic_data} also shows that pre-trained OPR \cite{wan2020bringing} is only better for real old photo evaluation, while worse for synthetic data evaluation.
Compared to retrained OPR (OPR-R), using the pre-trained OPR (OPR) decreases PSNR, SSIM, and increases LPIPS by an average of 0.980, 0.079, and 0.060 respectively for all stylization baselines.

The same observation can also be seen in the quantitative results of modernization on real old photos shown in Table. \ref{tab:quantitative_real_old_photos}.
On average, other variations: `enhancement + stylization' and `stylization' achieves better (lower) average NIQE \cite{mittal2012making} scores and worse (higher) BRISQUE \cite{mittal2012no} scores compared to `stylization + enhancement'.
In our observation, the BRISQUE score is a better metric for real old photo evaluation that better matches the qualitative results of modernization on real old photos.
For example, we show the qualitative results of OPR and OPR-R in Fig. \ref{fig:5_1_opr_training}, where OPR achieves better results.
However, the NIQE performance of OPR-R is better than OPR, even though the qualitative results show the opposite.
In addition, the qualitative results of `ReHistoGAN \cite{afifi2021histogan} + OPR' are also better than `OPR + ReHistoGAN \cite{afifi2021histogan}' shown in Fig. \ref{fig:5_2_stylization_enhnancement_part_1}.
All in all, we show the qualitative results of both `stylization + enhancement' and `enhancement + stylization' for every baseline in Fig. \ref{fig:5_2_stylization_enhnancement_part_1} and Fig. \ref{fig:5_3_stylization_enhnancement_part_2}, where the results show that `stylization + enhancement' achieves better results than the `enhancement + stylization'.
In addition, we also show the qualitative results of `stylization + enhancement' and only `stylization' for every baseline in Fig. \ref{fig:5_4_stylization_enhancement_only_stylization}.
The results show that using enhancement (`stylization + enhancement') improves the stylization output, making it look cleaner and sharper, and have a better color (yellow and red boxes).
Thus, choosing `stylization + enhancement' as the sequence for the baselines is the better choice to provide a fair comparison.

\begin{figure*}[ht]
\centering
\includegraphics[width=0.9\textwidth]{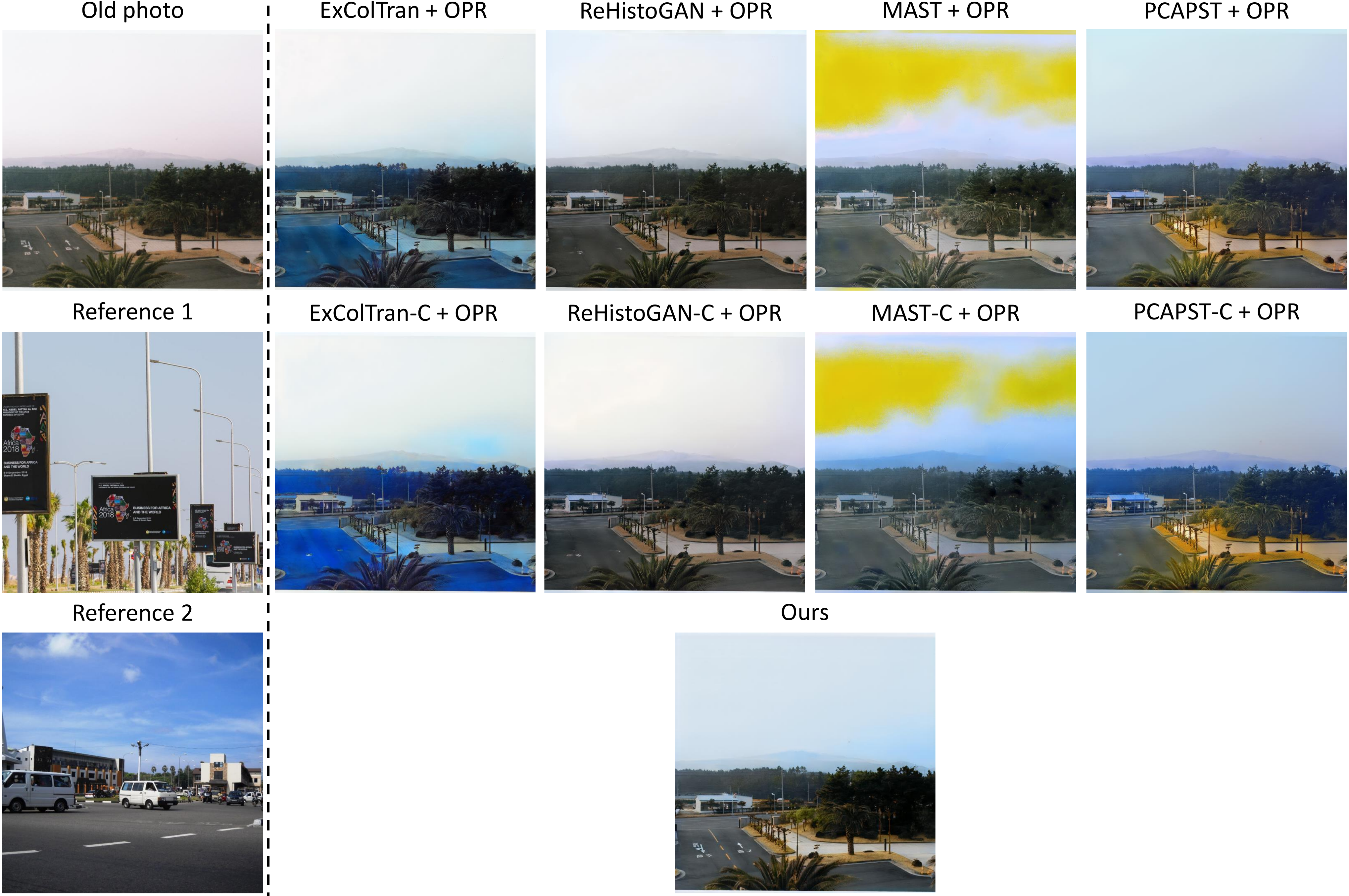}
\vspace{-0.5em}
\caption{The results of multi-reference stylization using spatial concatenation and single-reference stylization. Baseline, e.g., ReHistoGAN denotes the result of performing single-reference stylization using reference 1. Meanwhile, Baseline-C, e.g., ReHistoGAN-C denotes the result of performing multi-reference stylization by spatially concatenating references 1 and 2 into a single reference.}
\label{fig:5_5_spatial_concatenation}
\vspace{-0.5em}
\end{figure*}

\vspace{0.3em}
\noindent
\textbf{The results of spatial concatenation as the baseline.}
One naive way to make single-reference baselines able to handle multi-reference is by spatially concatenating multiple references into a single reference.
We show the results of spatial concatenation baselines in Fig. \ref{fig:5_5_spatial_concatenation}.
The results show that using a single reference for all of the baselines is mostly better compared to using the spatial concatenation of multiple references since the results of concatenation look more unnatural in most cases, e.g., unnatural tree color.
This is likely caused by the inability of the baselines to perform local style/color transfer properly.

\section{Additional Ablation Studies}

\begin{figure*}[ht]
\centering
\includegraphics[width=0.95\textwidth]{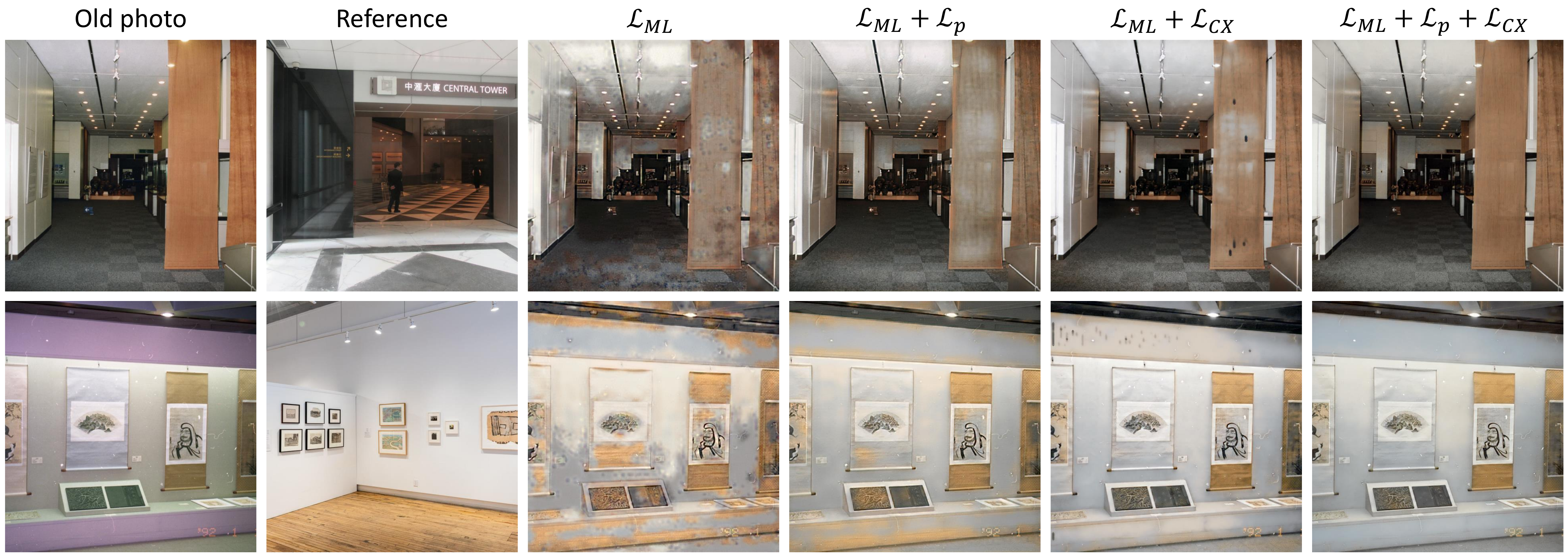}
\vspace{-1em}
\caption{Ablation study on loss functions for single stylization subnet.}
\label{fig:6_1_single_stylization_subnet_loss_ablation}
\vspace{-0.5em}
\end{figure*}

\noindent
\textbf{Ablation study on loss functions for single stylization subnet.}
Fig. \ref{fig:6_1_single_stylization_subnet_loss_ablation} shows the visual results of the ablation study on loss functions for the single stylization subnet.
Training the subnet with only $\mathcal{L}_{ML}$ is insufficient, making the subnet produce severe artifacts far from photorealistic results.
Meanwhile, adding $\mathcal{L}_{p}$ can reduce the artifact and enable the subnet to achieve faithful stylization at the semantic level, e.g., the wall and the painting, but the results still have some style artifacts.
Changing $\mathcal{L}_{p}$ to $\mathcal{L}_{CX}$ can produce better semantic style transfer with fewer artifacts. 
However, it produces weird artifacts, e.g., black dots in the wall region of the second row in Fig. \ref{fig:6_1_single_stylization_subnet_loss_ablation}.
By applying all three losses $\mathcal{L}_{ML}$, $\mathcal{L}_{p}$, and $\mathcal{L}_{CX}$ to train the subnet, we achieve the best photorealistic style transfer results that can faithfully stylize the old photos both on pixel and semantic levels, and can perform local style transfer without any semantic segmentation mask.

\begin{figure*}[ht]
\centering
\includegraphics[width=0.78\textwidth]{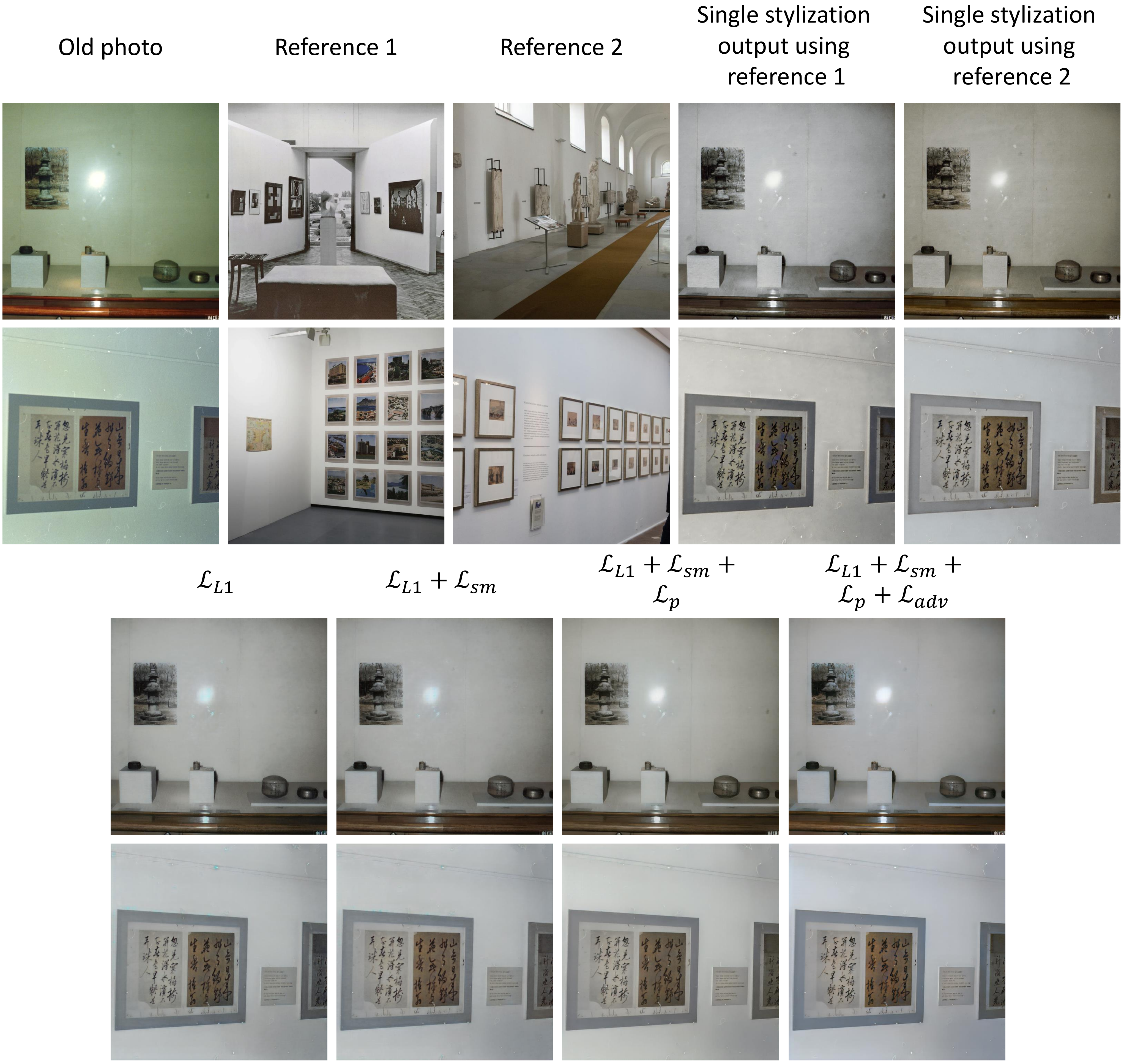}
\vspace{-1em}
\caption{Ablation study on loss functions for merging-refinement subnet.}
\label{fig:6_2_merging_refinement_loss_ablation}
\vspace{-1em}
\end{figure*}

\noindent
\textbf{Ablation study on loss functions for merging-refinement subnet.}
Fig. \ref{fig:6_2_merging_refinement_loss_ablation} shows the visual results of the ablation study on loss functions for the merging-refinement subnet.
Training the subnet with only reconstruction loss $\mathcal{L}_{L1}$ can make the subnet produce accurate merging and better refinement.
However, it produces several artifacts, e.g., rough textures around the wall regions.
Even though adding the local smoothness loss $\mathcal{L}_{sm}$ can produce spatially smooth output, it still contains some artifacts, e.g., the bluish color around the painting frame in the second row of Fig. \ref{fig:6_2_merging_refinement_loss_ablation}.
All of the artifacts can be removed by additionally adding a perceptual loss $\mathcal{L}_{p}$, but it has dull and unattractive (unsaturated) colors and blurry texture.
Adding a GAN loss $\mathcal{L}_{adv}$ to the loss function further makes the modernization results more realistic so that the texture becomes sharper and the saturation of color increases, making the output look more like modern images.

\begin{figure*}[ht]
\centering
\includegraphics[width=0.6\textwidth]{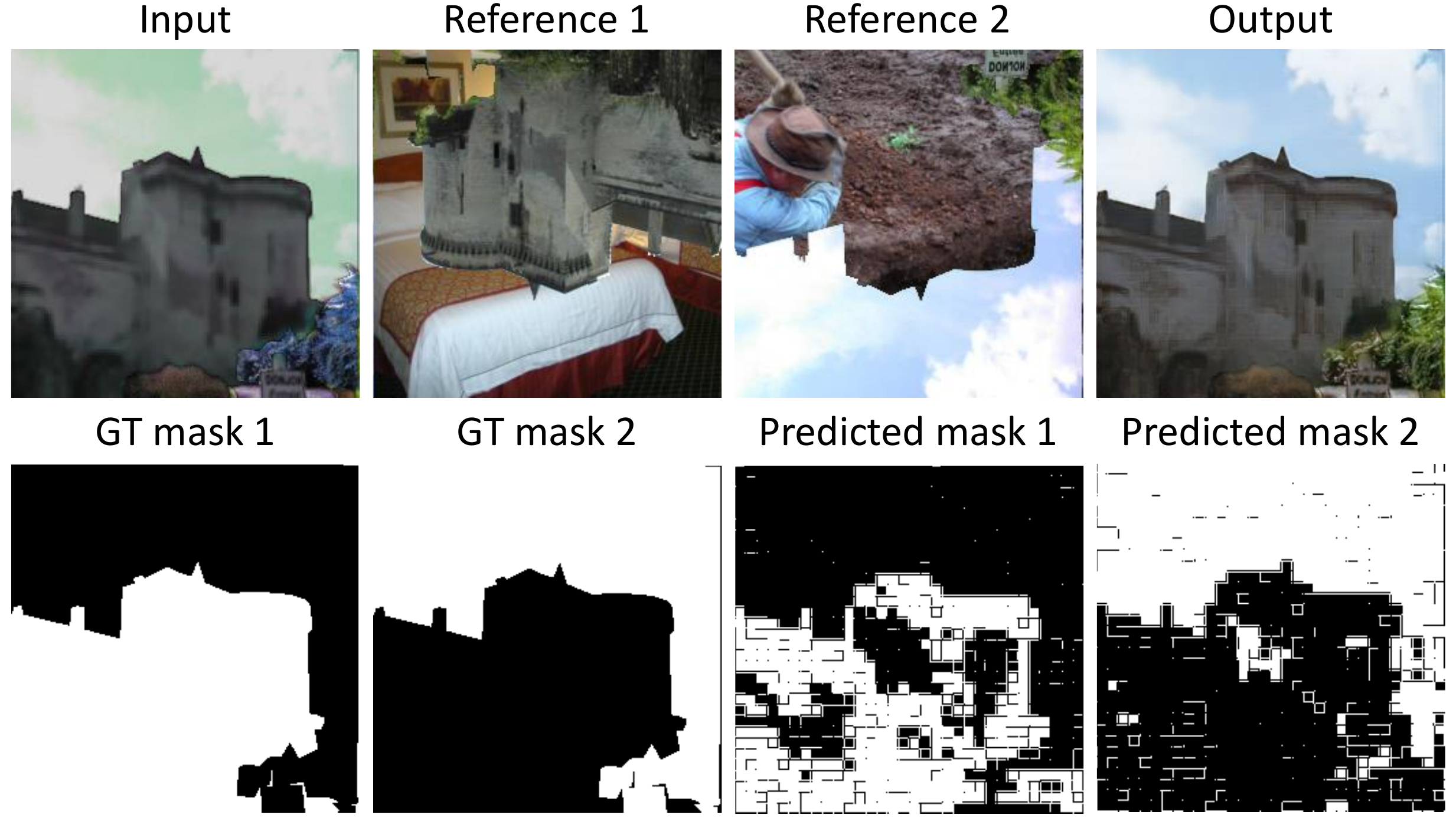}
\vspace{-0.5em}
\caption{Study on the capability of merging-refinement subnet $\mathcal{M}$ to select relevant regions from multiple references to transfer their styles to the corresponding regions in the input.}
\label{fig:6_3_merging_refinement_synthetic}
\vspace{-1em}
\end{figure*}

\noindent
\textbf{Exploration study for the merging-refinement subnet.}
In this study, we explore the capability of the merging-refinement subnet $\mathcal{M}$ in selecting relevant regions from multiple references to transfer their styles to the corresponding regions in the input.
To evaluate this capability, we use a synthetic sample generated using our synthetic data generation pipeline, where we can get the ground truth segmentation mask.
Since our merging-refinement subnet uses spatial attention, we can generate the prediction mask by simple thresholding of the attention weight.
This prediction mask denotes the regions in the input where the corresponding style from multiple references will be transferred to.
Furthermore, we can use the mIoU (mean intersection over union) between the predicted masks and ground truth masks to measure the accuracy of the merging-refinement subnet $\mathcal{M}$.
As shown in Fig. \ref{fig:6_3_merging_refinement_synthetic}, our $\mathcal{M}$ can select relevant regions in the input where it achieves an average of 70.70\% mIoU for both predictions.

\section{Study on the Method’s Capability}

\begin{figure*}[ht]
\centering
\includegraphics[width=0.75\textwidth]{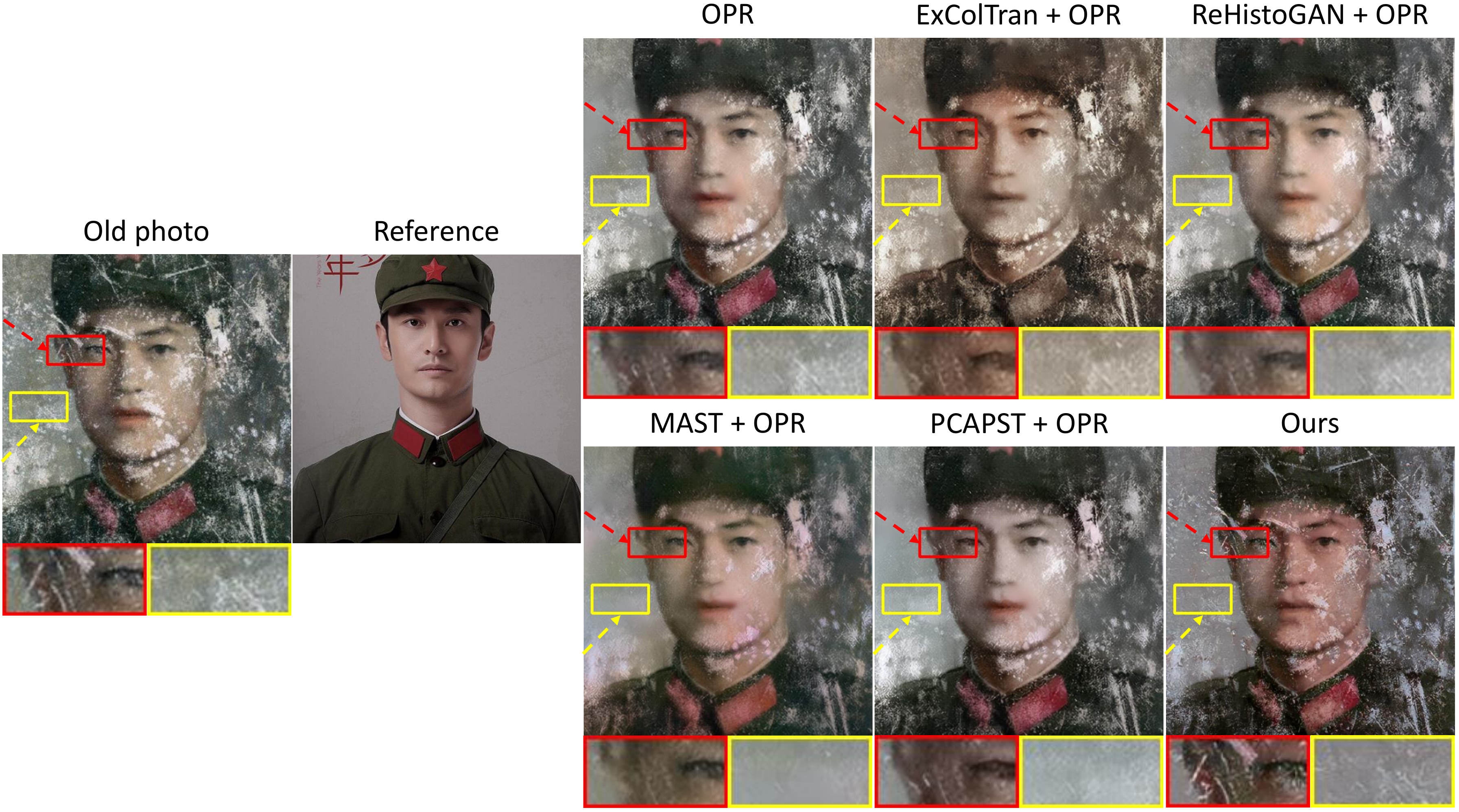}
\vspace{-0.5em}
\caption{Results of our method compared to other baselines on an old photo from RealOld dataset \cite{xu2022pik} with severe structured and unstructured degradations.}
\label{fig:7_0_severe_ud_sd}
\vspace{-1em}
\end{figure*}

\noindent
\textbf{Nature and capability of the enhancement in this work.}
In this work, the enhancement primarily focuses on unstructured degradation (UD) restoration such as deblurring, denoising, and artifact removal, commonly found in old photos. 
The capability of our enhancement can be seen in Fig. \ref{fig:5_4_stylization_enhancement_only_stylization}, where the output of our method is sharper and less noisy compared to the baselines denoting better enhancement capability.
Despite primarily focusing on UD, we find that our method can still generalize to some extent to structured degradation (SD).
We provide additional results on an old photo from RealOld dataset \cite{xu2022pik} with severe SD and UD in Fig. \ref{fig:7_0_severe_ud_sd} to better show the enhancement capability of our method.
All the stylization baselines coupled with OPR can restore SD (scratches and holes) better than ours (red boxes) since it is explicitly trained for such degradations, even though the baselines also fail to remove all SDs like ours.
Nevertheless, our method can enhance the image by restoring small scratches and UD (blur and noise) better than the baselines without excessive blurry artifacts like the results of MAST + OPR (yellow boxes).

\begin{figure*}[ht]
\centering
\includegraphics[width=0.82\textwidth]{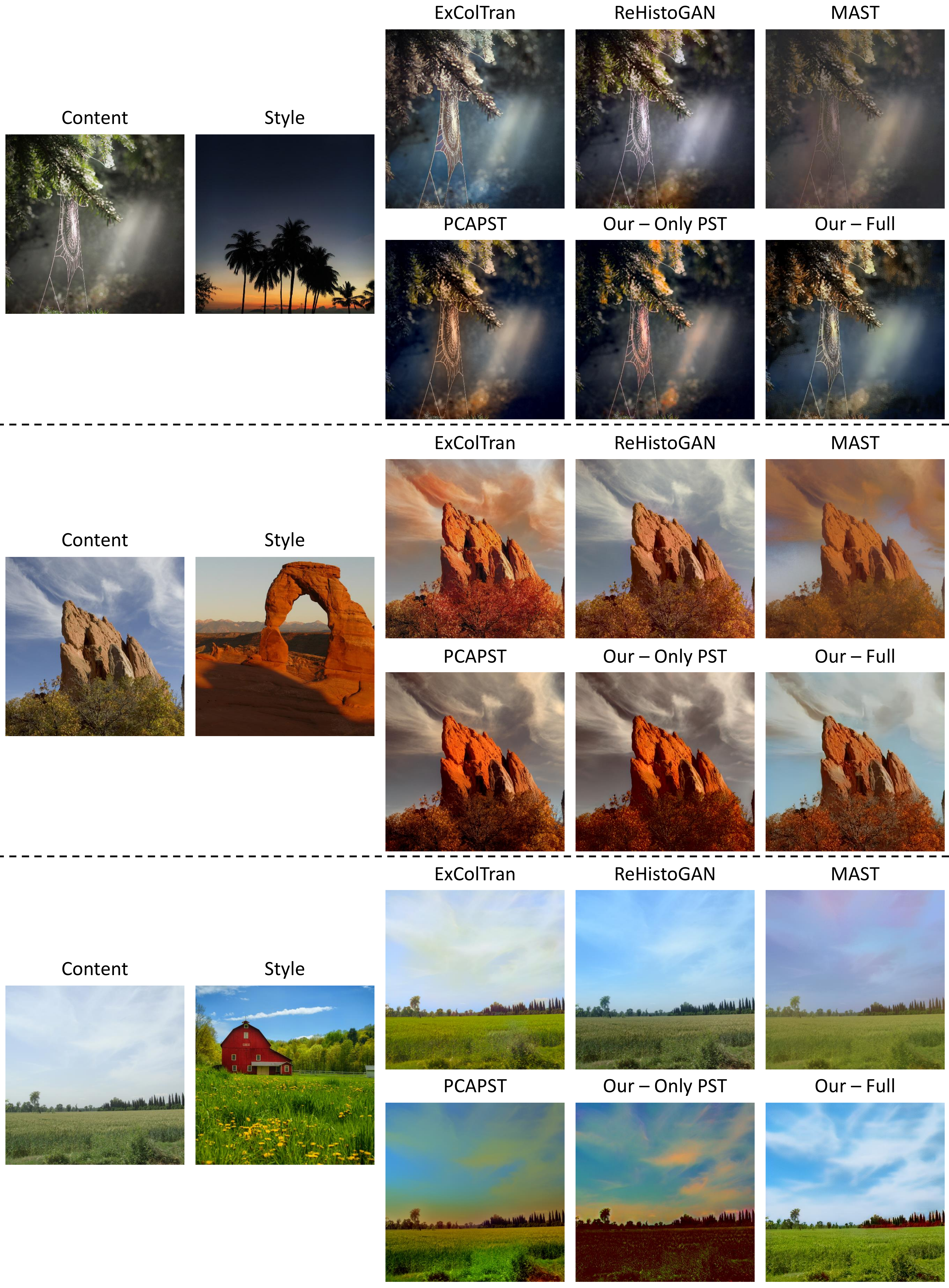}
\vspace{-1em}
\caption{Comparison of stylization results on modern images. Our MROPM-Net, denoted as Our -- Full, achieves the best local PST results on all examples compared to other PST baselines such as MAST \cite{huo2021manifold} and PCAPST \cite{chiu2022pca}, and other baselines such as ExColTran \cite{yin2021yes} and ReHistoGAN \cite{afifi2021histogan}. Our -- Only PST denotes the results of PST using our PST network (without style code predictor and merging-refinement subnet).}
\label{fig:7_1_modern_images}
\vspace{-1em}
\end{figure*}

\noindent
\textbf{Stylization on modern images.}
We provide stylization results on modern photos which have no degradations using our network (MROPM-Net) and other stylization baselines.
As shown in Fig. \ref{fig:7_1_modern_images}, our MROPM-Net (denoted as Our -- Full) achieves the best local style transfer on all images.
In addition, our PST network (without style code predictor and merging-refinement subnet) achieves faithful stylization as shown in the first and second examples of Fig. \ref{fig:7_1_modern_images} and is on par with the SOTA PST network (PCAPST \cite{chiu2022pca}) results.

\begin{figure*}[ht]
\centering
\includegraphics[width=0.8\textwidth]{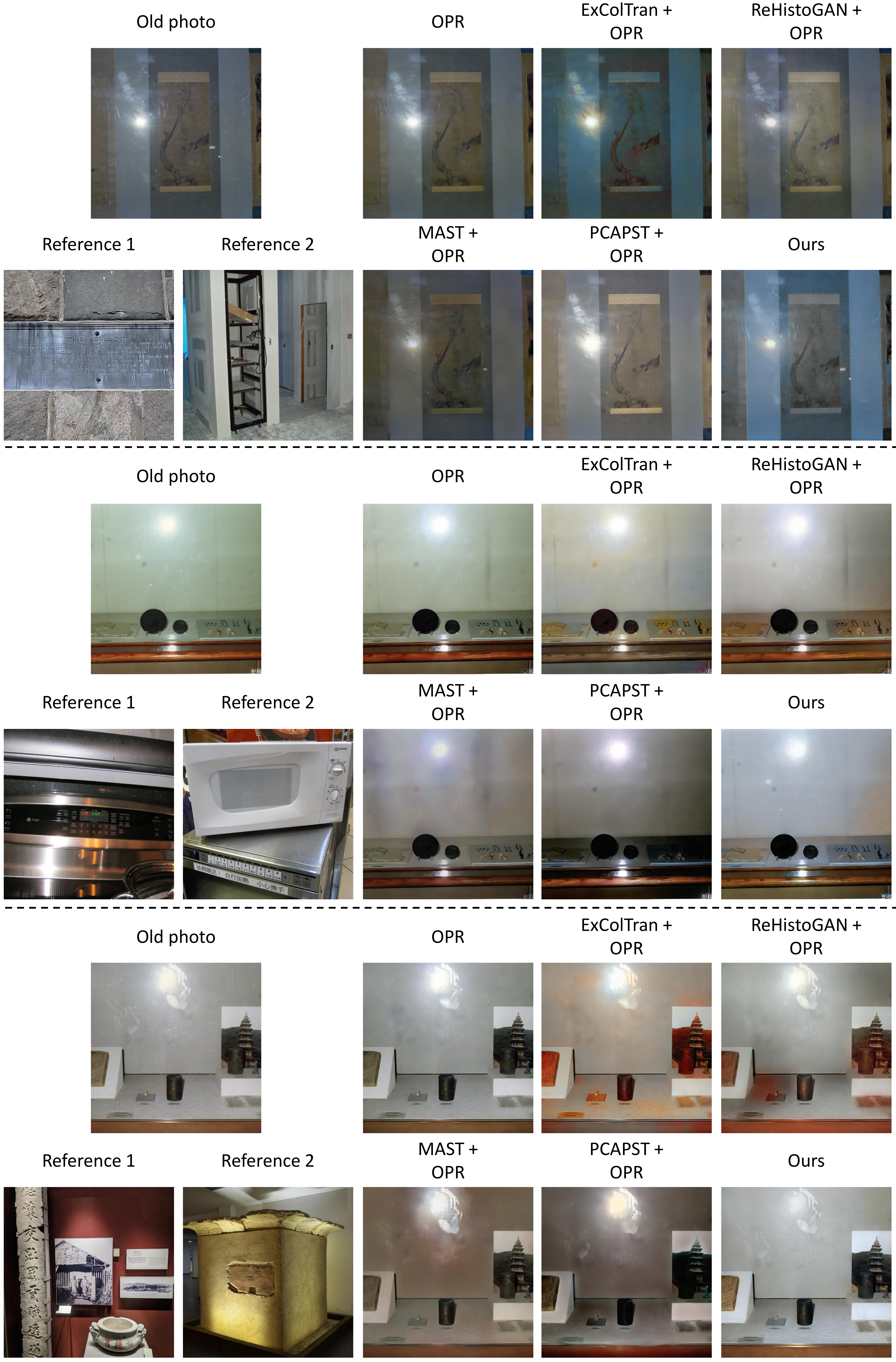}
\vspace{-1em}
\caption{Old photo modernization results using unrelated references. In most cases, our method outperforms baselines (OPR\cite{wan2020bringing}, ExColTran\cite{yin2021yes} + OPR, ReHistoGAN\cite{afifi2021histogan} + OPR, MAST\cite{huo2021manifold} + OPR, and PACPST\cite{chiu2022pca} + OPR) even though the references are unrelated with the old photo. Reference-based baselines use reference 1 as their reference.}
\label{fig:7_4_unrelated_references}
\vspace{-1em}
\end{figure*}

\begin{figure*}[ht]
\centering
\includegraphics[width=0.8\textwidth]{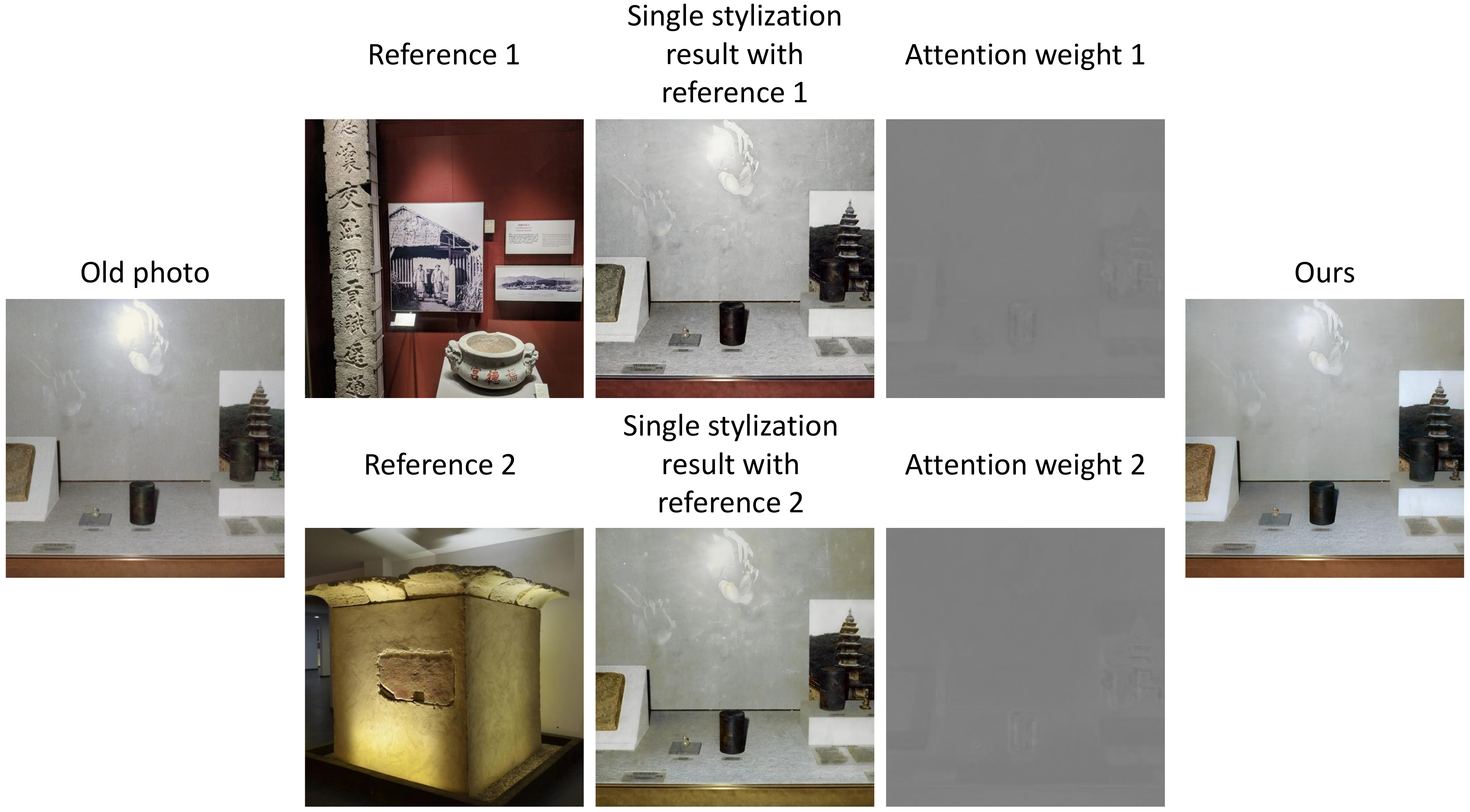}
\vspace{-1em}
\caption{The internal working of our MROPM-Net when handling unrelated references.}
\label{fig:7_4_2_unrelated_references_with_attention}
\vspace{-1em}
\end{figure*}

\noindent
\textbf{Modernization results using unrelated references.}
Fig. \ref{fig:7_4_unrelated_references} shows the visual examples of the robustness of our method when the references are highly unrelated.
Our method outperforms other baselines in terms of handling unrelated references.
We further show the internal working of our MROPM-Net when handling one of the unrelated references in Fig. \ref{fig:7_4_2_unrelated_references_with_attention}.
In this example, our single stylization subnet can robustly find a better style that can modernize the specific regions in the old photos, e.g., the style of concrete to stylize the wall region instead of the red wall in the first reference.
In addition, the merging-refinement subnet can further select the first reference style to stylize the wall region compared to the yellowish wall style in the second reference.

\begin{figure*}[ht]
\centering
\includegraphics[width=0.7\textwidth]{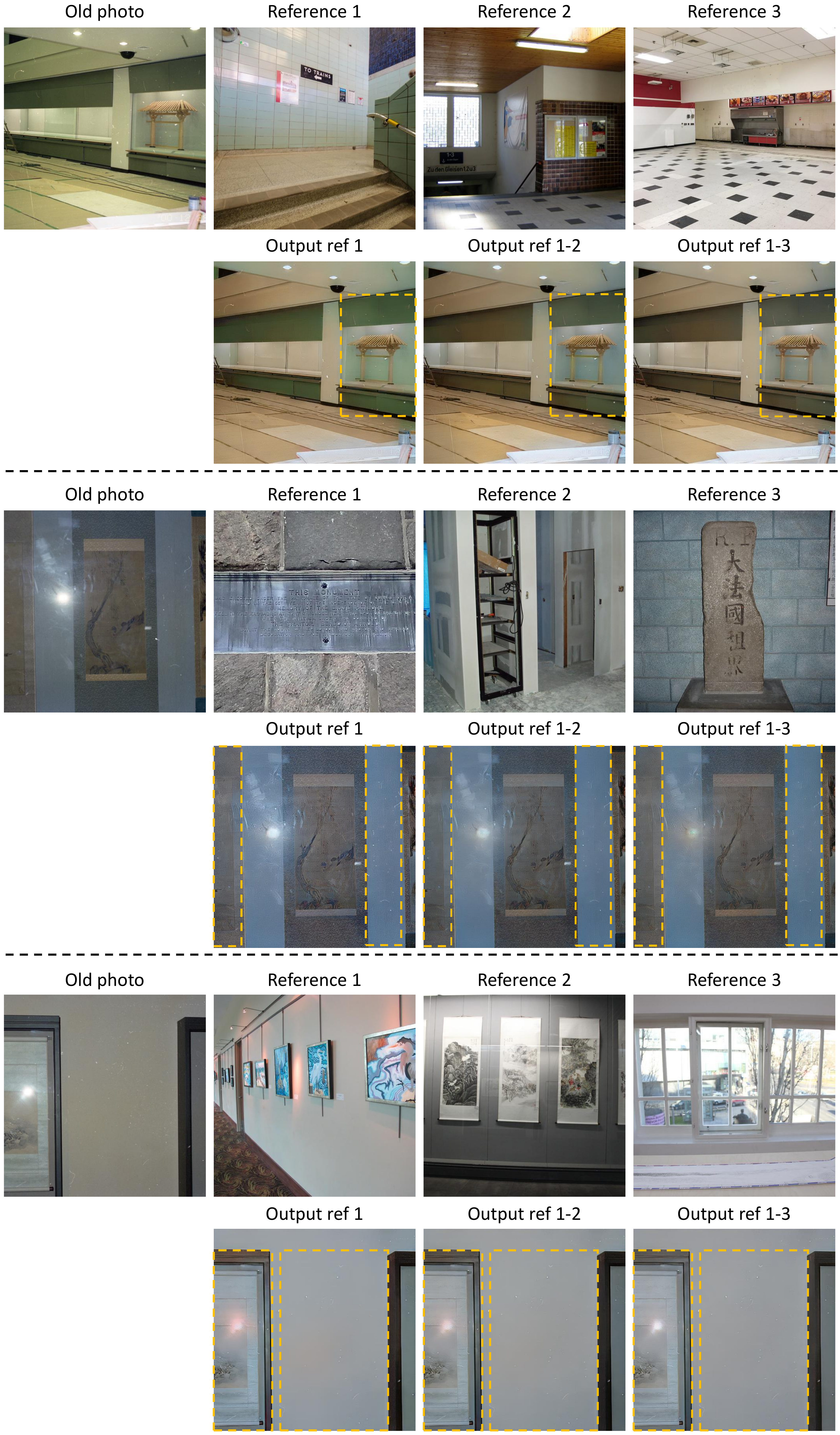}
\vspace{-1em}
\caption{Progressive old photo modernization results using three references. Some regions with distinctive improvements are shown inside yellow boxes.}
\label{fig:7_5_three_references_part_1}
\vspace{-1em}
\end{figure*}

\begin{figure*}[ht]
\centering
\includegraphics[width=0.7\textwidth]{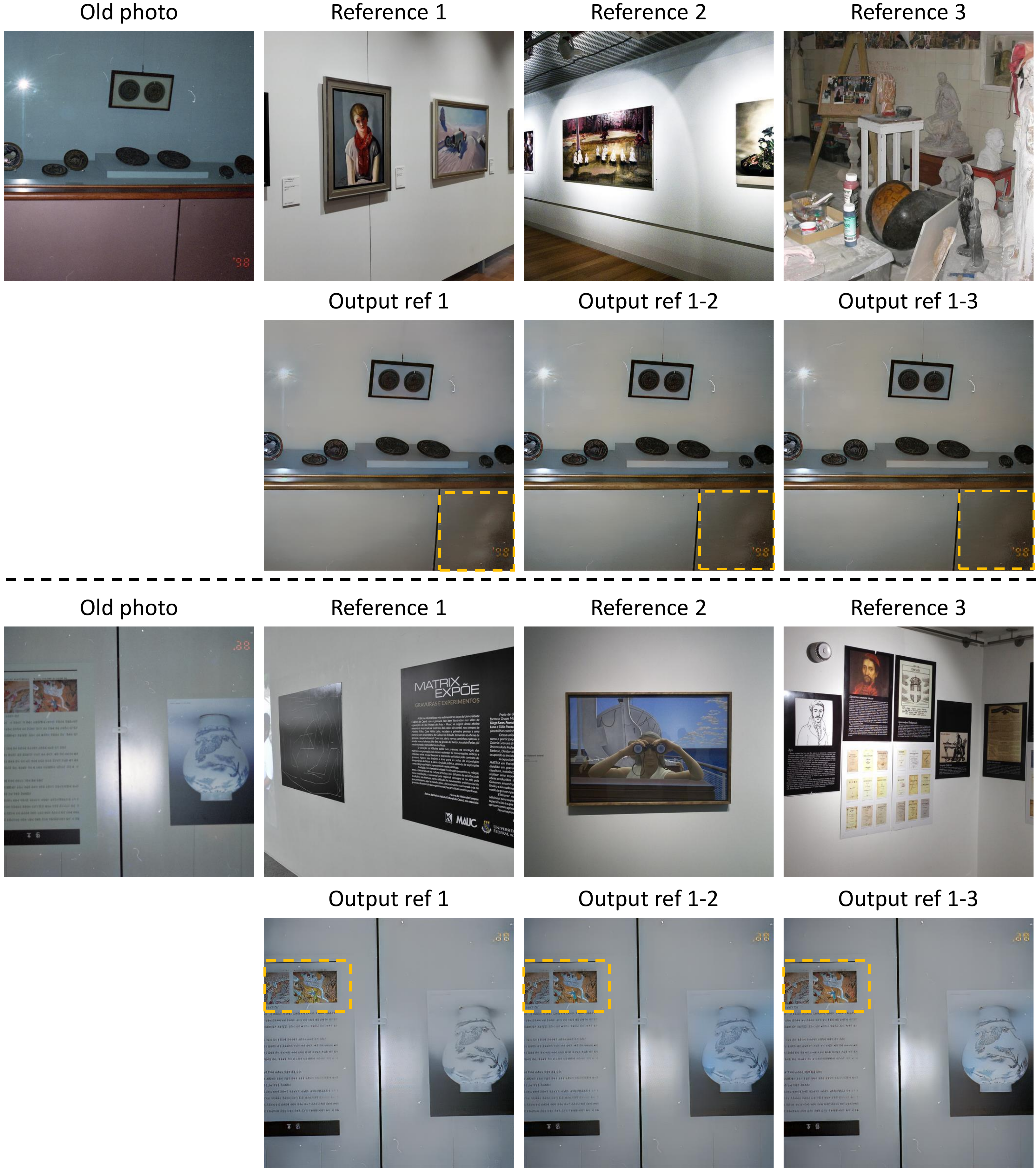}
\vspace{-1em}
\caption{Progressive old photo modernization results using three references. Some regions with distinctive improvements are shown inside yellow boxes.}
\label{fig:7_6_three_references_part_2}
\vspace{-1em}
\end{figure*}

\begin{figure*}[ht]
\centering
\includegraphics[width=0.82\textwidth]{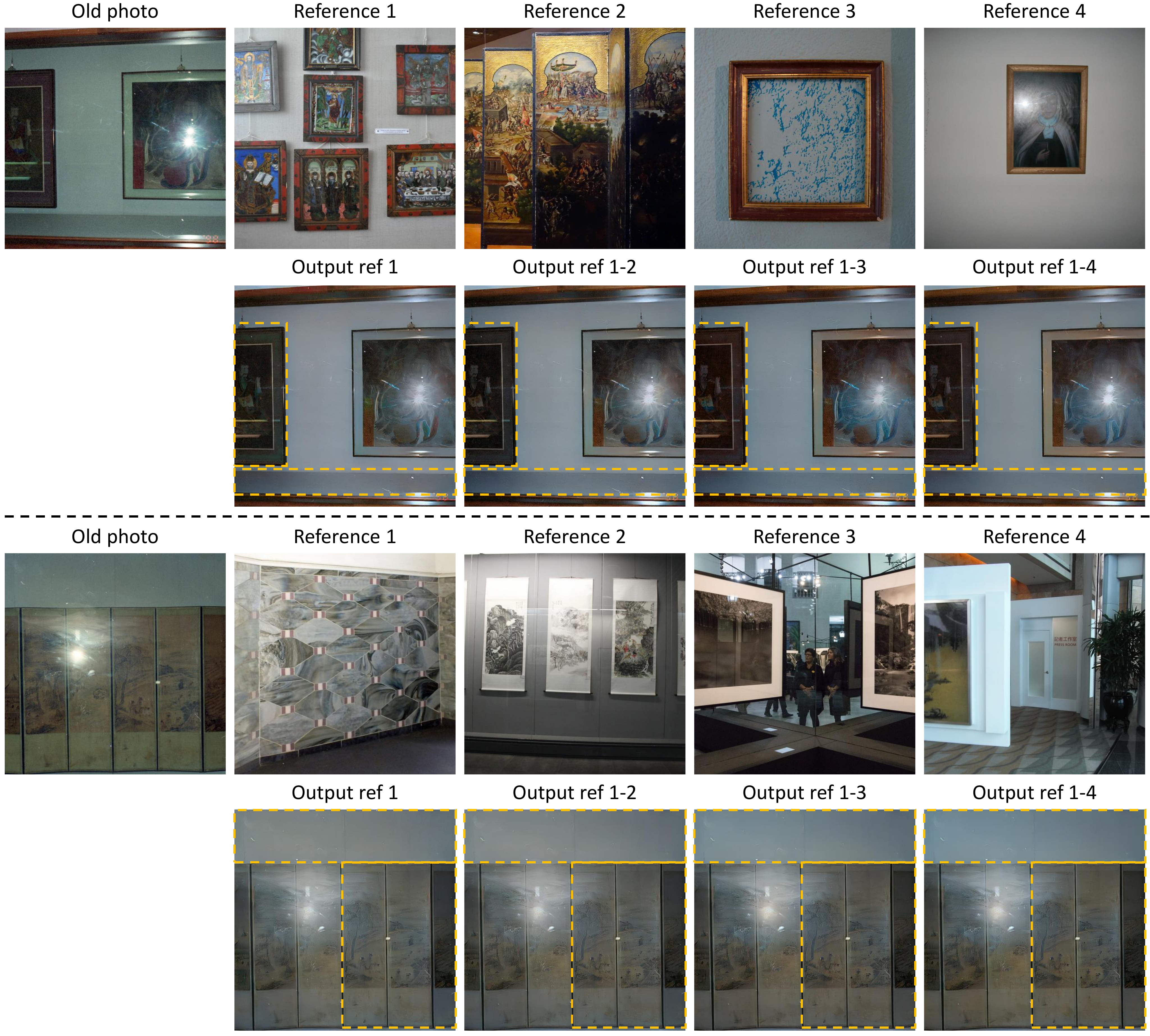}
\vspace{-1em}
\caption{Progressive old photo modernization results using four references. Some regions with distinctive improvements are shown inside yellow boxes.}
\label{fig:7_7_four_references_part_1}
\vspace{-1em}
\end{figure*}

\begin{figure*}[ht]
\centering
\includegraphics[width=0.82\textwidth]{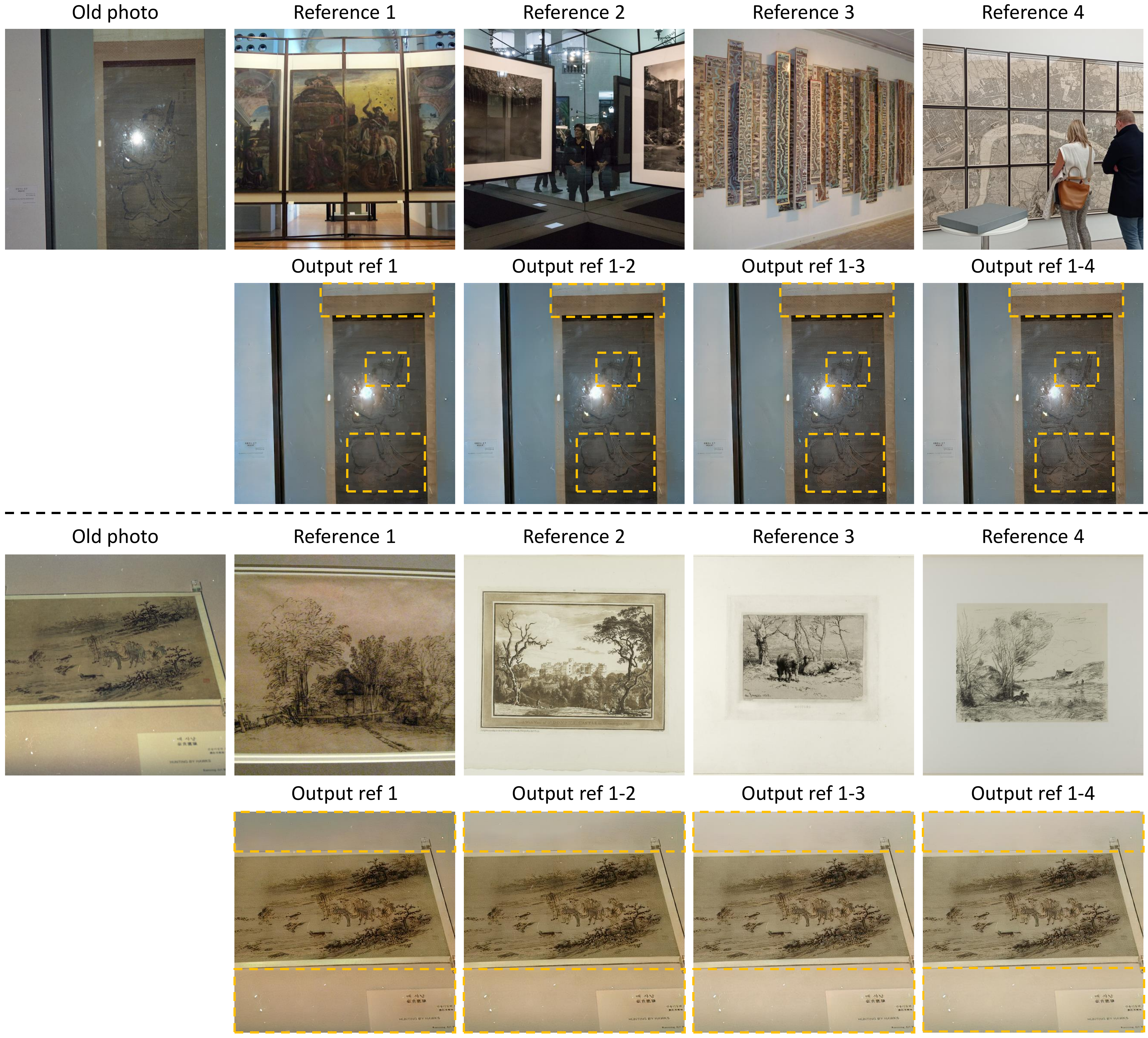}
\vspace{-1em}
\caption{Progressive old photo modernization results using four references. Some regions with distinctive improvements are shown inside yellow boxes.}
\label{fig:7_7_2_four_references_part_2}
\vspace{-1em}
\end{figure*}

\noindent
\textbf{Modernization results using more than two references.}
We provide additional results when using more than two references in Fig. \ref{fig:7_5_three_references_part_1}, Fig. \ref{fig:7_6_three_references_part_2}, Fig. \ref{fig:7_7_four_references_part_1}, and Fig. \ref{fig:7_7_2_four_references_part_2}.
As shown in all of the figures, our MROPM-Net can adaptively select appropriate styles from multiple references to further improve modernization performance.
Some results show distinctive improvement in specific regions shown inside yellow dashed boxes.
In some other results, the overall improvement of the old photos can also be seen outside the yellow dashed boxes.
Users can choose which region is important and accordingly choose references that can improve the specific regions depending on the availability of similar objects in references.
Since using more than four references with the resolution of $1024\times1024$ could not be processed with our GPU (NVIDIA RTX 3090), we resize the images (old photo and references) into the resolution of $512\times512$ to handle more than two references.

\begin{figure*}[ht]
\centering
\includegraphics[width=0.8\textwidth]{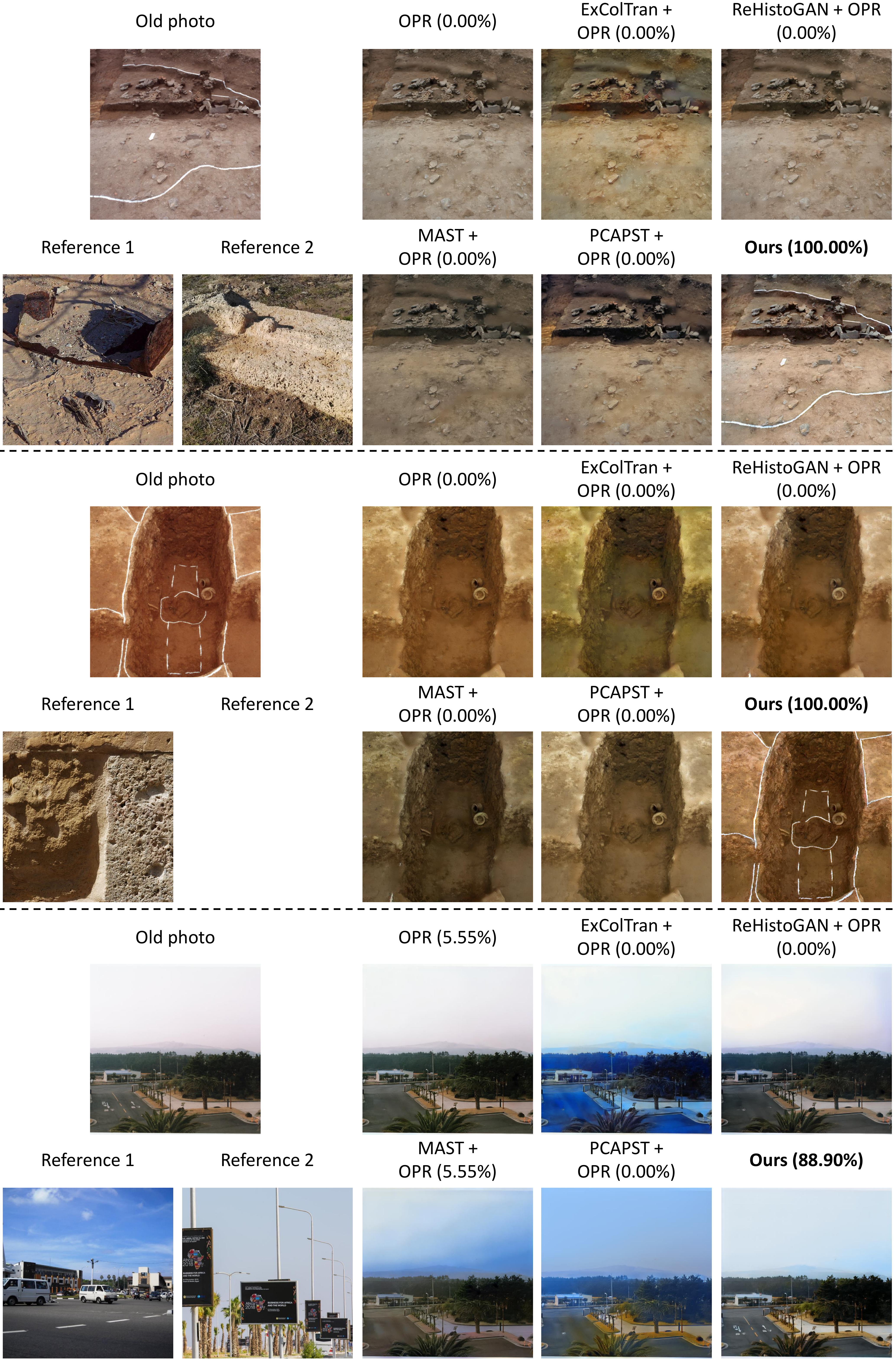}
\vspace{-1em}
\caption{User study results with the percentage of user voting. Our method compares favorably against baselines (OPR \cite{wan2020bringing}, ExColTran\cite{yin2021yes} + OPR, ReHistoGAN\cite{afifi2021histogan} + OPR, MAST\cite{huo2021manifold} + OPR, and PCAPST\cite{chiu2022pca} + OPR). Reference-based baselines use reference 1 as their reference.}
\label{fig:7_8_user_study_1}
\vspace{-1em}
\end{figure*}

\begin{figure*}[ht]
\centering
\includegraphics[width=0.8\textwidth]{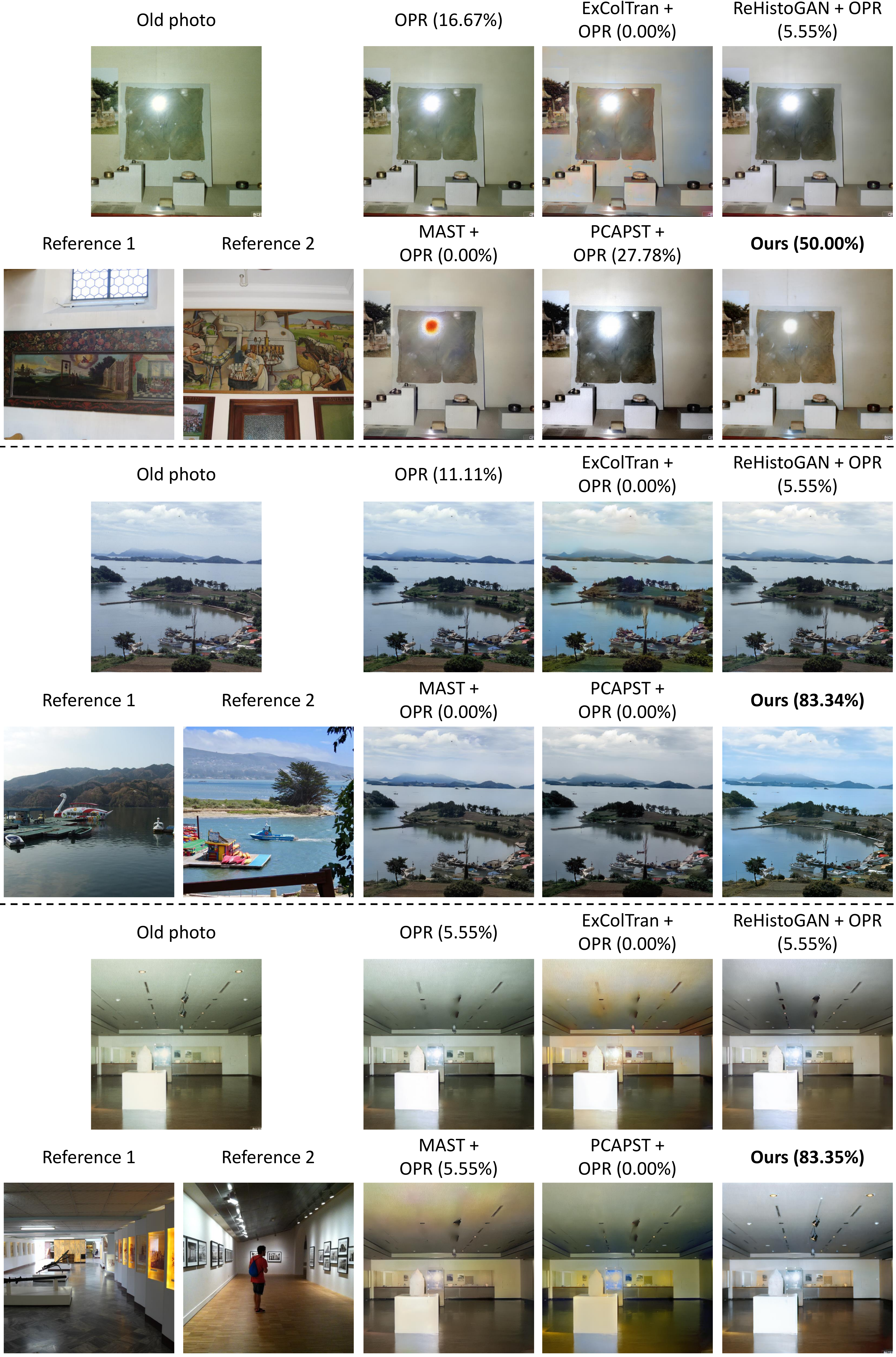}
\vspace{-1em}
\caption{User study results with the percentage of user voting. Our method compares favorably against baselines (OPR \cite{wan2020bringing}, ExColTran\cite{yin2021yes} + OPR, ReHistoGAN\cite{afifi2021histogan} + OPR, MAST\cite{huo2021manifold} + OPR, and PCAPST\cite{chiu2022pca} + OPR). Reference-based baselines use reference 1 as their reference.}
\label{fig:7_9_user_study_2}
\vspace{-1em}
\end{figure*}

\begin{figure*}[ht]
\centering
\includegraphics[width=0.8\textwidth]{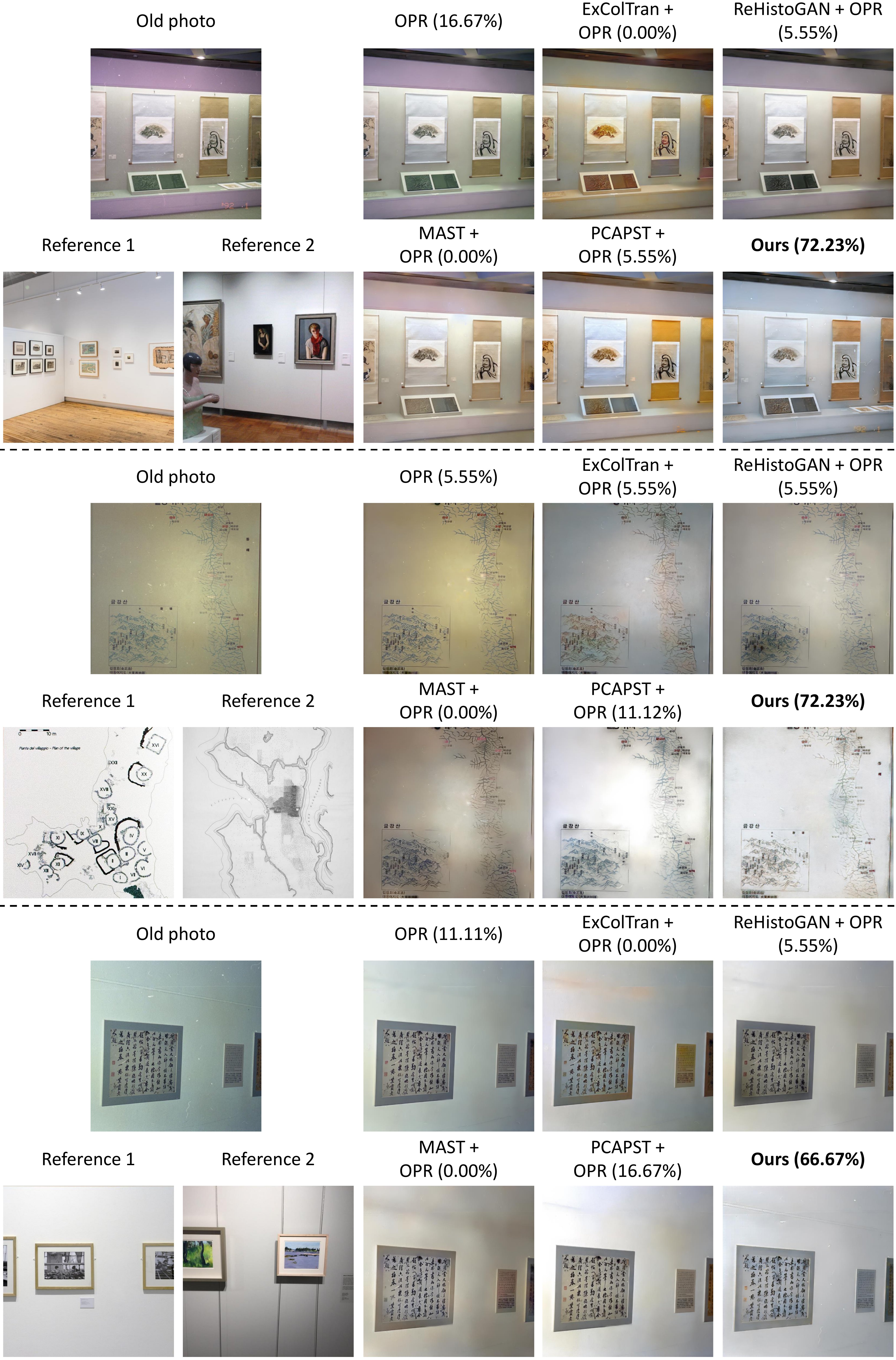}
\vspace{-1em}
\caption{User study results with the percentage of user voting. Our method compares favorably against baselines (OPR \cite{wan2020bringing}, ExColTran\cite{yin2021yes} + OPR, ReHistoGAN\cite{afifi2021histogan} + OPR, MAST\cite{huo2021manifold} + OPR, and PCAPST\cite{chiu2022pca} + OPR). Reference-based baselines use reference 1 as their reference.}
\label{fig:7_10_user_study_3}
\vspace{-1em}
\end{figure*}

\begin{figure*}[ht]
\centering
\includegraphics[width=0.8\textwidth]{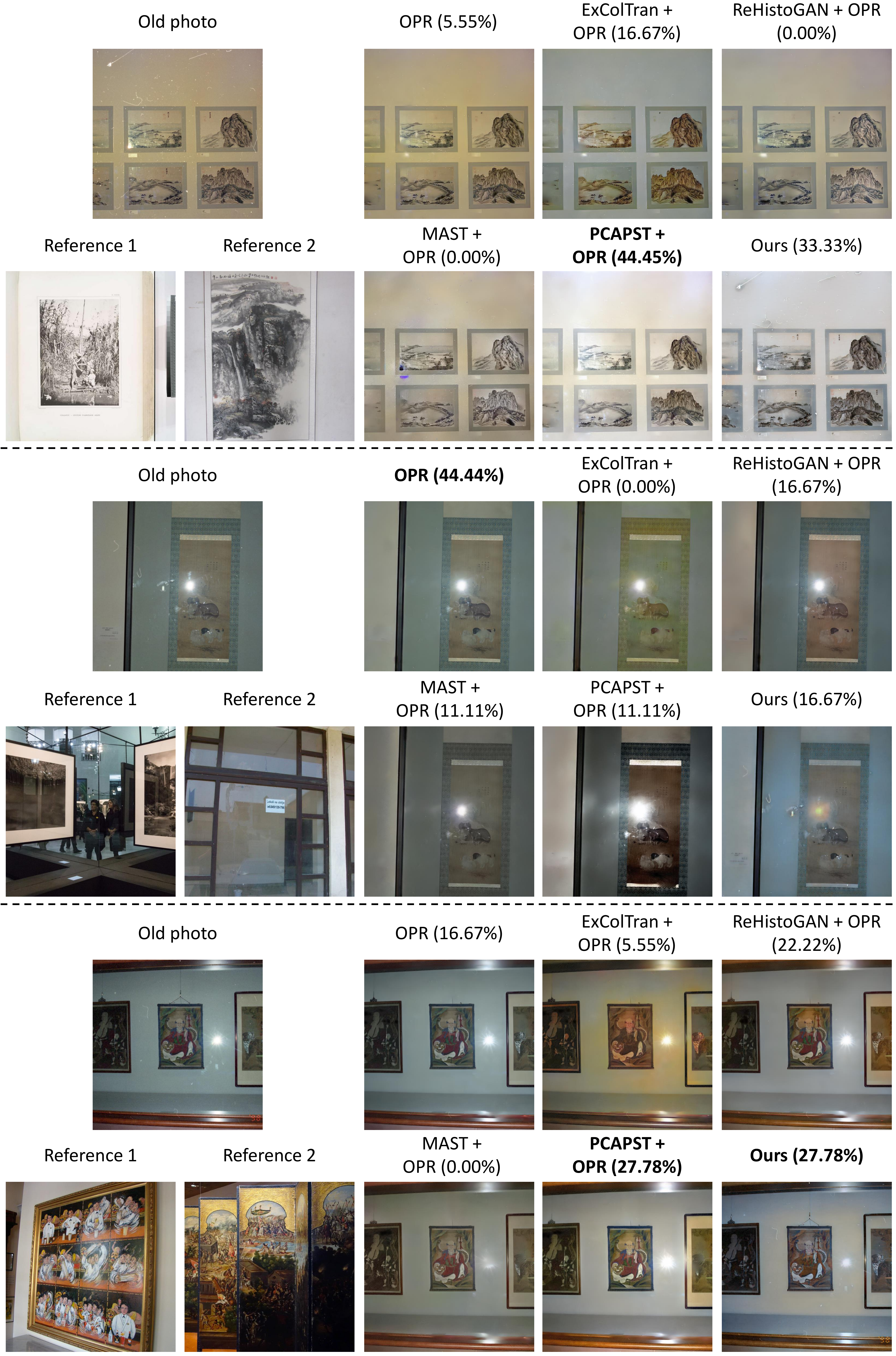}
\vspace{-1em}
\caption{User study results with the percentage of user voting. Our method compares favorably against baselines (OPR \cite{wan2020bringing}, ExColTran\cite{yin2021yes} + OPR, ReHistoGAN\cite{afifi2021histogan} + OPR, MAST\cite{huo2021manifold} + OPR, and PCAPST\cite{chiu2022pca} + OPR). Reference-based baselines use reference 1 as their reference.}
\label{fig:7_11_user_study_4}
\vspace{-1em}
\end{figure*}

\noindent
\textbf{Some examples of user study results.}
We provide some examples of user study results with varying user voting percentages.
The results are shown in Fig. \ref{fig:7_8_user_study_1}, Fig. \ref{fig:7_9_user_study_2}, Fig. \ref{fig:7_10_user_study_3}, and Fig. \ref{fig:7_11_user_study_4}.
In most cases, the results produced by our method are more preferably selected by the users compared to other baselines.

\end{document}